\documentclass{article} % For LaTeX2e
\usepackage{nips14submit_e,times}
\usepackage{hyperref}
\usepackage{url}
\usepackage{graphicx}
\usepackage{caption}
\usepackage{subcaption}
\usepackage{booktabs}
\usepackage{amsmath}
\usepackage{wrapfig}

\newcommand{\caffe}{\texttt{caffe}}

\newcommand{\ning}[1]{}
\newcommand{\trevor}[1]{}

\title{Do Convnets Learn Correspondence?}

\author{
Jonathan Long\hspace{3em}Ning Zhang\hspace{3em}Trevor Darrell \\
University of California -- Berkeley \\
\texttt{\{jonlong, nzhang, trevor\}@cs.berkeley.edu} \\
}

\nipsfinalcopy % Uncomment for camera-ready version

\begin{document}

\maketitle

\begin{abstract}
    Convolutional neural nets (convnets) trained from massive labeled datasets
    \cite{ImageNet} have substantially improved the state-of-the-art in image
    classification \cite{Krizhevsky} and object detection \cite{RossJeff}.
    However, visual understanding requires establishing correspondence on a finer
    level than object category.
    Given their large pooling regions and training from whole-image labels, it
    is not clear that convnets derive their success from an accurate
    correspondence model which could be used for precise localization.
    In this paper, we study the effectiveness of convnet activation features for
    tasks requiring correspondence.
    We present evidence that convnet features localize at a much finer scale
    than their receptive field sizes, that they can be used to perform intraclass
    alignment as well as conventional hand-engineered features, and that they
    outperform conventional features in keypoint prediction on objects from
    PASCAL VOC 2011 \cite{pascal}.

%\ning{Shall we say keypoint detection or keypoint localization? Since we assume bounding box at test time, `detection` might be confusing. }
\end{abstract}

\section{Introduction}

Recent advances in convolutional neural nets \cite{Krizhevsky} dramatically
improved the state-of-the-art in image classification.
Despite the magnitude of these results, many doubted
\cite{YannLeCunsGooglePlusPage} that the resulting features had the spatial
specificity necessary for localization; after all, whole image classification
can rely on context cues and overly large pooling regions to get the job done.
For coarse localization, such doubts were alleviated by record breaking results
extending the same features to detection on PASCAL \cite{RossJeff}.

Now, the same questions loom on a finer scale. Are the modern convnets that
excel at classification and detection also able to find precise correspondences
between object parts?
Or do large receptive fields mean that correspondence is effectively pooled
away, making this a task better suited for hand-engineered features?

In this paper, we provide evidence that convnet features perform at least as
well as conventional ones, even in the regime of point-to-point correspondence,
and we show considerable performance improvement in certain settings, including
category-level keypoint prediction.

\subsection{Related work}
\paragraph{Image alignment}

Image alignment is a key step in many computer vision tasks, including face
verification, motion analysis, stereo matching, and object recognition.
Alignment results in correspondence across different images by removing
intraclass variability and canonicalizing pose.
Alignment methods exist on a supervision spectrum from requiring
manually labeled fiducial points or landmarks, to requiring class labels, to
fully unsupervised joint alignment and clustering models.
Congealing \cite{congealing} is an unsupervised joint alignment method based on
an entropy objective.
Deep congealing \cite{deep_congealing} builds on this idea by
replacing hand-engineered features with unsupervised feature
learning from multiple resolutions.
Inspired by optical flow, SIFT flow \cite{sift-flow} matches densely sampled
SIFT features for correspondence and has been applied to motion prediction and
motion transfer.
In Section \ref{sec:flow}, we apply SIFT flow using deep features for aligning
different instances of the same class.

\paragraph{Keypoint localization}
Semantic parts carry important information for object recognition, object
detection, and pose estimation. In particular, fine-grained categorization,
the subject of many recent works, depends strongly on part localization
\cite{iccv13_keypoint, poof}.
Large pose and appearance variation across examples make
part localization for generic object categories a challenging task.

Most of the existing works on part localization or keypoint prediction focus on
either facial landmark localization \cite{Belhumeur_Localizing_2011} or human
pose estimation.
Human pose estimation has been approached
using tree structured methods
to model the spatial relationships between parts \cite{YangRamananCVPR11,
Min_Sun_ICCV11, Deva_2012}, and also using poselets
\cite{BourdevMalikICCV09} as an intermediate step to localize human keypoints
\cite{kposelet, armlet}.
Tree structured models and poselets may struggle when applied to generic objects
with large articulated deformations and wide shape variance.

\paragraph{Deep learning}
Convolutional neural networks have gained much recent attention due to their
success in image classification \cite{Krizhevsky}. Convnets trained with
backpropagation were initially
succesful in digit recognition~\cite{Lecun89} and OCR~\cite{Lecun98OCR}.
The feature representations learned from large data
sets have been found to generalize well to other image classification tasks
\cite{decaf} and even to object detection \cite{RossJeff, sermanet-cvpr13}.
Recently, Toshev et al.\ \cite{deep_pose} trained a cascade of regression-based
convnets for human pose estimation and Jain et al.\ \cite{HumanPoseICLR} combine
a weak spatial model with deep learning methods.

The latter work trains multiple small, independent convnets on $64 \times 64$ patches
for binary body-part detection. In contrast, we employ a powerful pretained
ImageNet model that shares mid-elvel feature representations among all parts
in Section \ref{sec:pred}.

Several recent works have attempted to analyze and explain this overwhelming
success. Zeiler and Fergus \cite{ZF} provide several heuristic visualizations
suggesting coarse localization ability.
Szegedy et al.\ \cite{Szegedy} show
counterintuitive properties of the convnet representation, and suggest that
individual feature channels may not be more semantically meaningful than other
bases in feature space.
A concurrent work \cite{Fischer}
compares convnet features with SIFT in a standard descriptor matching task.
This work illuminates and extends that comparison by providing visual analysis
and by moving beyond single instance matching to intraclass correspondence and keypoint prediction.

\subsection{Preliminaries}

We perform experiments using a network architecture almost
identical\footnote{Ours reverses the order of the response normalization and
pooling layers.} to that popularized by Krizhevsky et
al.\ \cite{Krizhevsky} and trained for classification using the 1.2 million
images of the ILSVRC 2012 challenge dataset \cite{ImageNet}.
All experiments are implemented using \caffe\ \cite{caffe}, and our network is
the publicly available \caffe\ reference model.
We use the activations of each layer as features, referred to as
\texttt{conv}$n$, \texttt{pool}$n$, or \texttt{fc}$n$ for the $n$th convolutional,
pooling, or fully connected layer, respectively.
We will use the term \emph{receptive field}, abbreviated rf, to refer to the
set of input pixels that are path-connected to a particular unit in the convnet.

\section{Feature visualization}

\begin{wraptable}{r}{0.5\textwidth}
    \centering
    \small
    \caption{Convnet receptive field sizes and strides, for an input of size
    $227 \times 227$. }
    \begin{tabular}{lll}
        layer & rf size & stride \\
        \hline
        \texttt{conv}1 & $11 \times 11$ & $4 \times 4$ \\
        \texttt{conv}2 & $51 \times 51$ & $8 \times 8$ \\
        \texttt{conv}3 & $99 \times 99$ & $16 \times 16$ \\
        \texttt{conv}4 & $131 \times 131$ & $16 \times 16$ \\
        \texttt{conv}5 & $163 \times 163$ & $16 \times 16$ \\
        \texttt{pool}5 & $195 \times 195$ & $32 \times 32$
    \end{tabular}
    \label{tab:rfs}
\end{wraptable}

\begin{figure}[t]
\centering
\renewcommand{\tabcolsep}{2pt}
\hspace{-2em}
\begin{minipage}[t]{0.23\linewidth}
\vspace{0pt}
\centering
\includegraphics[width=\textwidth]{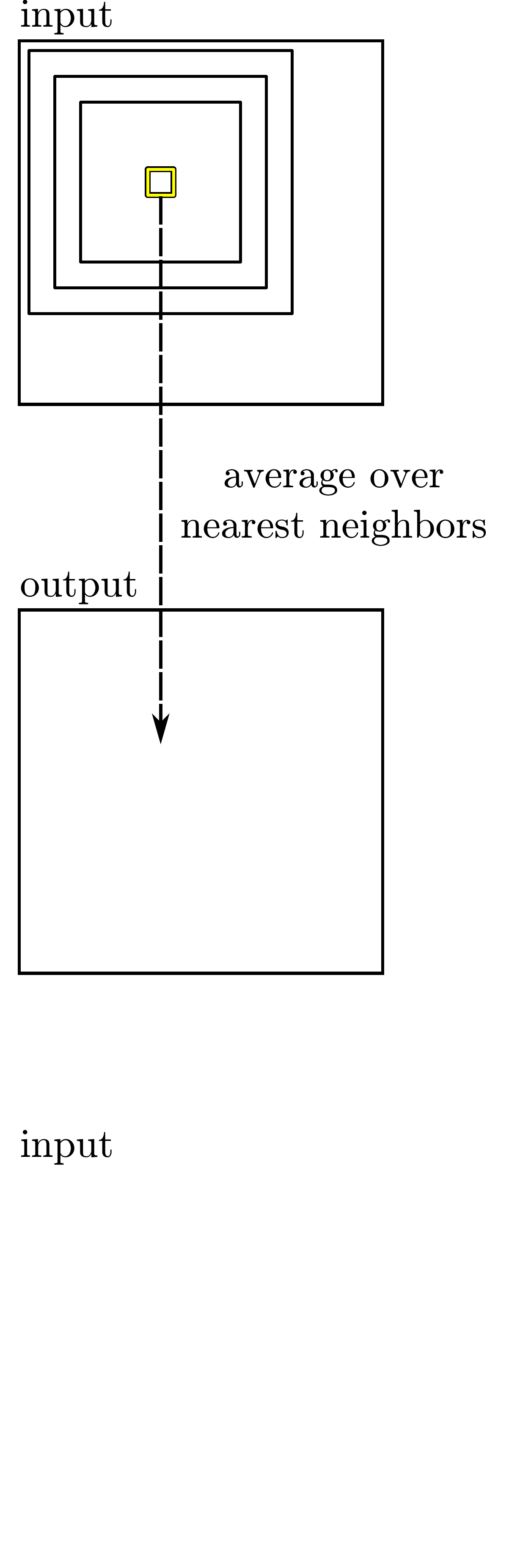}
\end{minipage}
%\hspace{0.5em}
\begin{minipage}[t]{0.65\linewidth}
\vspace{0pt}
\centering
\begin{tabular}{rccc|c}
& \texttt{conv}3 & \texttt{conv}4 & \texttt{conv}5 & uniform rf \\
\rotatebox{90}{1 neighbor} &
\includegraphics[width=0.25\textwidth]{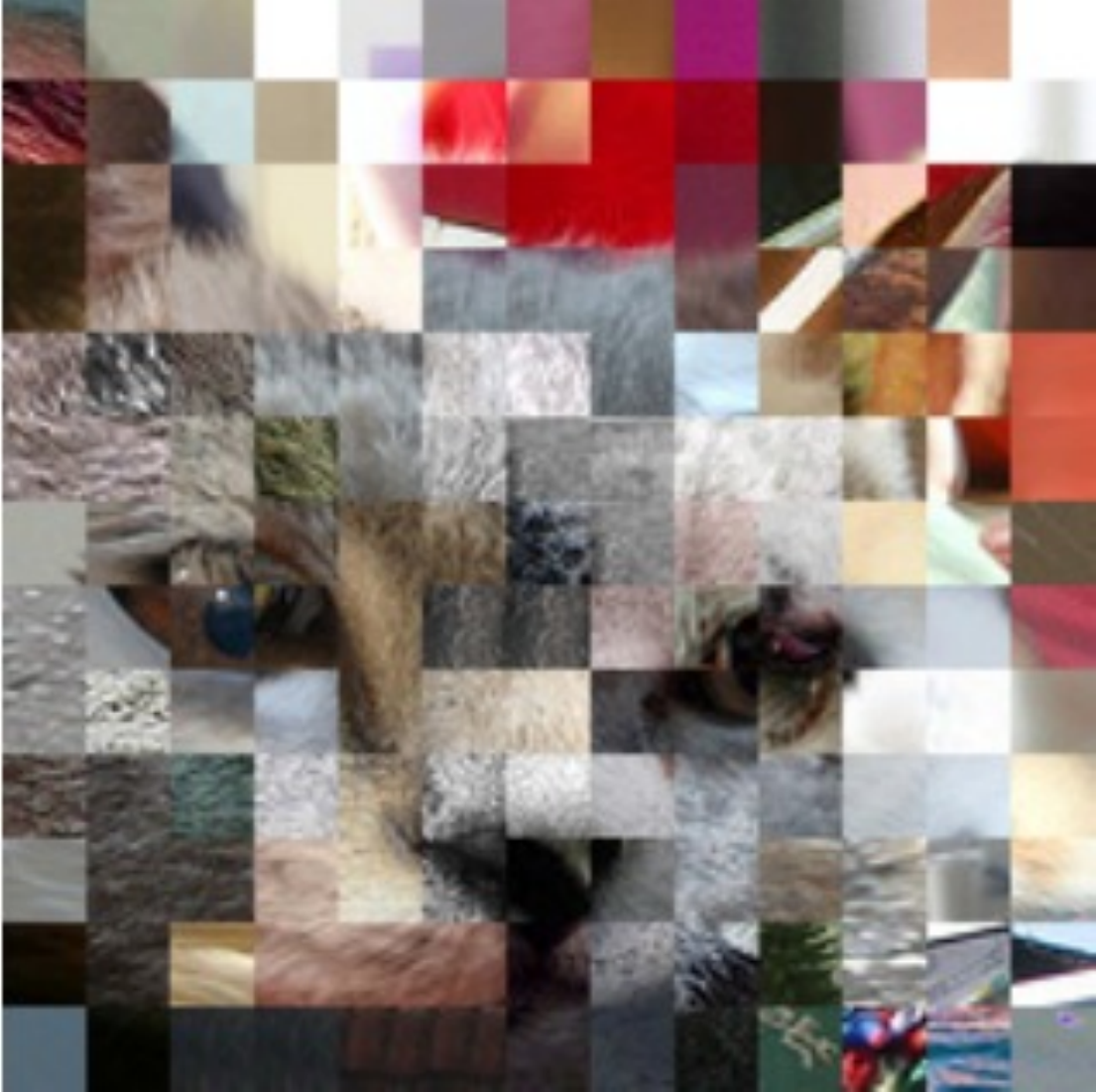} &
\includegraphics[width=0.25\textwidth]{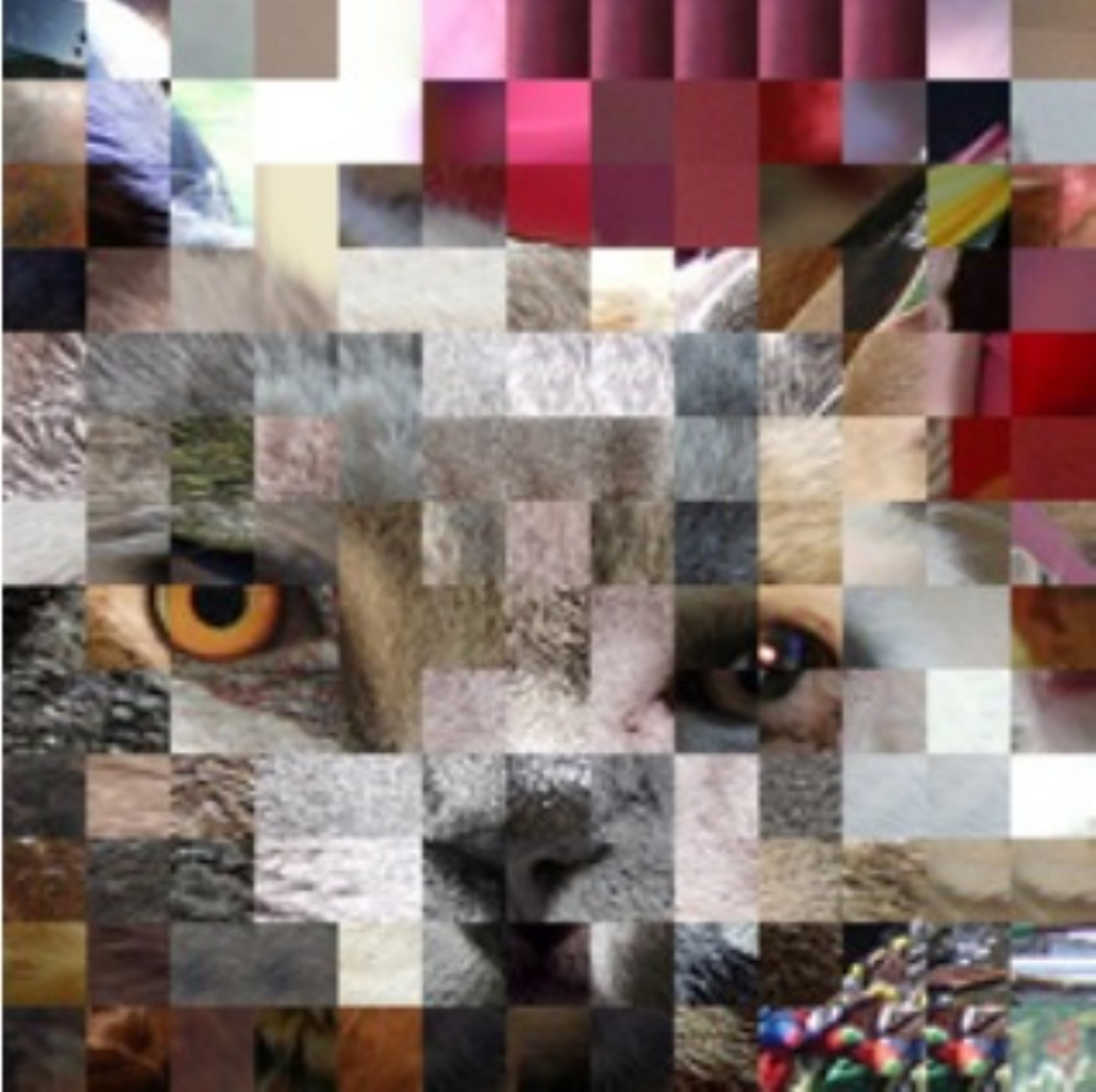} &
\includegraphics[width=0.25\textwidth]{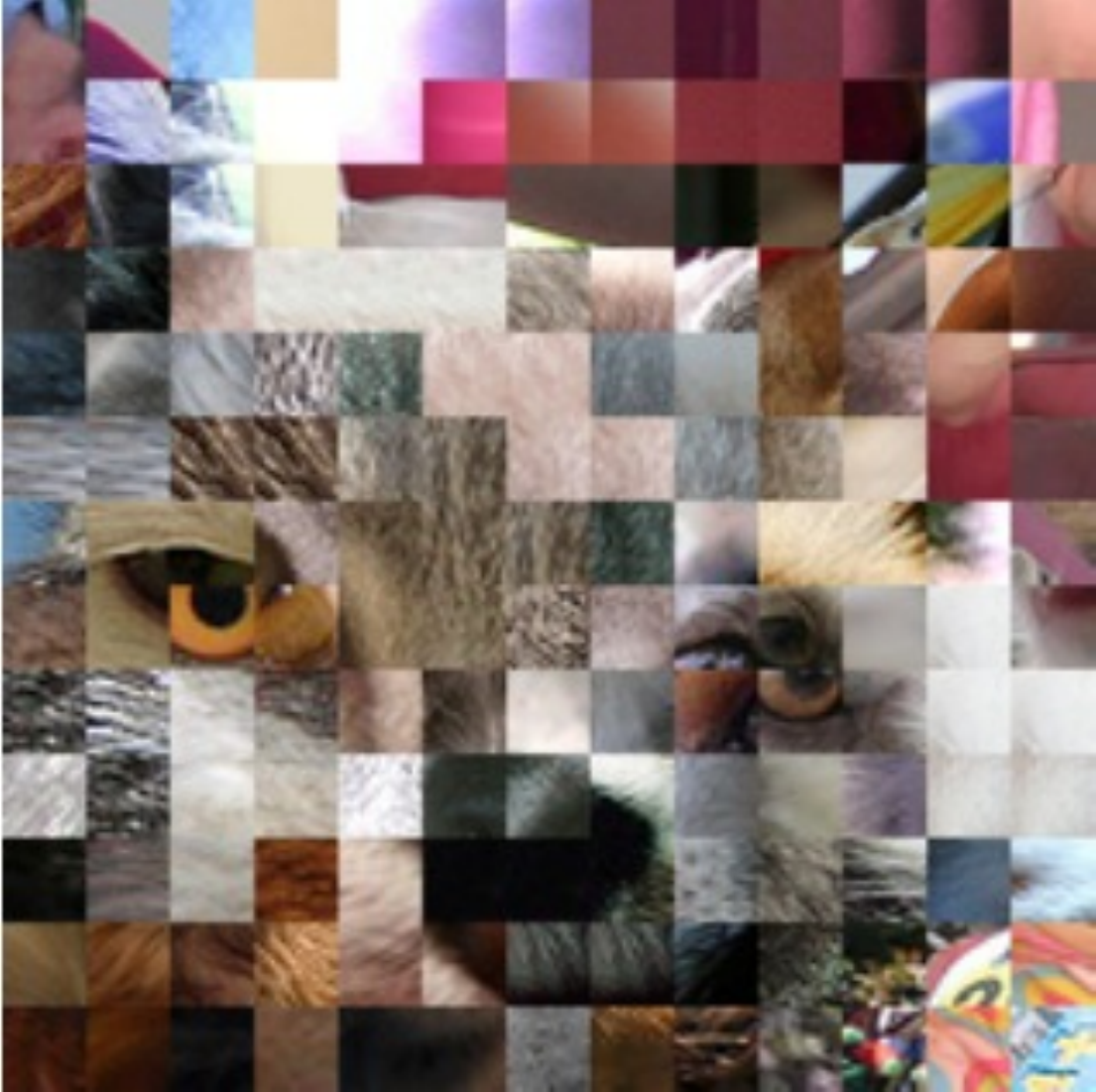} &
\includegraphics[width=0.25\textwidth]{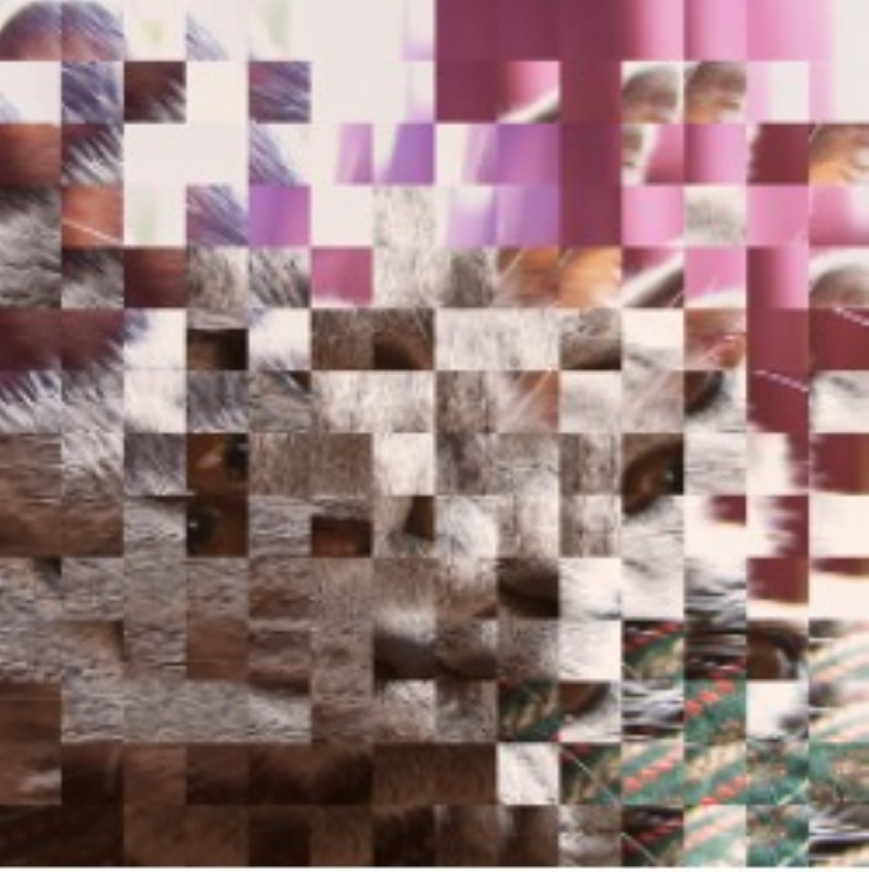} \\
\rotatebox{90}{5 neighbors} &
\includegraphics[width=0.25\textwidth]{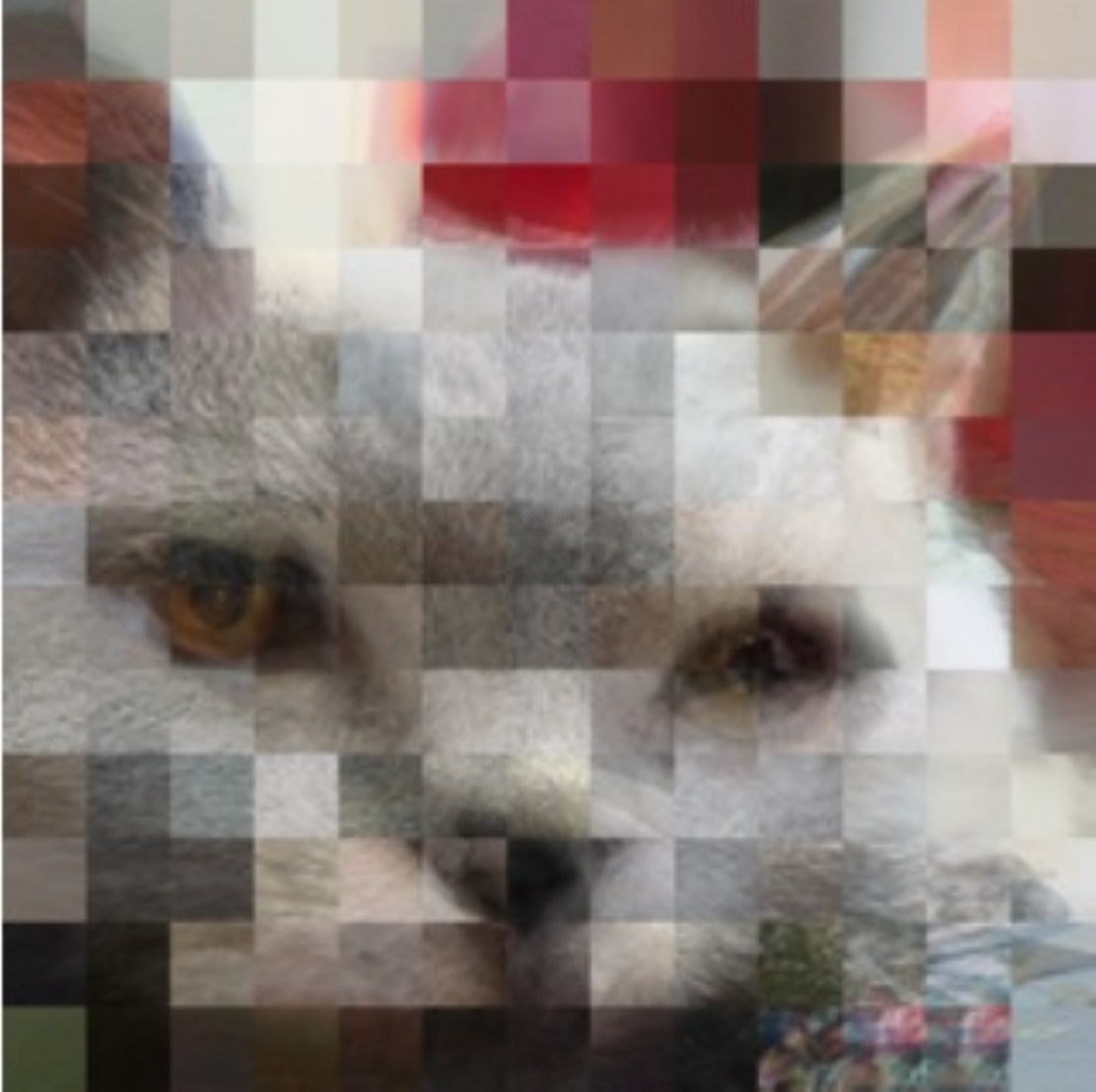} &
\includegraphics[width=0.25\textwidth]{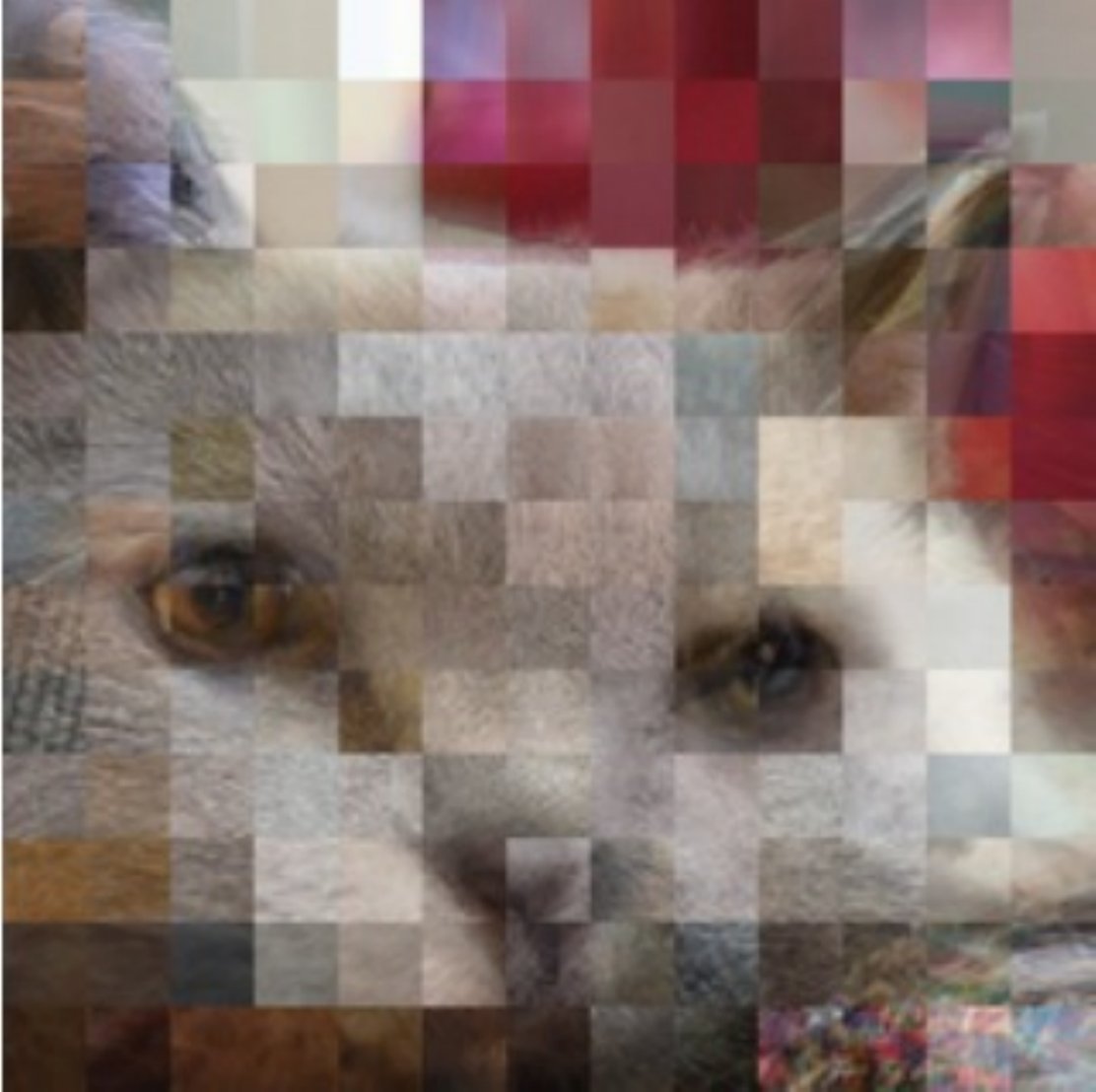} &
\includegraphics[width=0.25\textwidth]{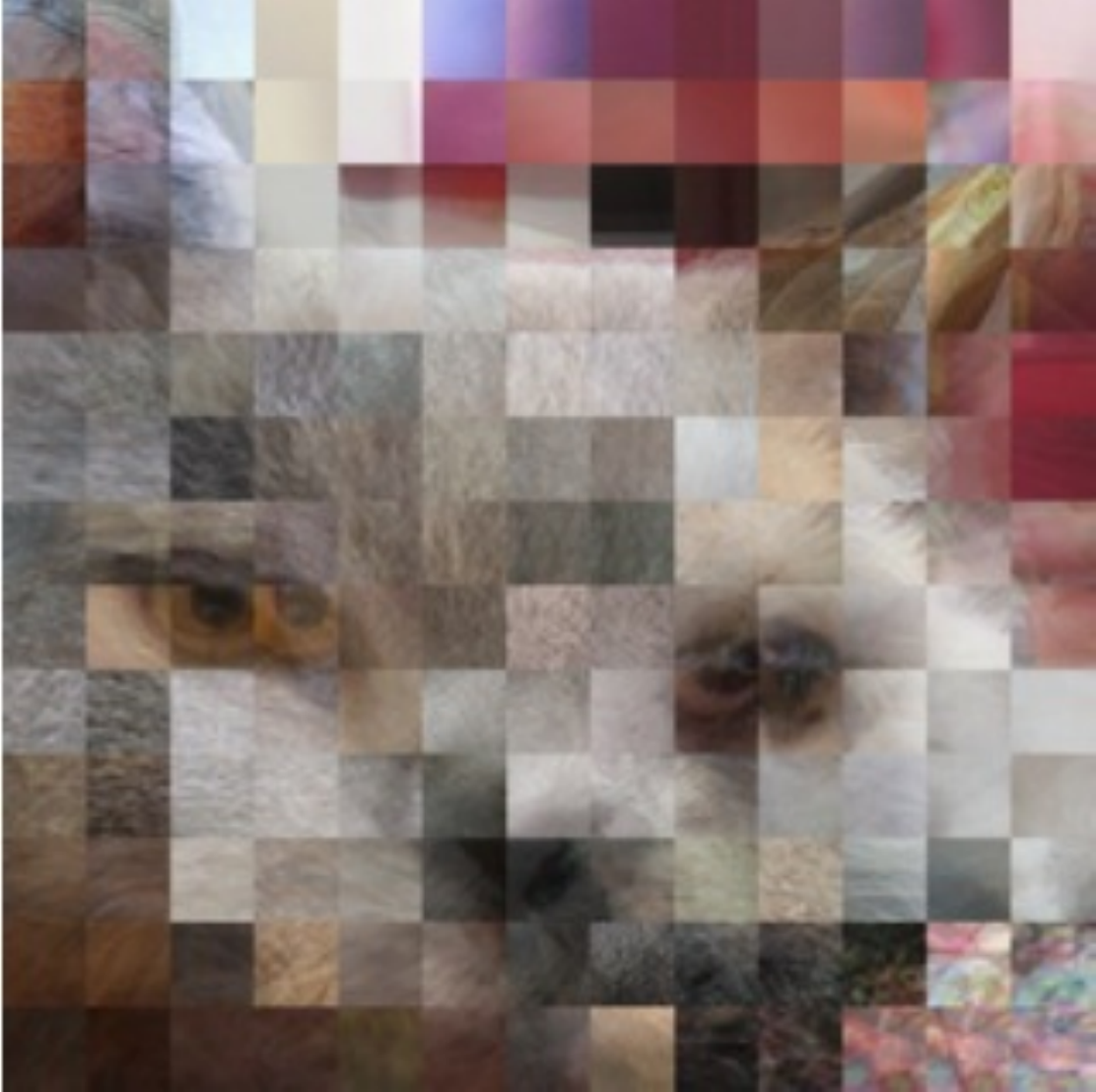} &
\includegraphics[width=0.25\textwidth]{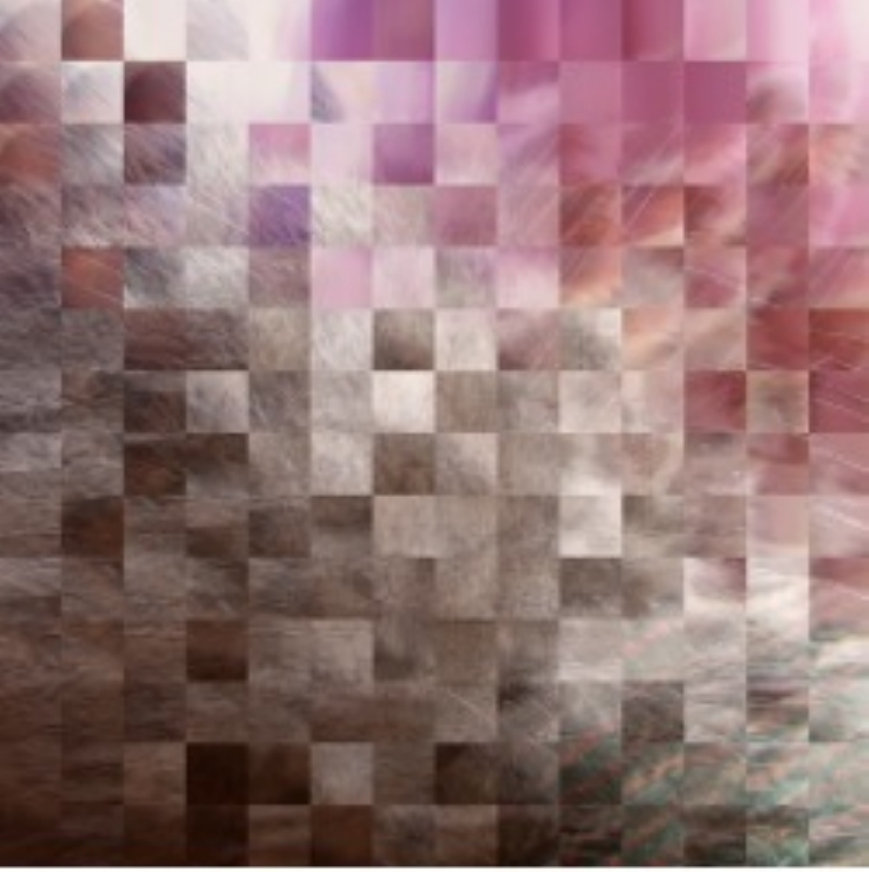} \\
\\
\rotatebox{90}{1 neighbor} &
\includegraphics[width=0.25\textwidth]{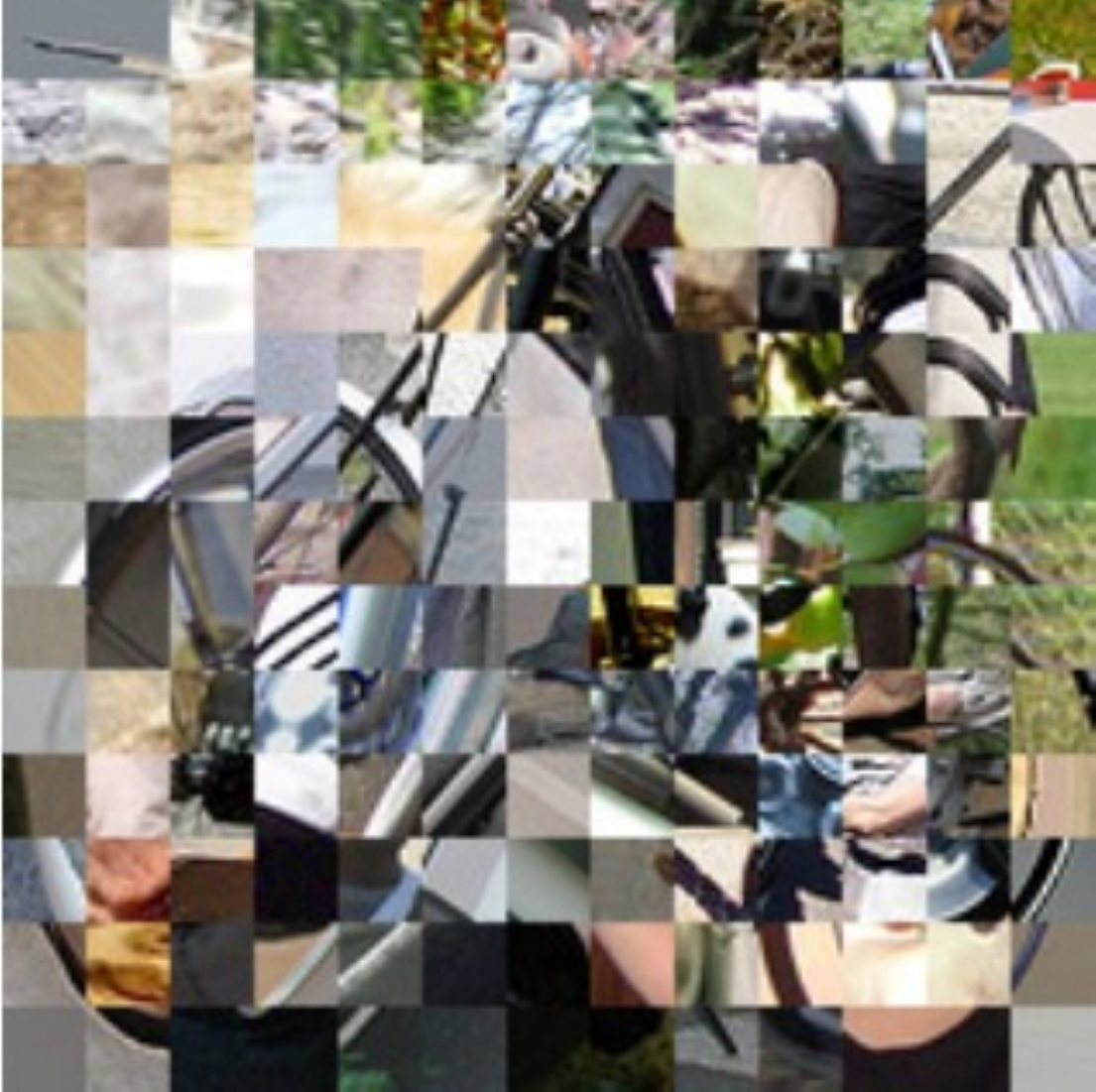} &
\includegraphics[width=0.25\textwidth]{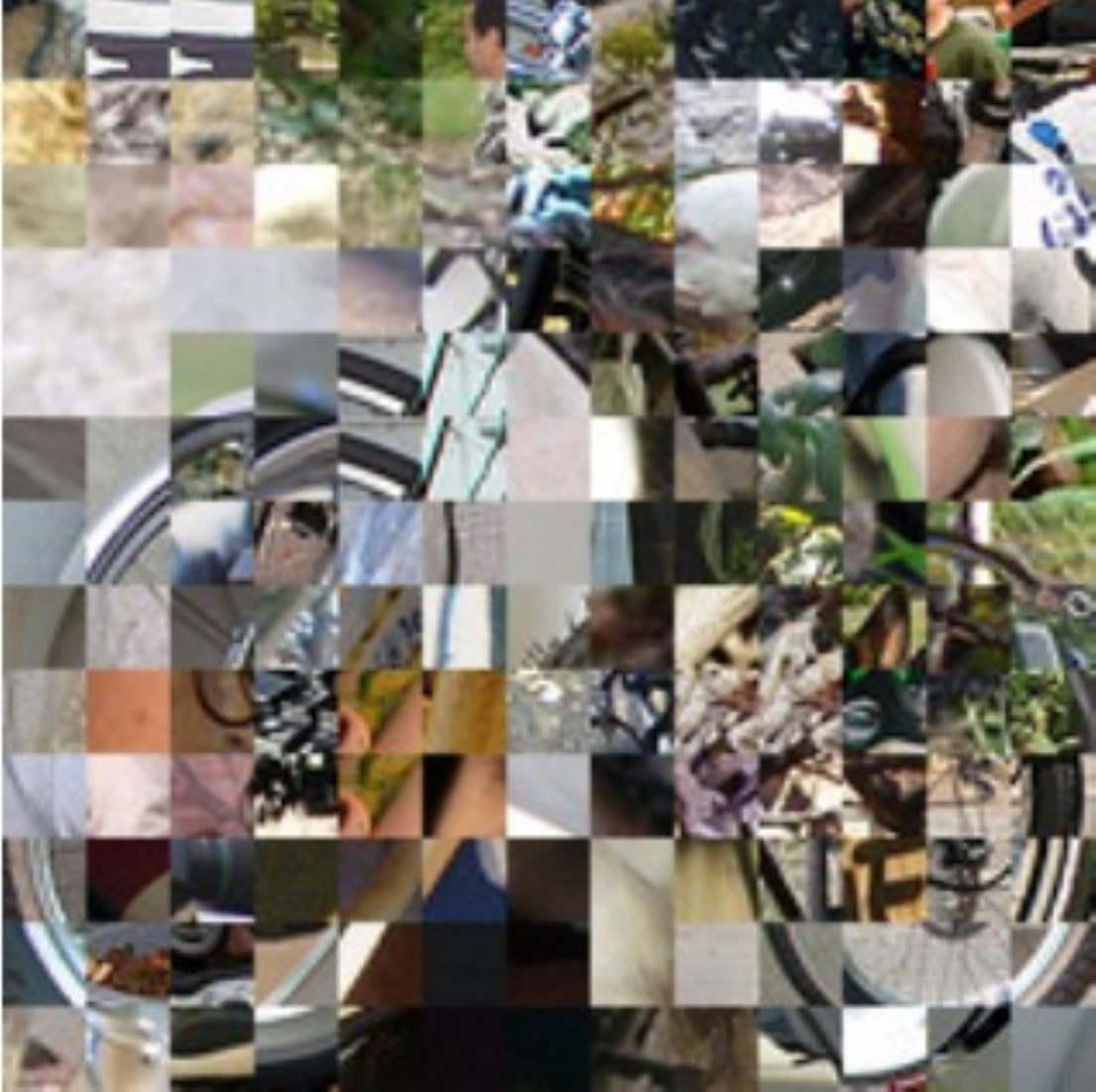} &
\includegraphics[width=0.25\textwidth]{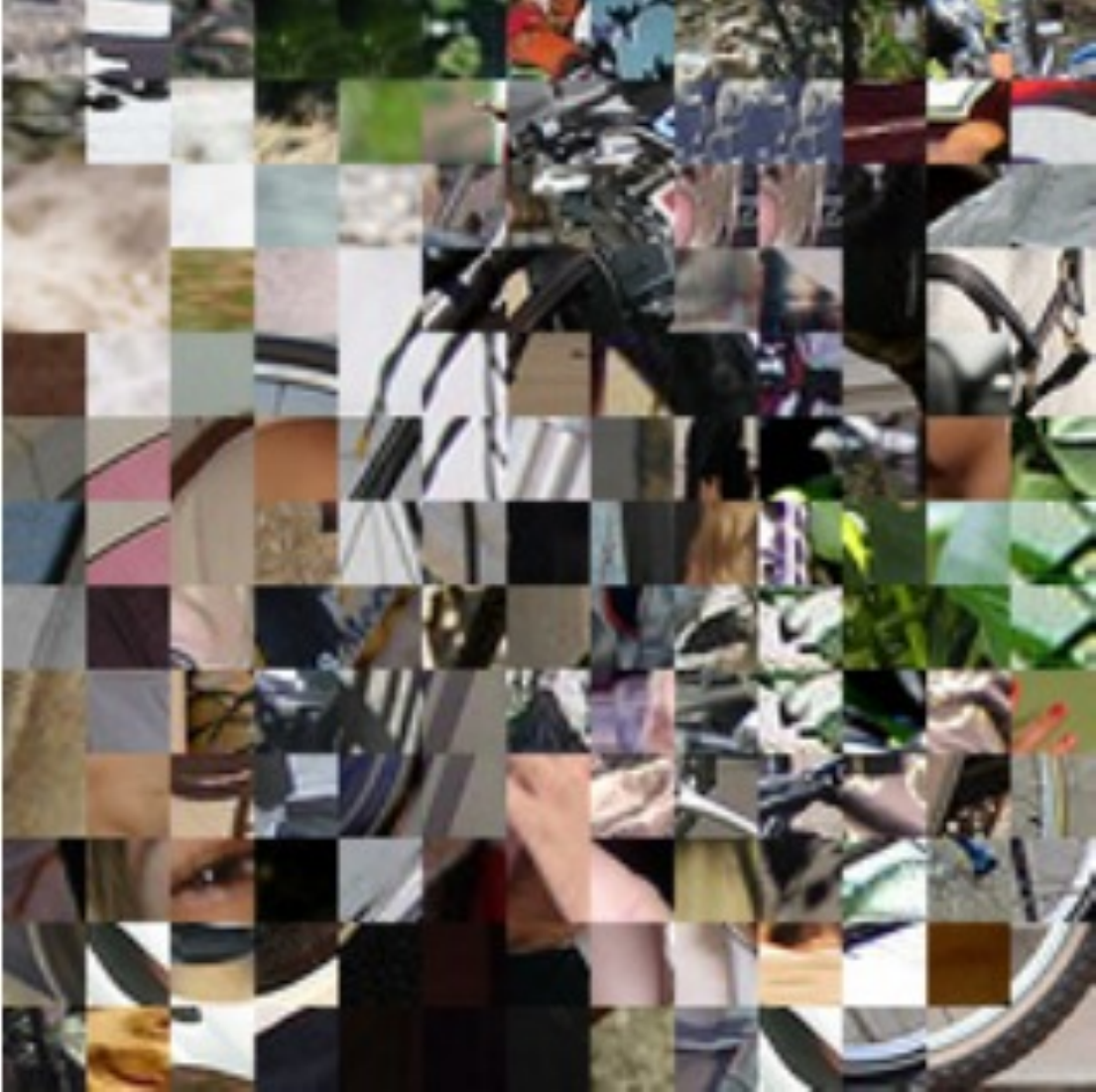} &
\includegraphics[width=0.25\textwidth]{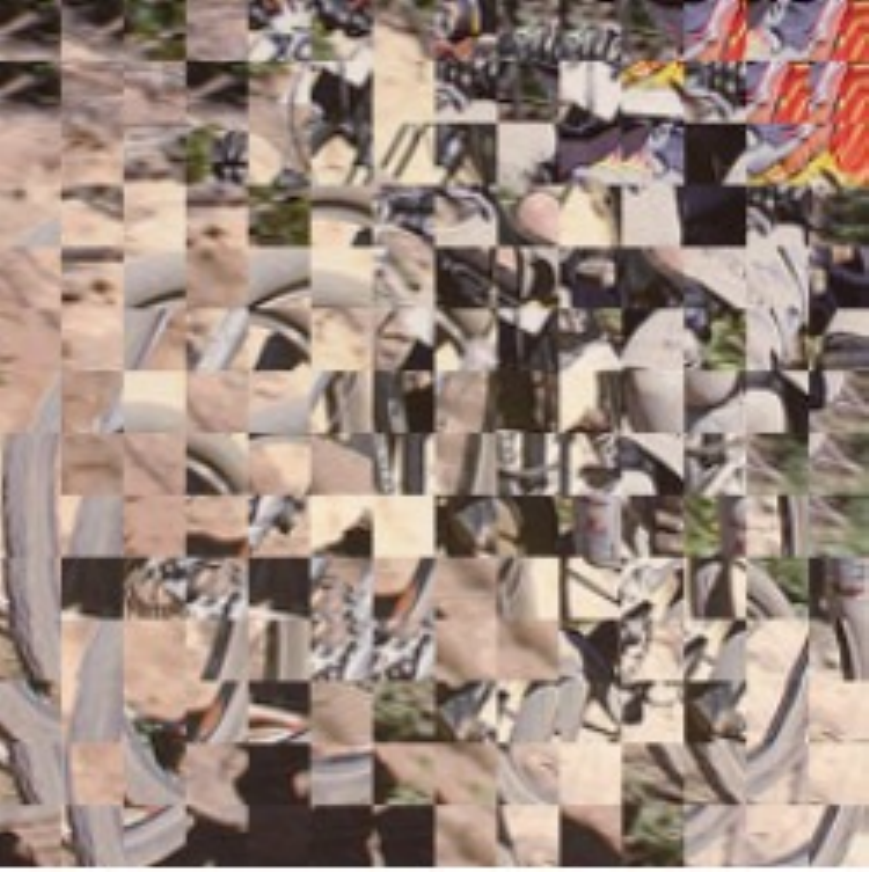} \\
\rotatebox{90}{5 neighbors} &
\includegraphics[width=0.25\textwidth]{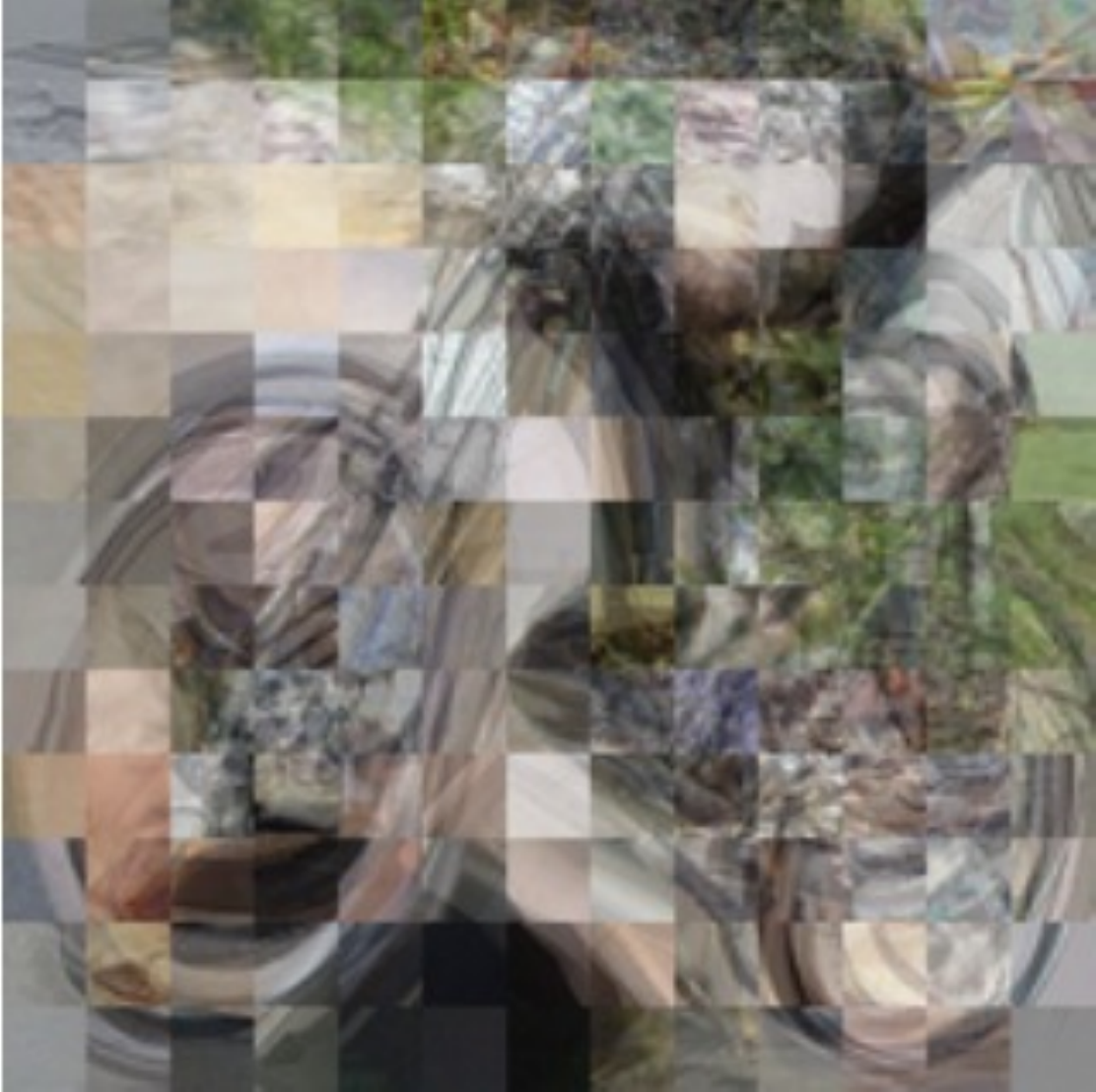} &
\includegraphics[width=0.25\textwidth]{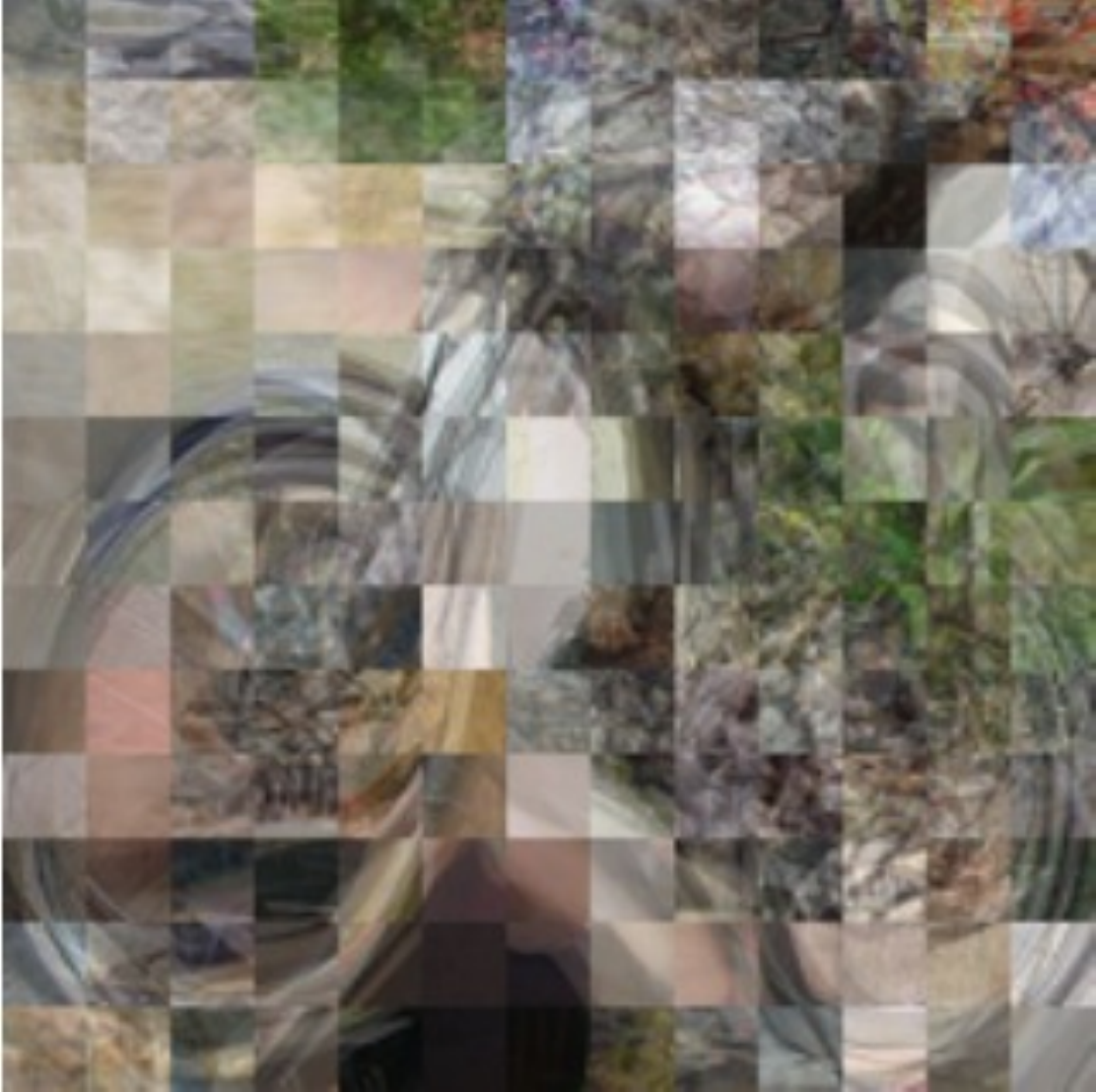} &
\includegraphics[width=0.25\textwidth]{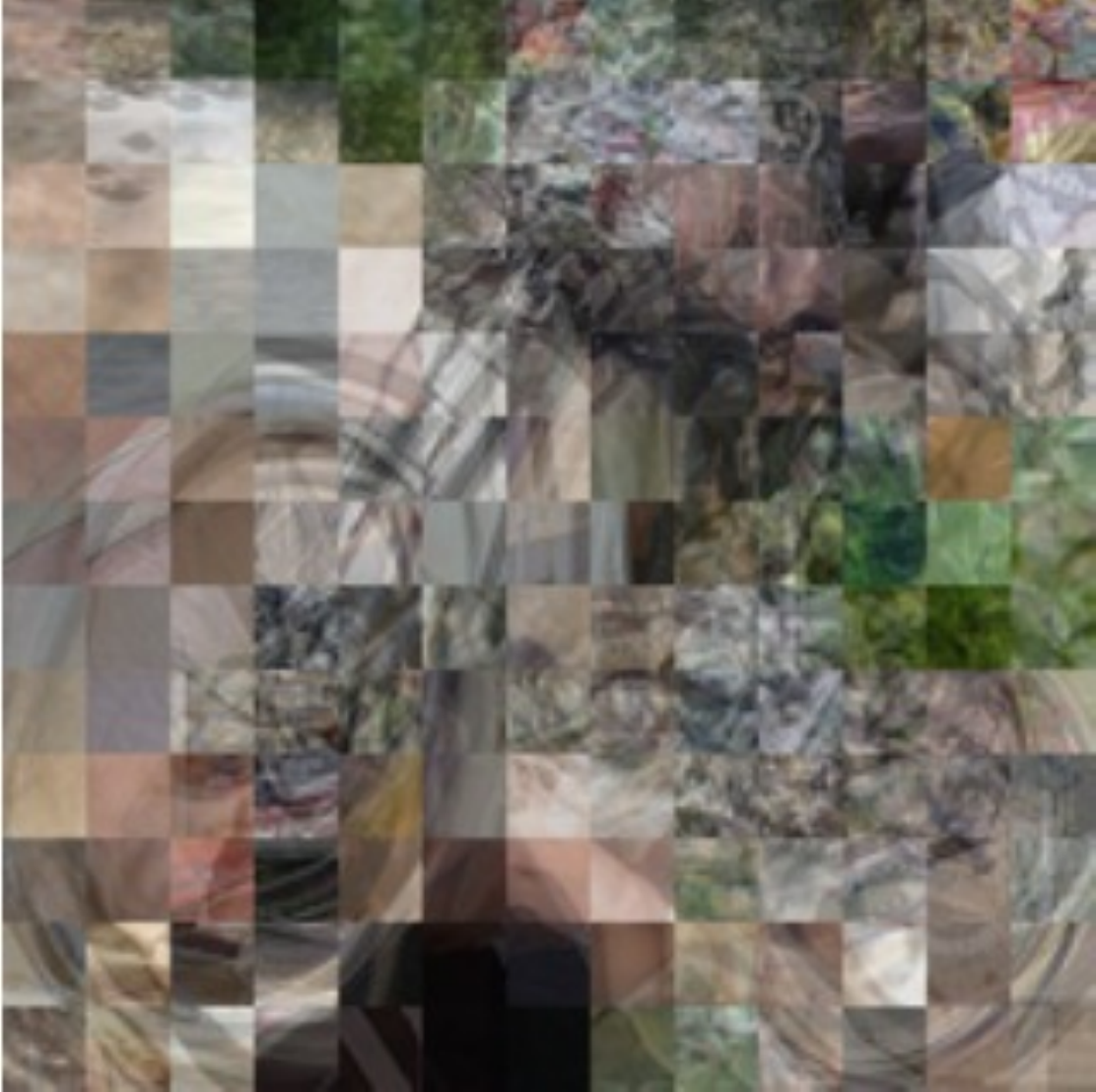} &
\includegraphics[width=0.25\textwidth]{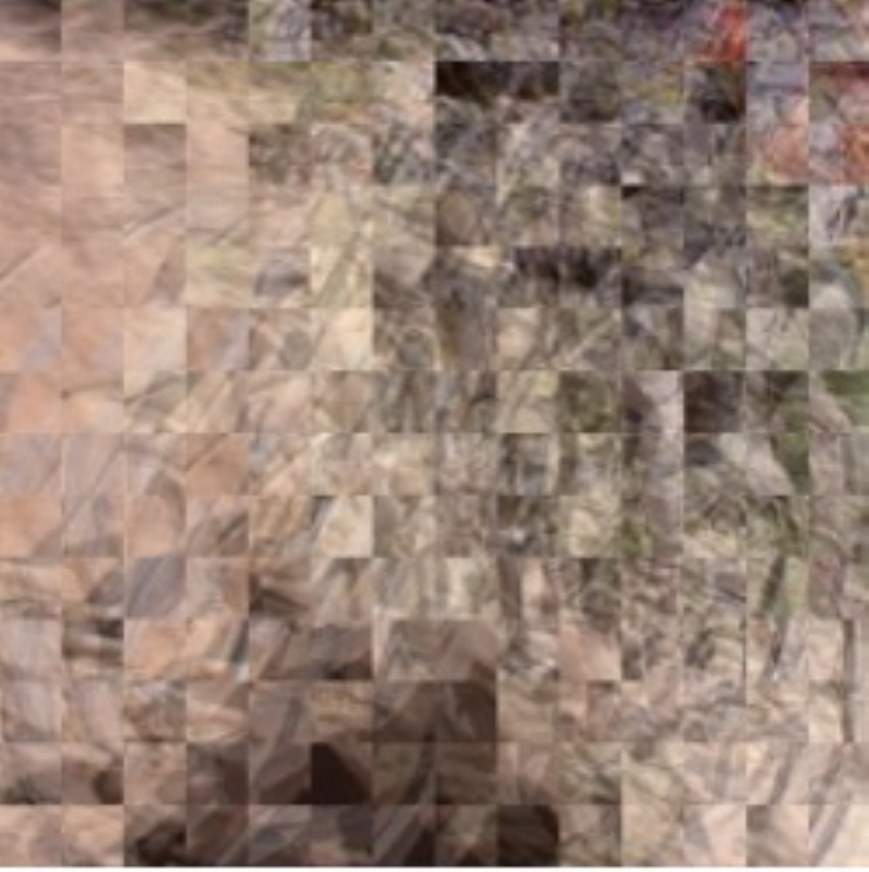}
\end{tabular}
\end{minipage}

\caption{
Even though they have large receptive fields, convnet features carry
local information at a finer scale.
Upper left: given an input image, we replaced $16 \times 16$ patches with
averages over 1 or 5 nearest neighbor patches, computed using convnet features
centered at those patches.
The yellow square illustrates one input patch, and the black squares show the
corresponding rfs for the three layers shown.
Right: Notice that the features retrieve reasonable matches for the centers of
their receptive fields, even though those rfs extend over large regions of the
source image.
In the ``uniform rf'' column, we show the best that could be expected if convnet
features discarded all spatial information within their rfs, by choosing input
patches uniformly at random from \texttt{conv}3-sized neighborhoods.
(Best viewed electronically.)
}
\label{fig:patches}
\end{figure}

\begin{figure}
\centering
\renewcommand{\tabcolsep}{1pt}
\renewcommand{\arraystretch}{0.7}
\scalebox{0.9}{
\begin{tabular}{rcccc@{\hskip 5pt}cccc@{\hskip 5pt}cccc}
& \multicolumn{4}{c}{\texttt{conv}3}
& \multicolumn{4}{c}{\texttt{conv}4}
& \multicolumn{4}{c}{\texttt{conv}5} \\
\rotatebox{90}{\small 5 nbrs} &
\includegraphics[width=0.08\textwidth]{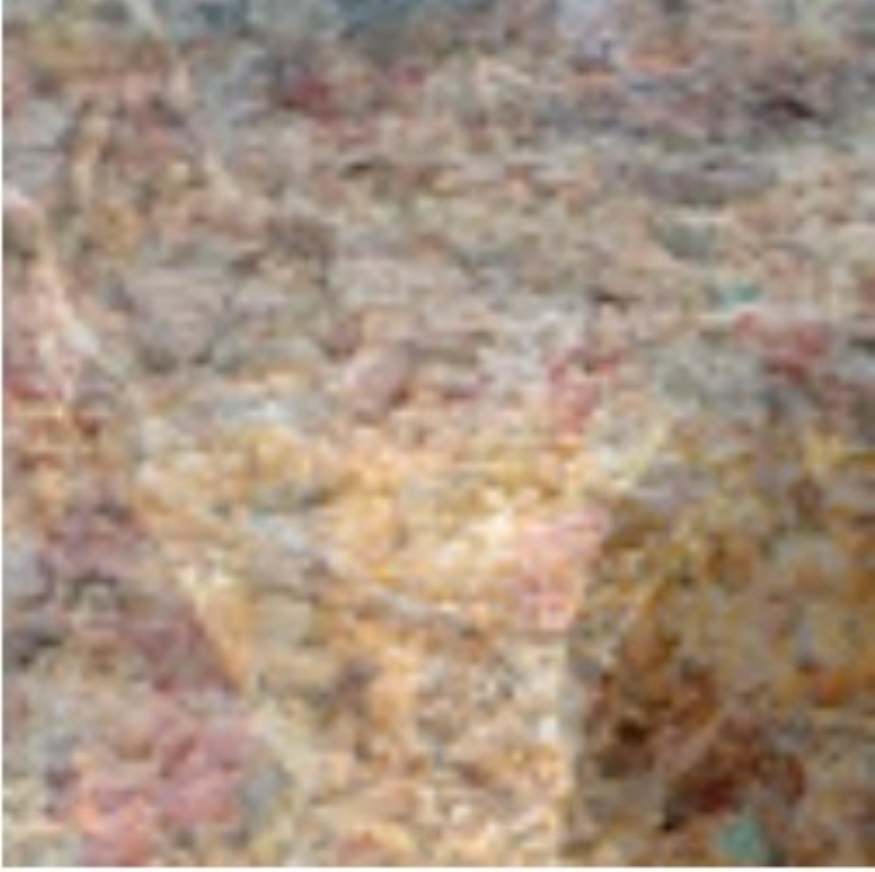} &
\includegraphics[width=0.08\textwidth]{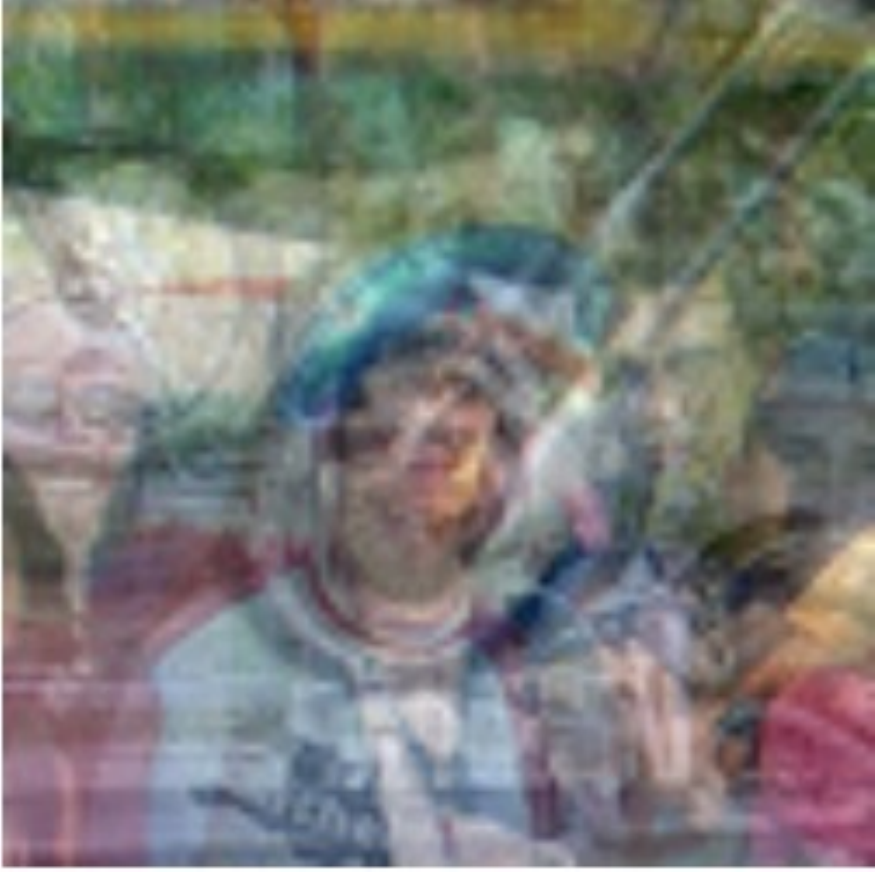} &
\includegraphics[width=0.08\textwidth]{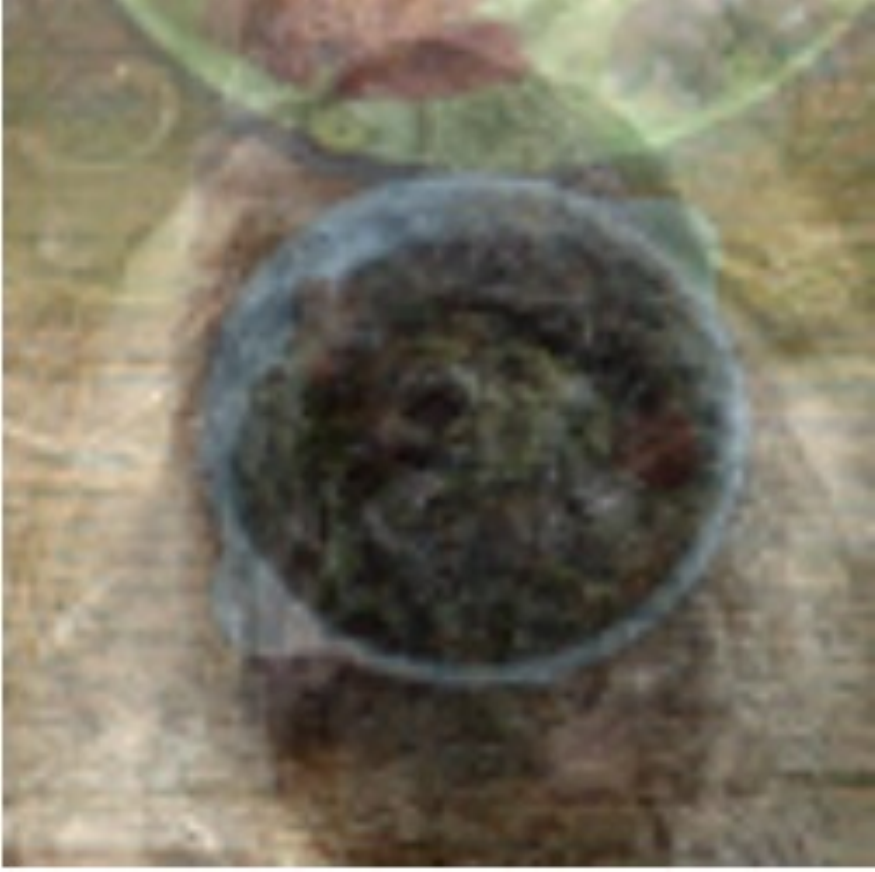} &
\includegraphics[width=0.08\textwidth]{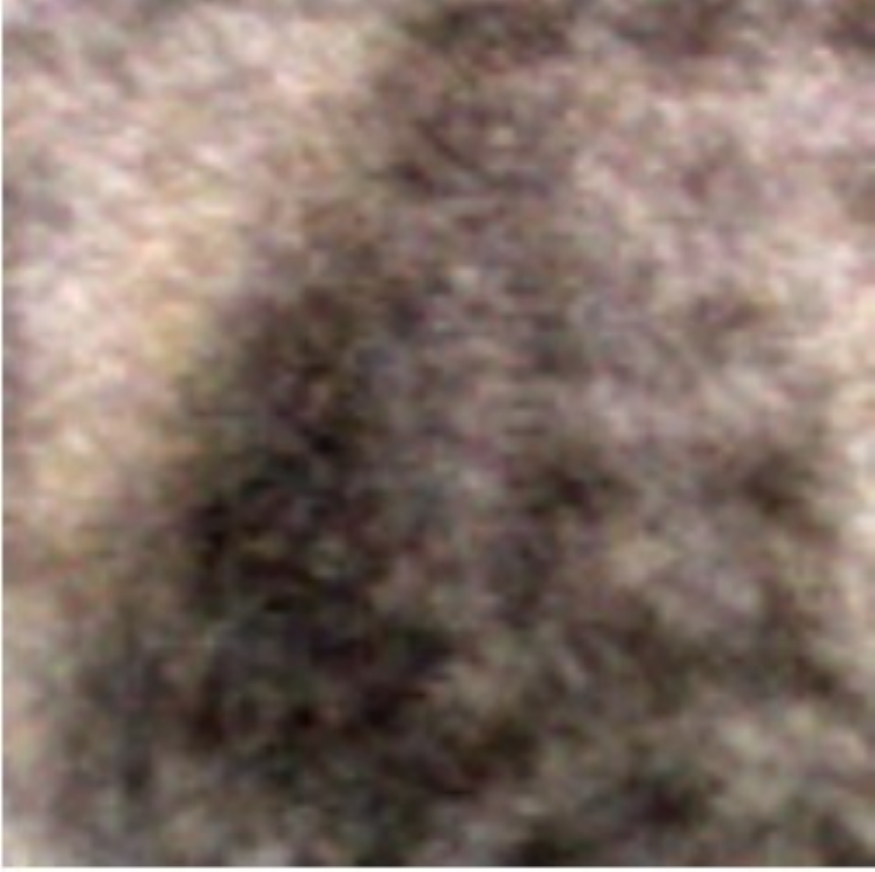} &
\includegraphics[width=0.08\textwidth]{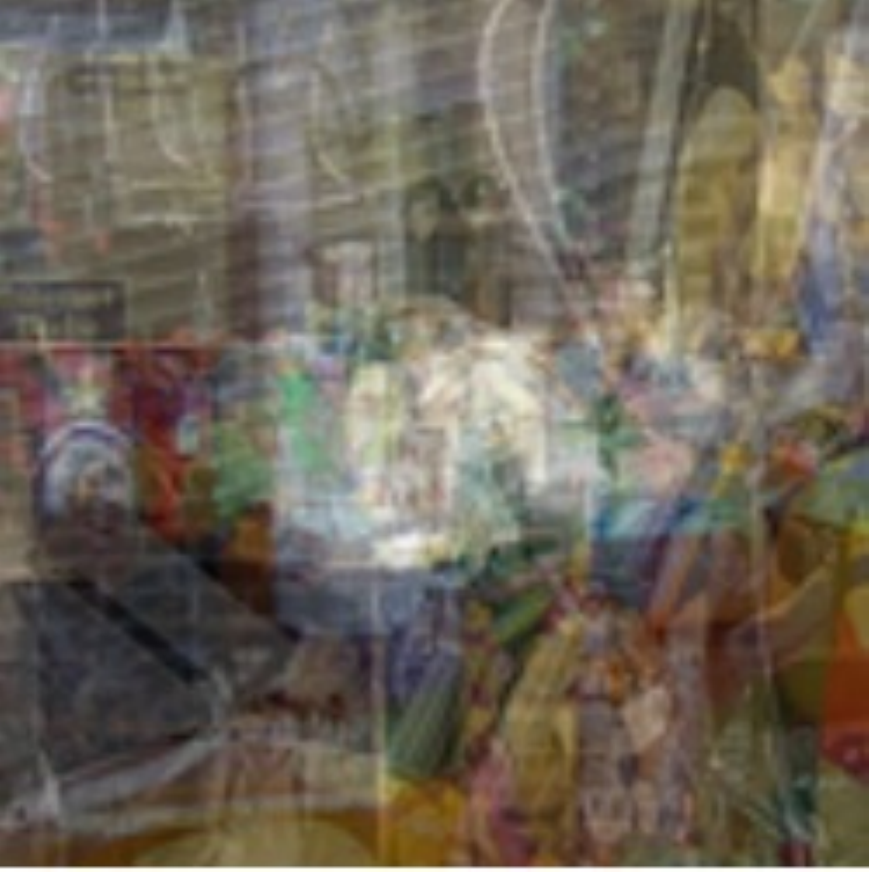} &
\includegraphics[width=0.08\textwidth]{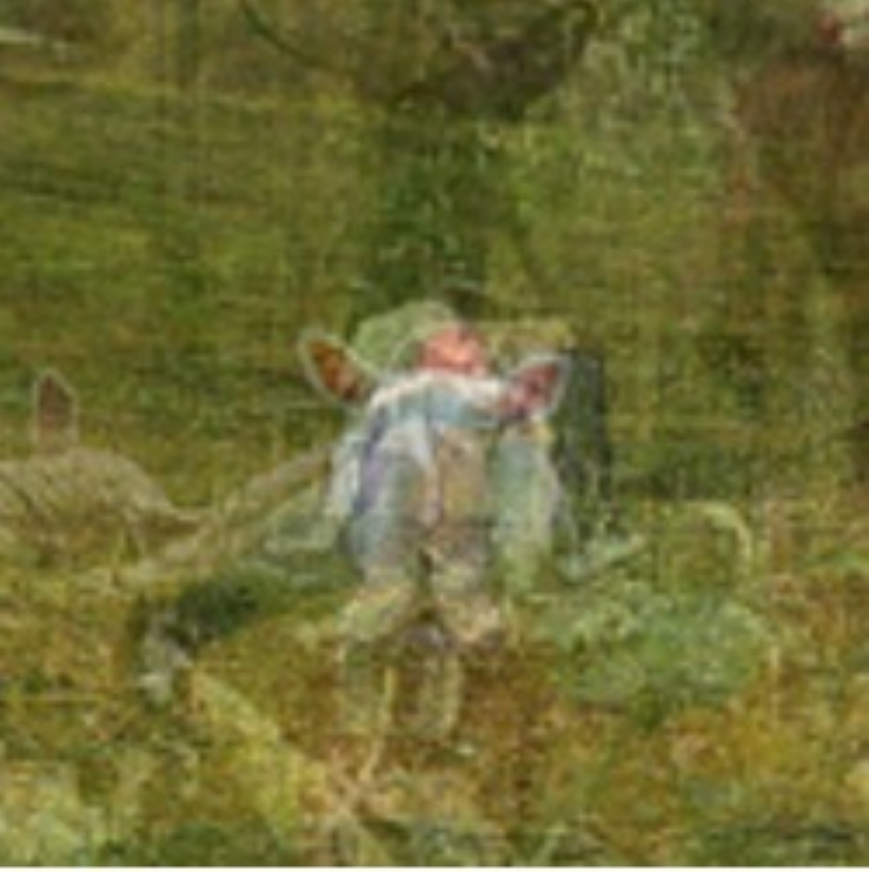} &
\includegraphics[width=0.08\textwidth]{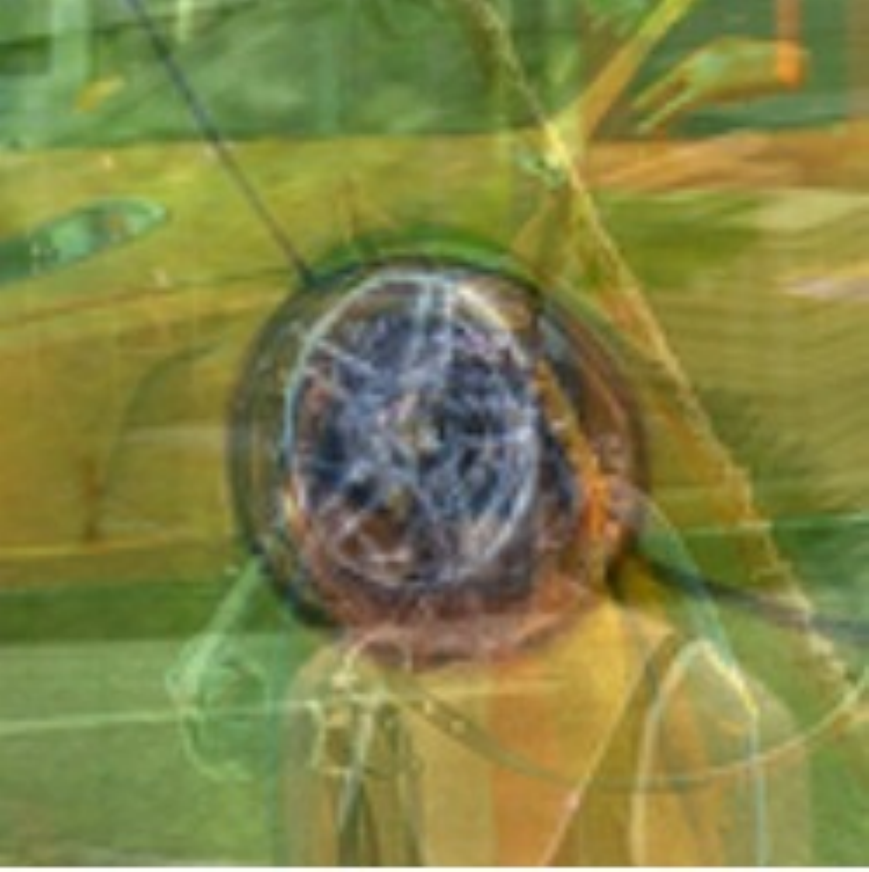} &
\includegraphics[width=0.08\textwidth]{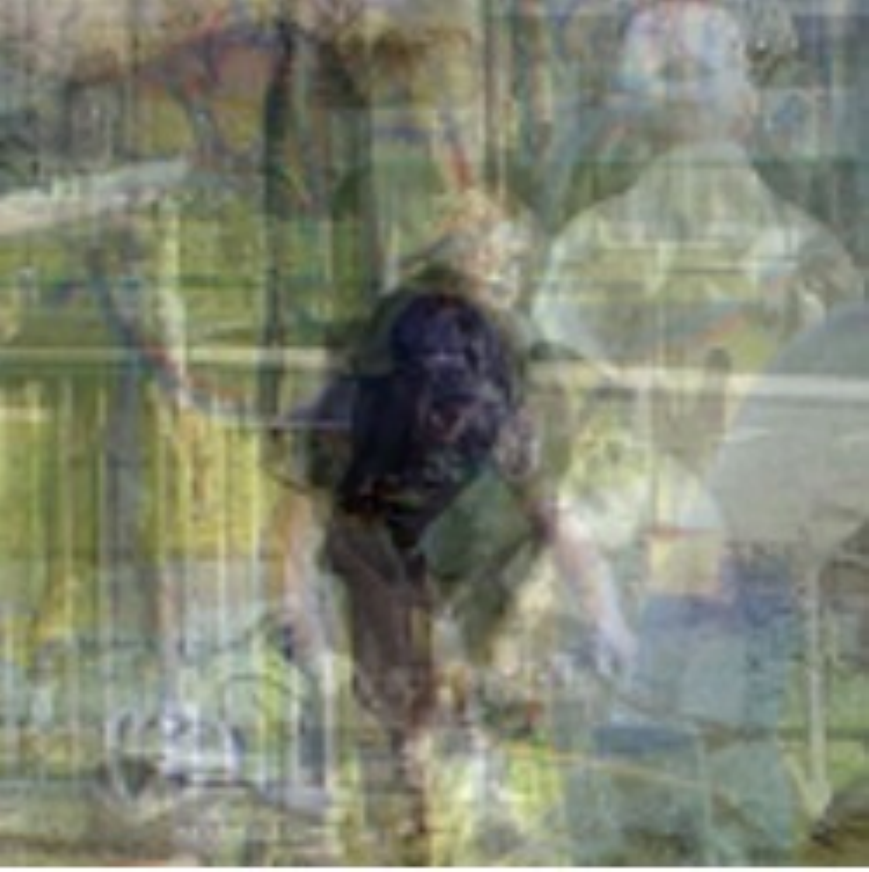} &
\includegraphics[width=0.08\textwidth]{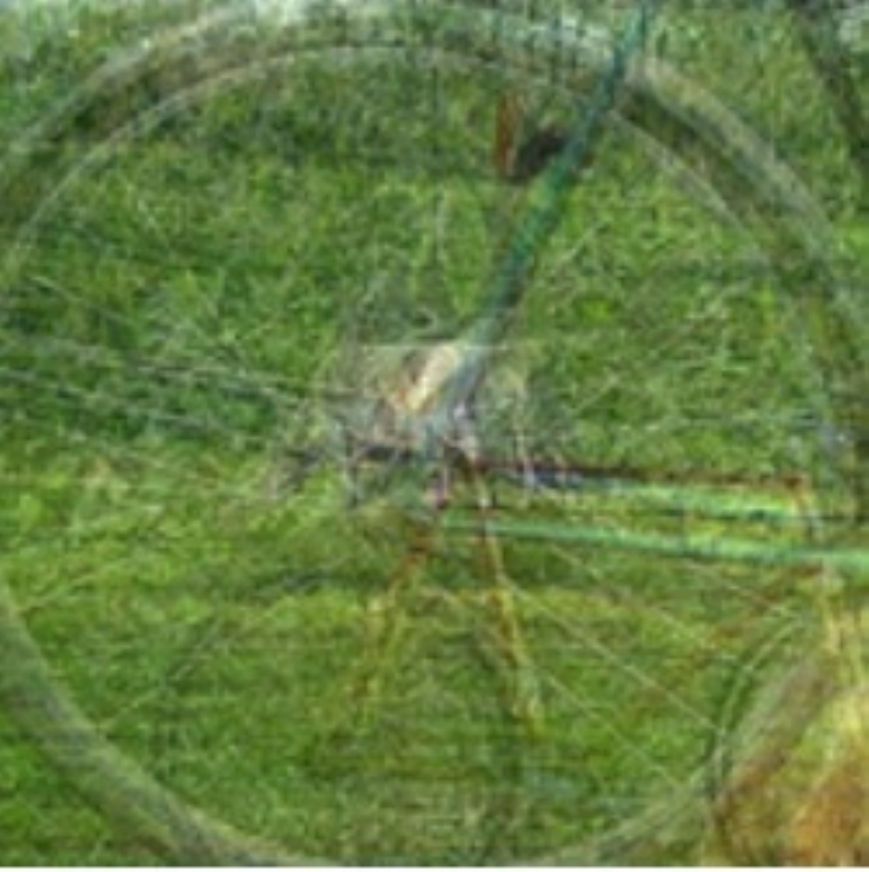} &
\includegraphics[width=0.08\textwidth]{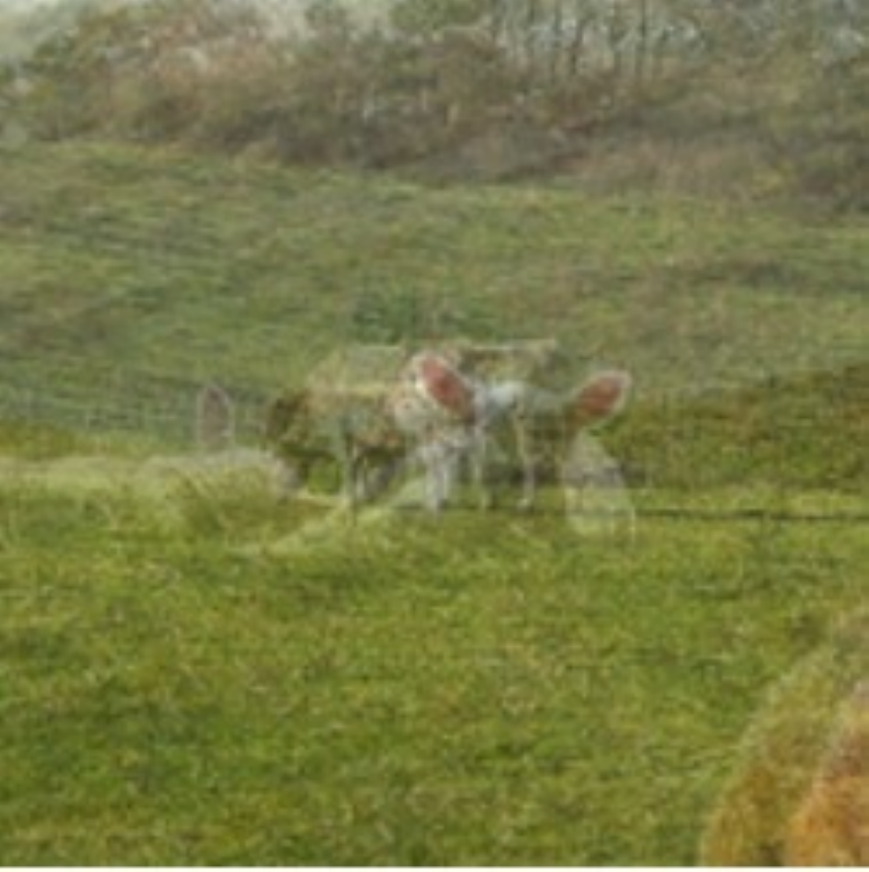} &
\includegraphics[width=0.08\textwidth]{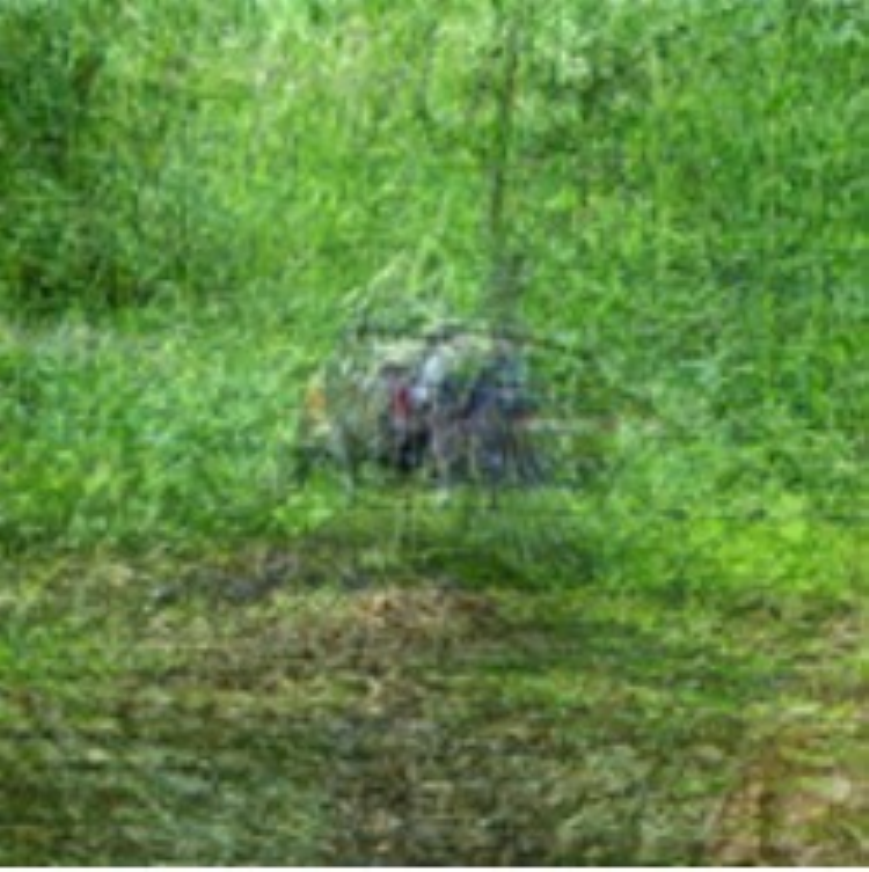} &
\includegraphics[width=0.08\textwidth]{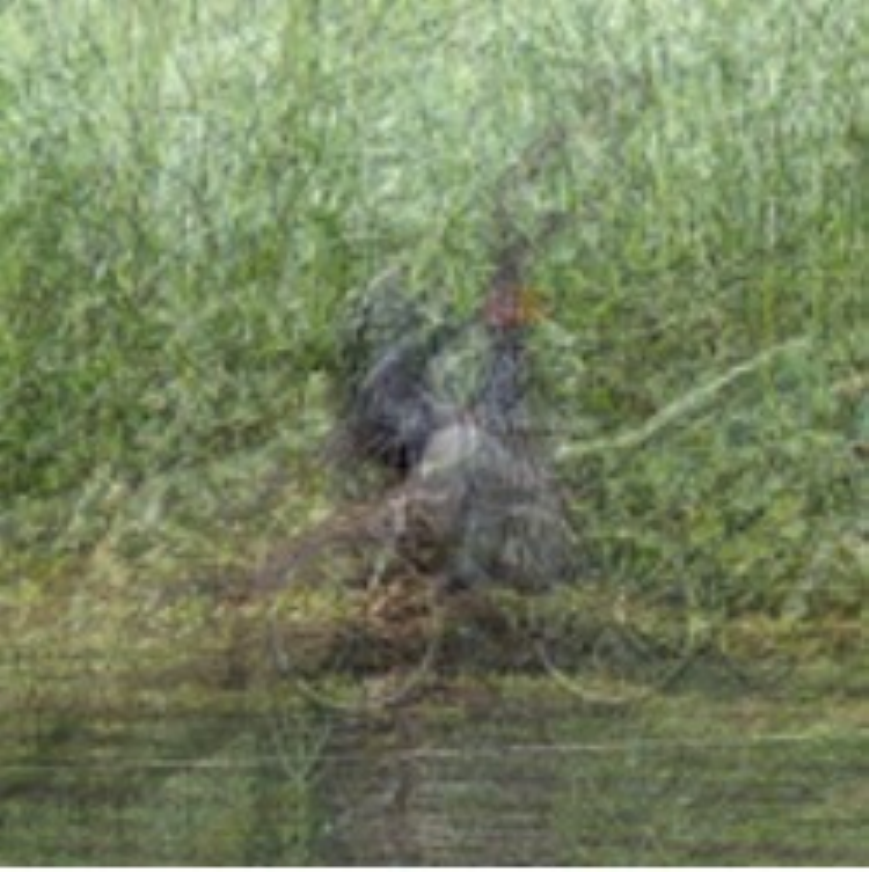} \\
\rotatebox{90}{\small 50 nbrs} &
\includegraphics[width=0.08\textwidth]{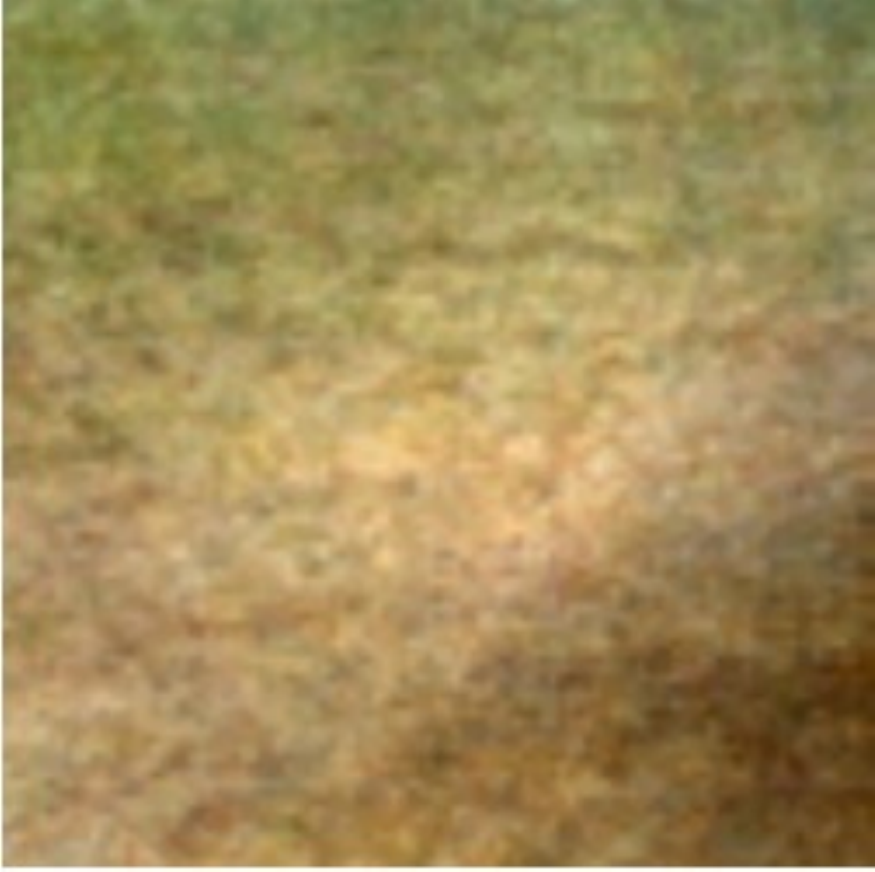} &
\includegraphics[width=0.08\textwidth]{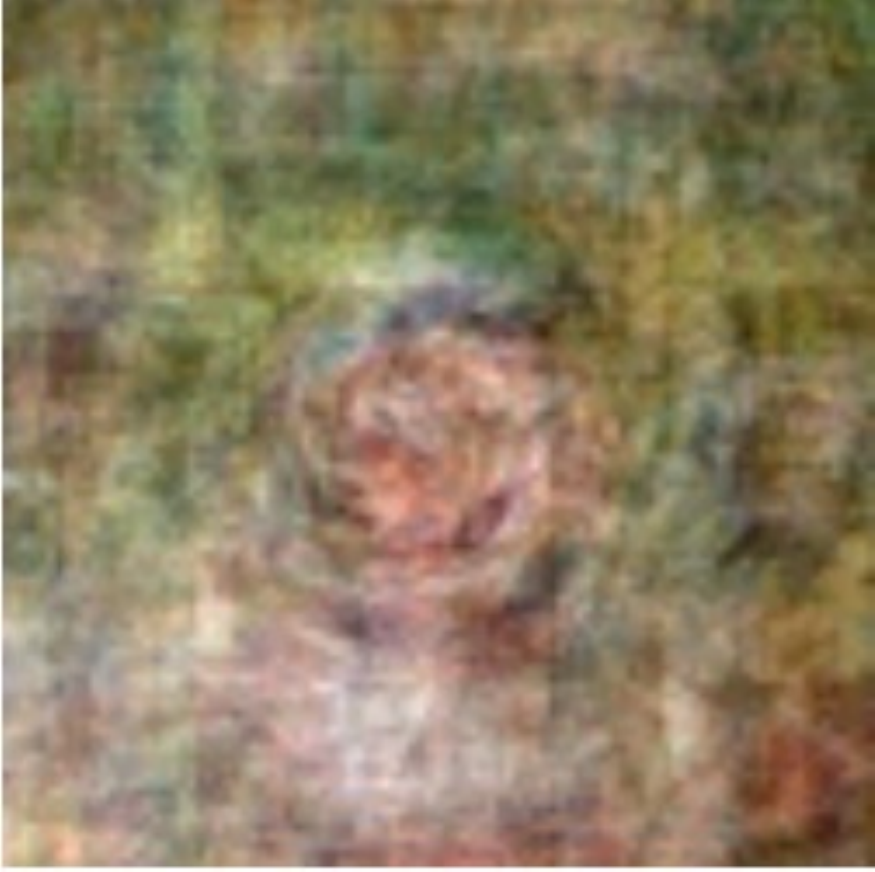} &
\includegraphics[width=0.08\textwidth]{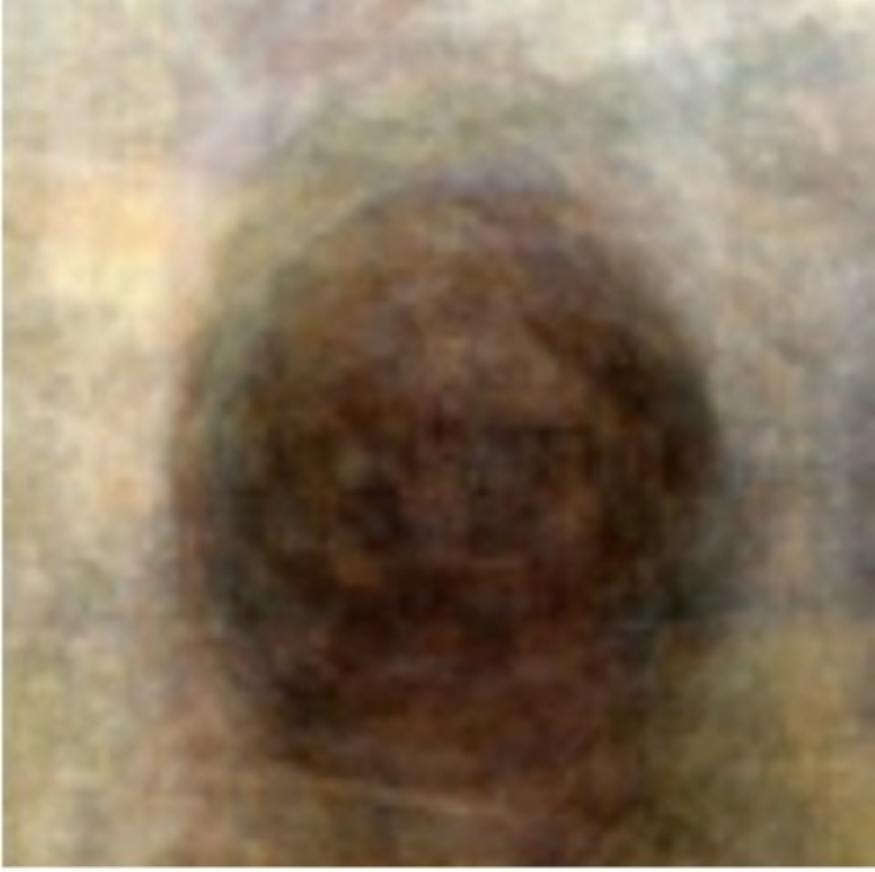} &
\includegraphics[width=0.08\textwidth]{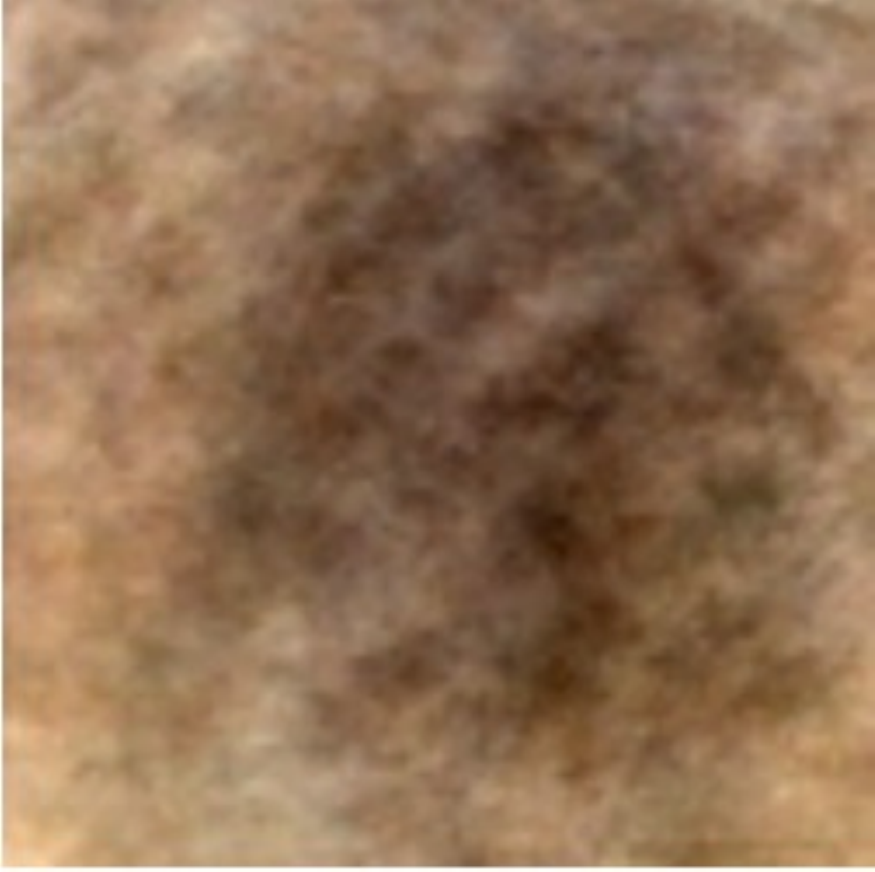} &
\includegraphics[width=0.08\textwidth]{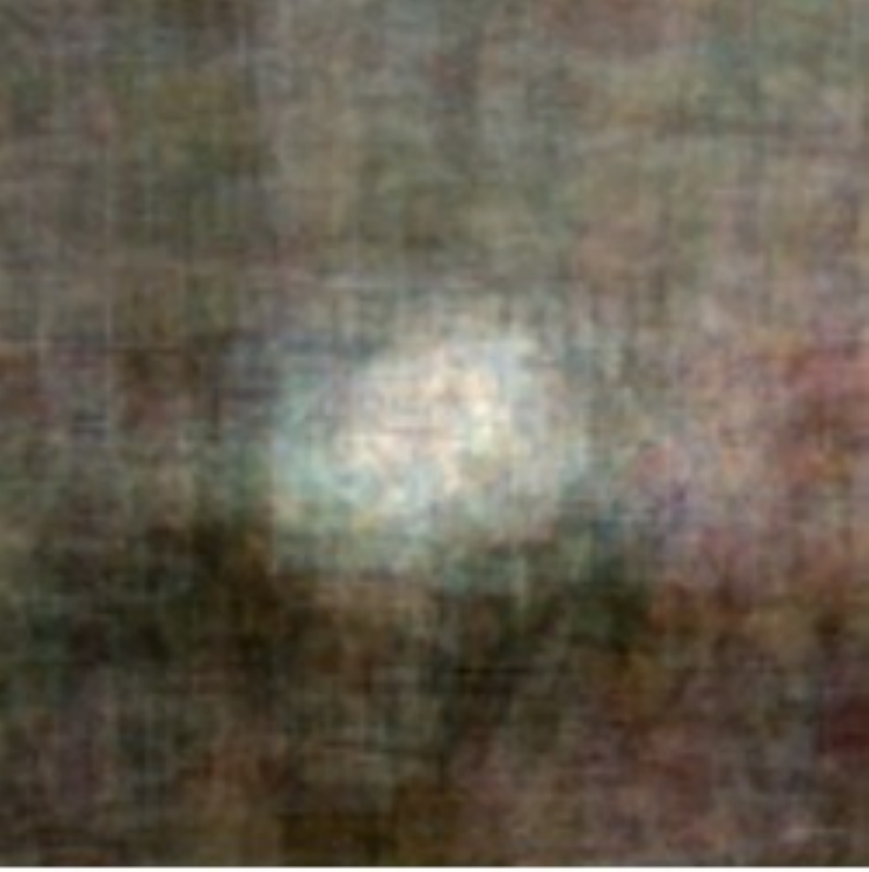} &
\includegraphics[width=0.08\textwidth]{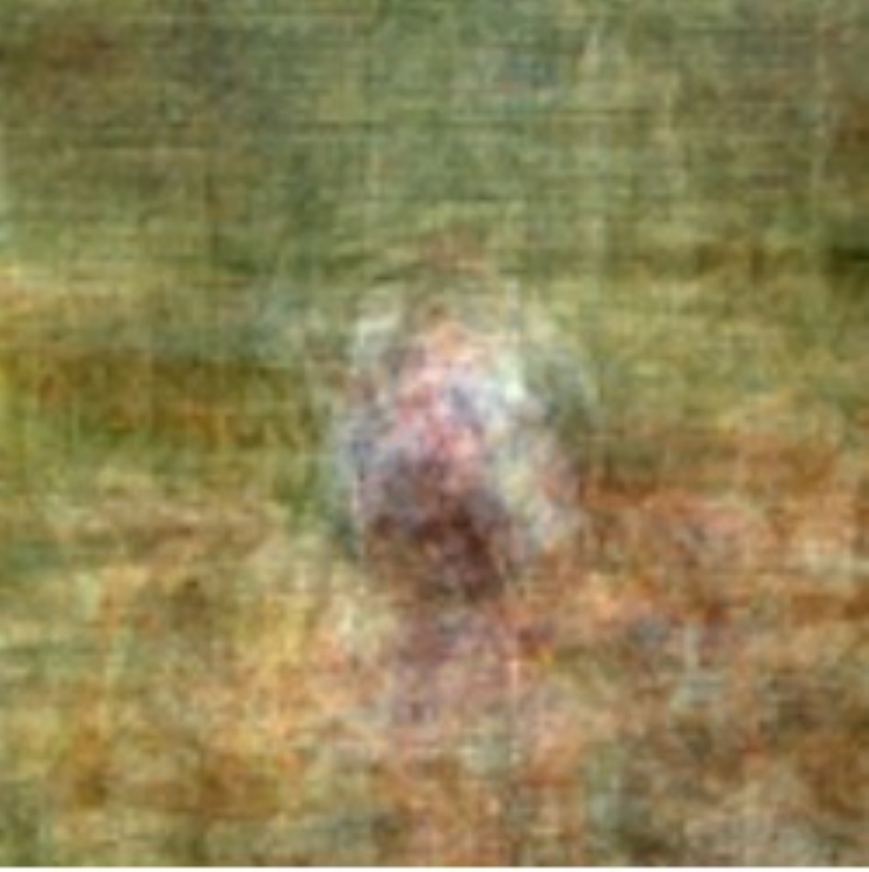} &
\includegraphics[width=0.08\textwidth]{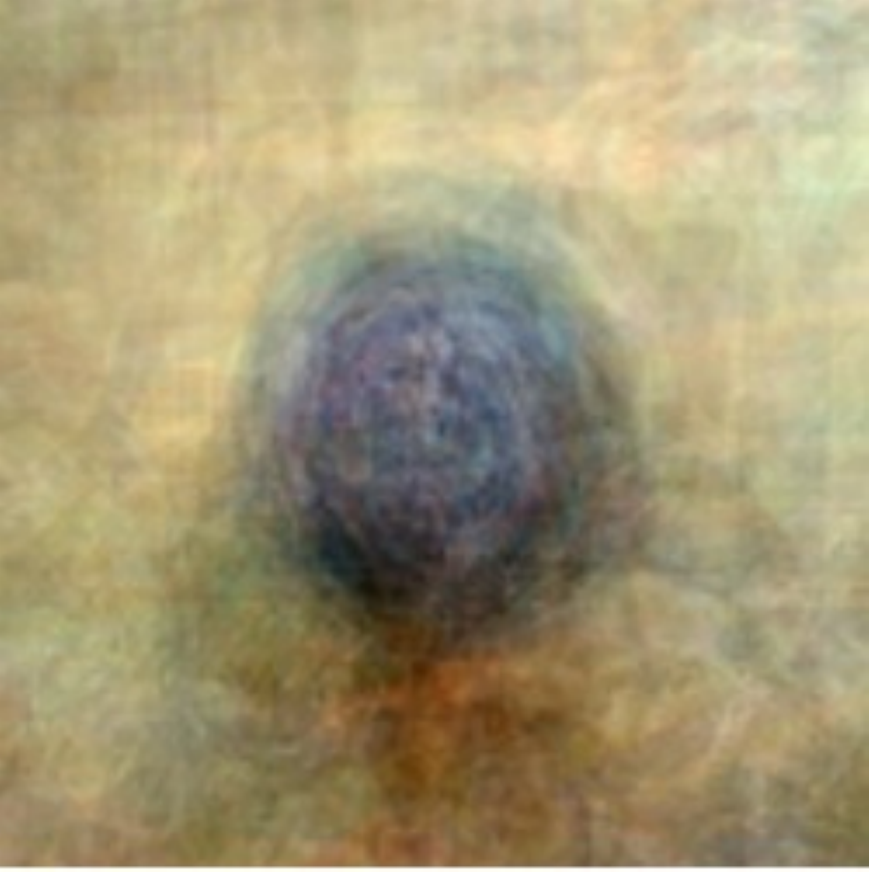} &
\includegraphics[width=0.08\textwidth]{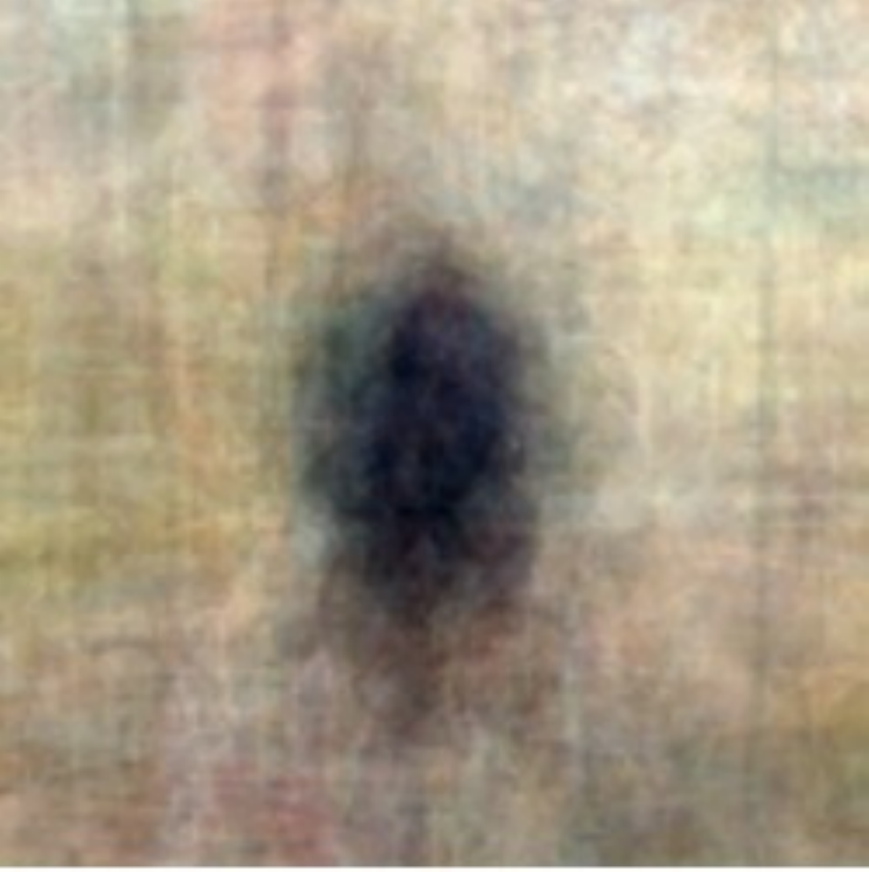} &
\includegraphics[width=0.08\textwidth]{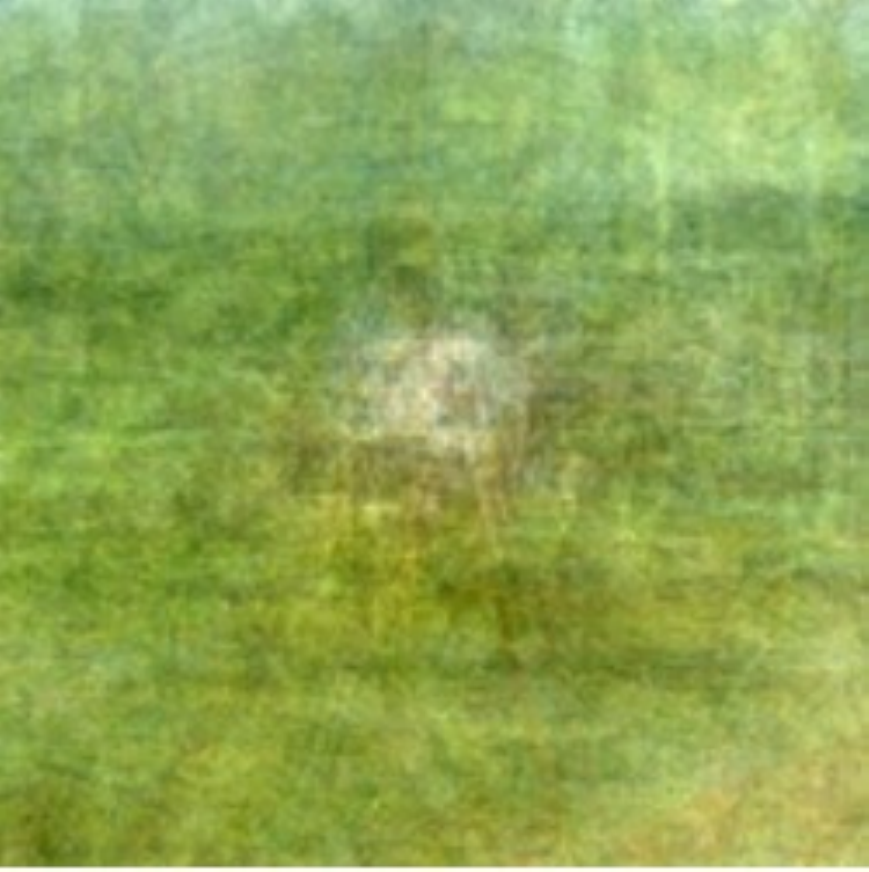} &
\includegraphics[width=0.08\textwidth]{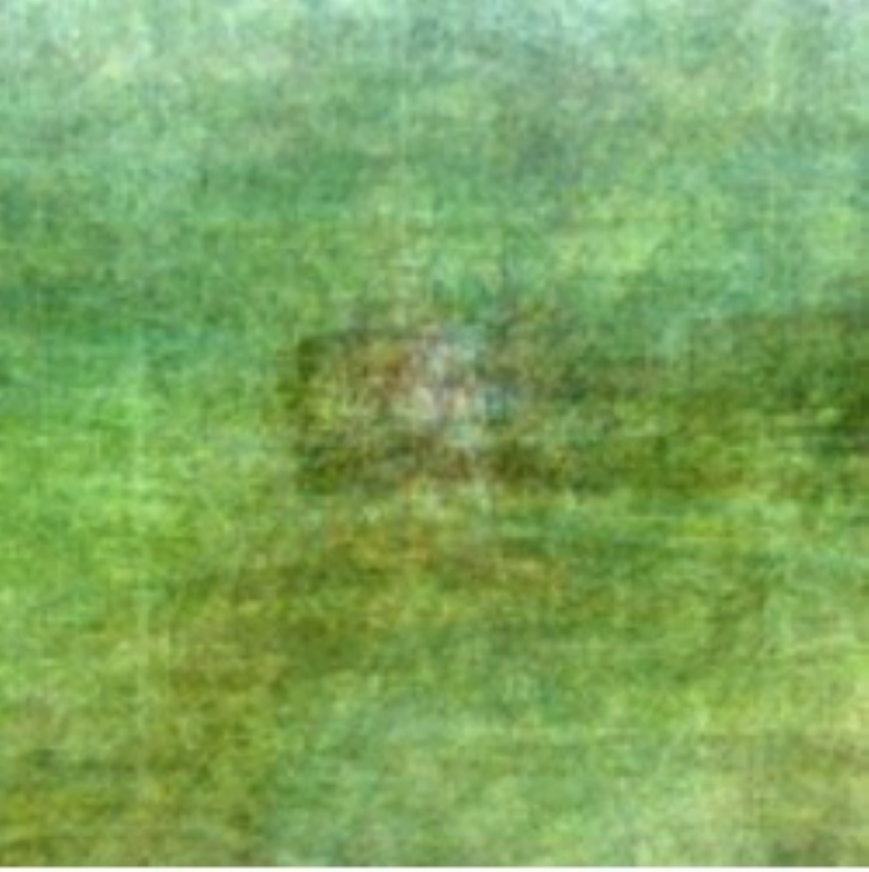} &
\includegraphics[width=0.08\textwidth]{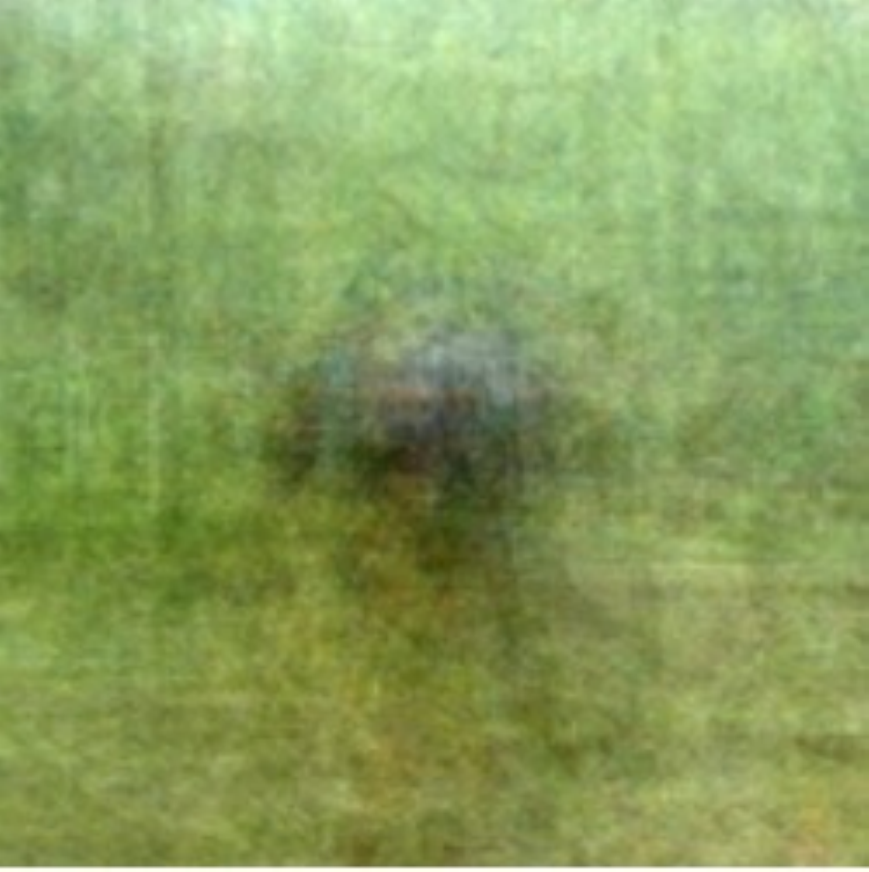} &
\includegraphics[width=0.08\textwidth]{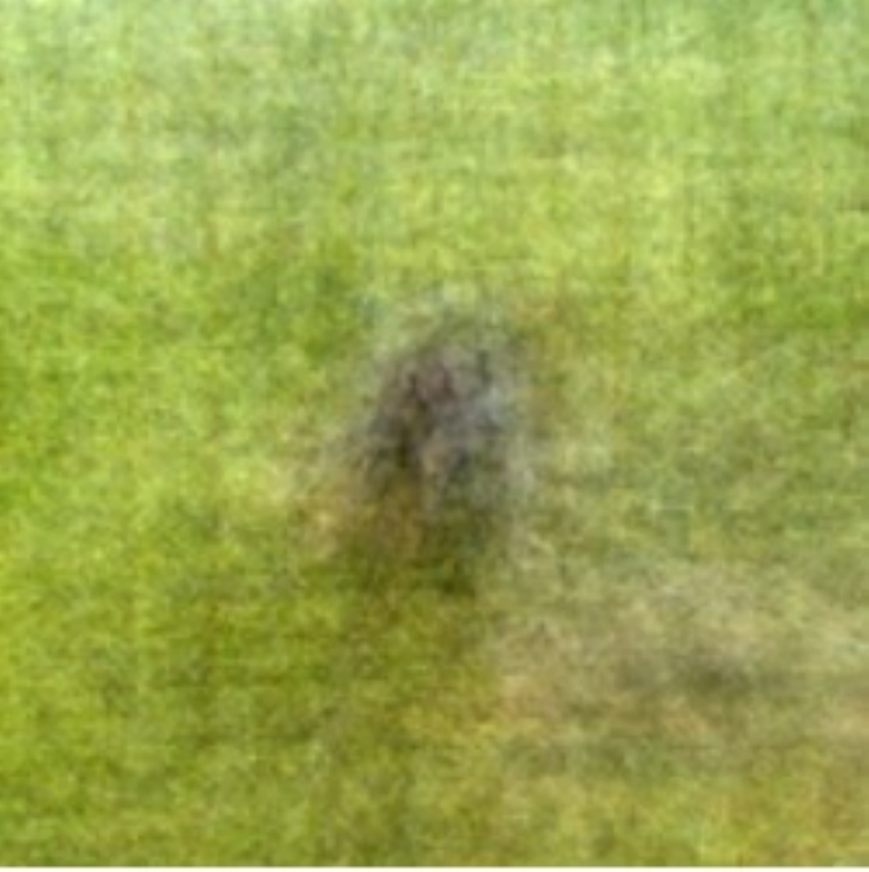} \\
\rotatebox{90}{\small 500 nbrs} &
\includegraphics[width=0.08\textwidth]{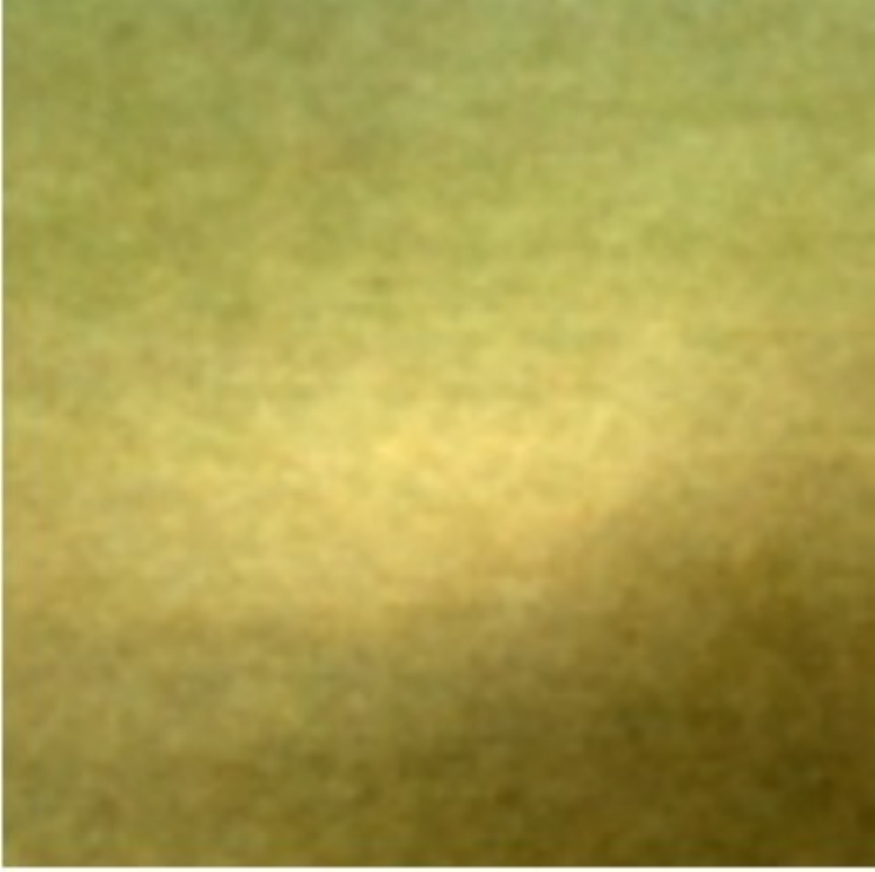} &
\includegraphics[width=0.08\textwidth]{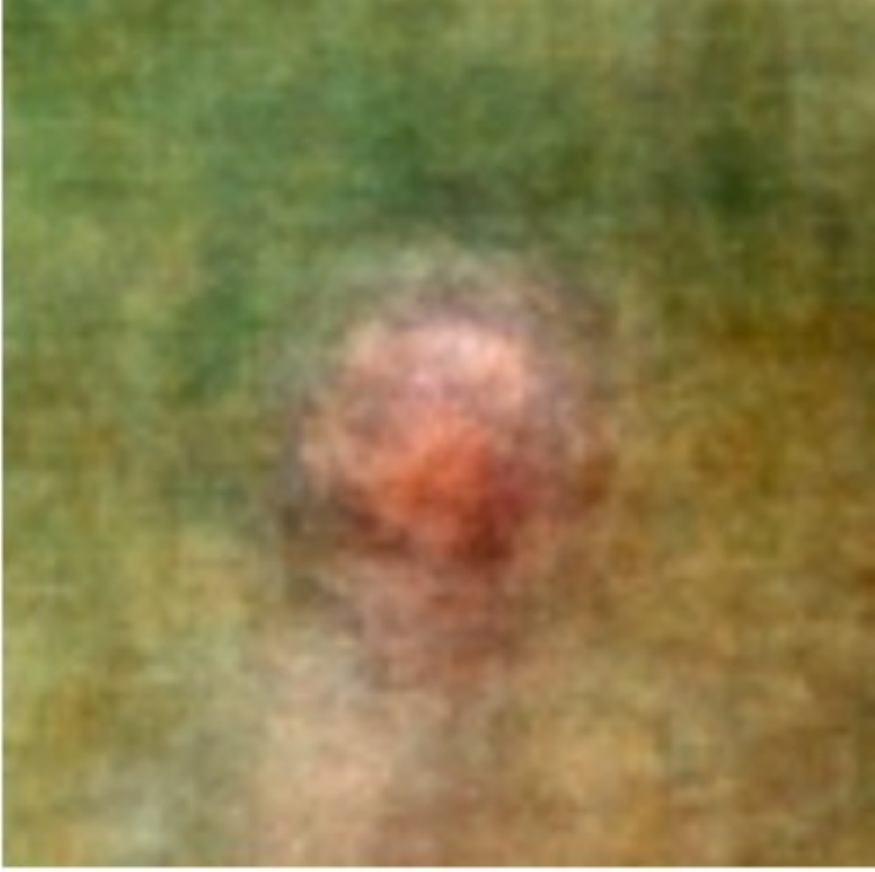} &
\includegraphics[width=0.08\textwidth]{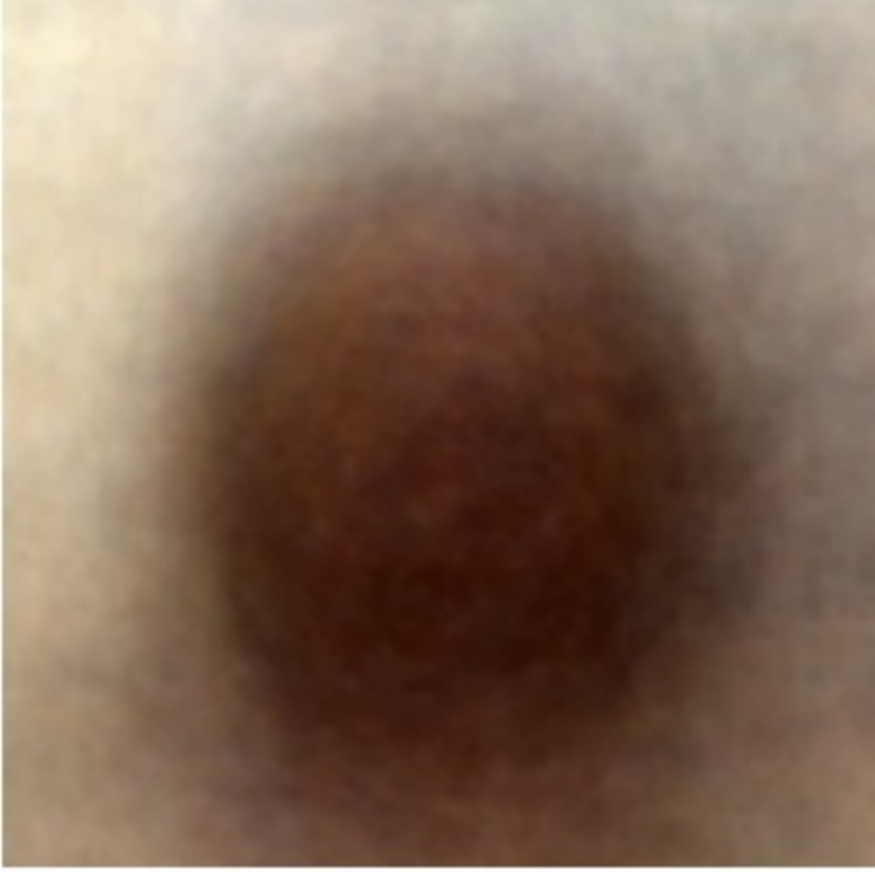} &
\includegraphics[width=0.08\textwidth]{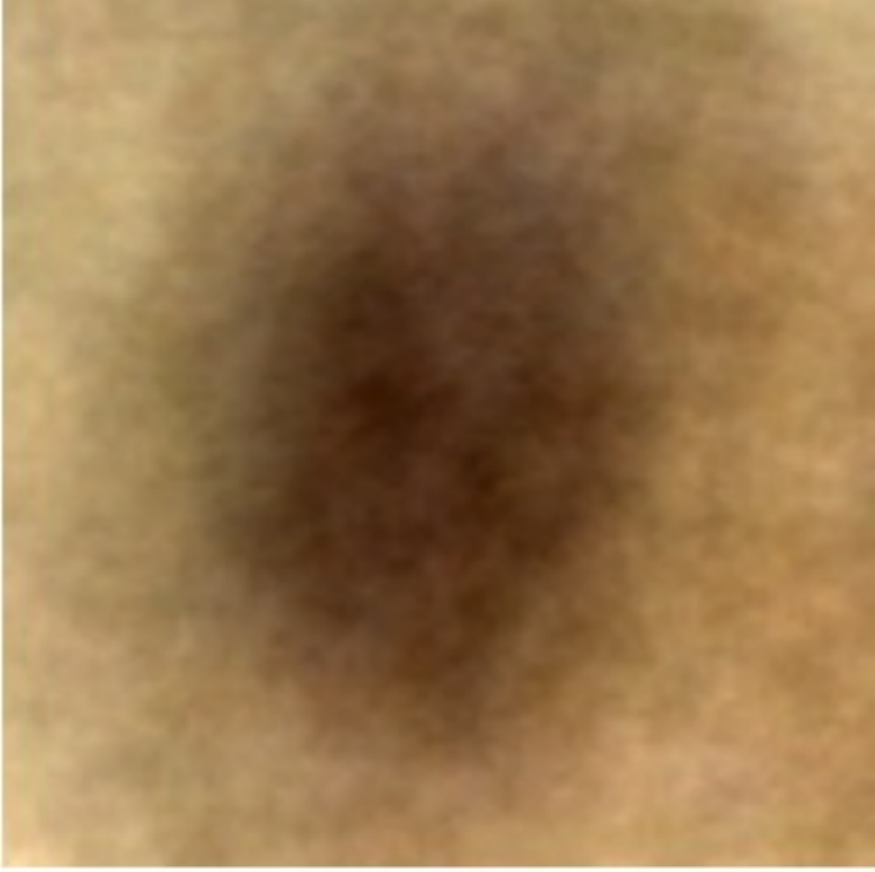} &
\includegraphics[width=0.08\textwidth]{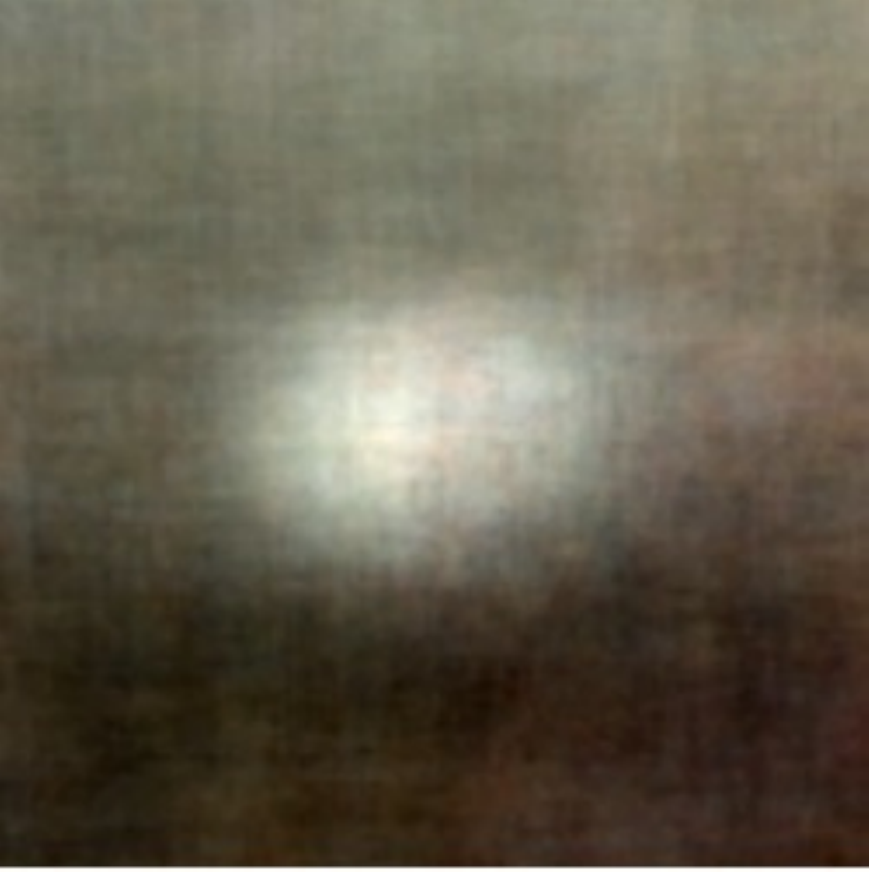} &
\includegraphics[width=0.08\textwidth]{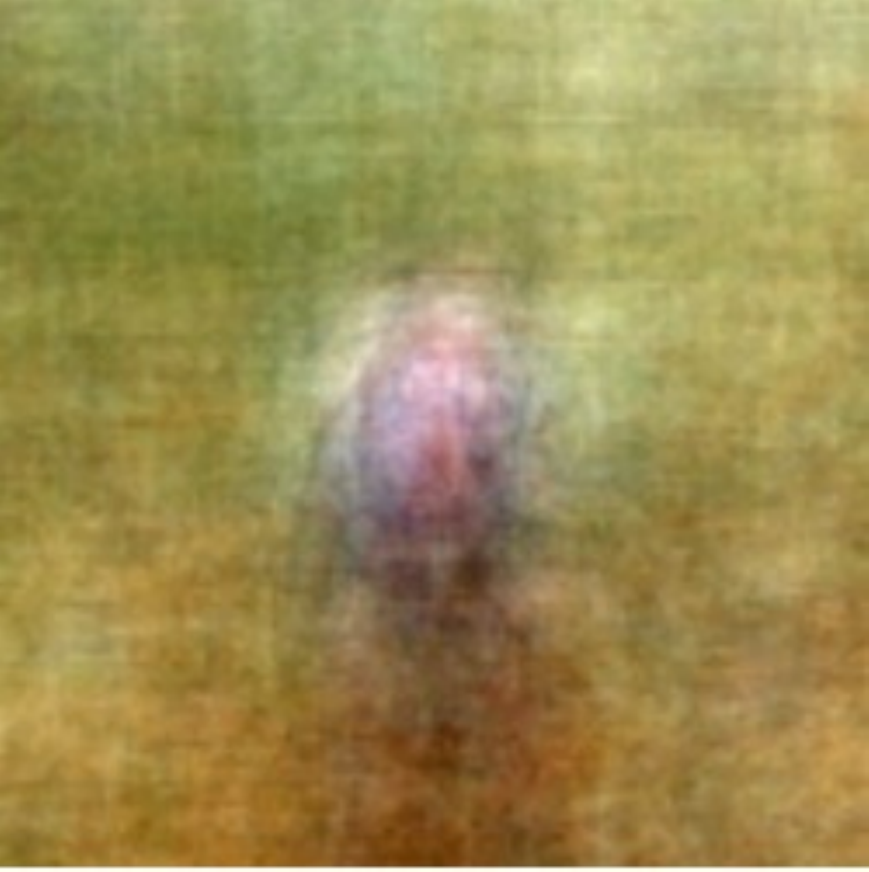} &
\includegraphics[width=0.08\textwidth]{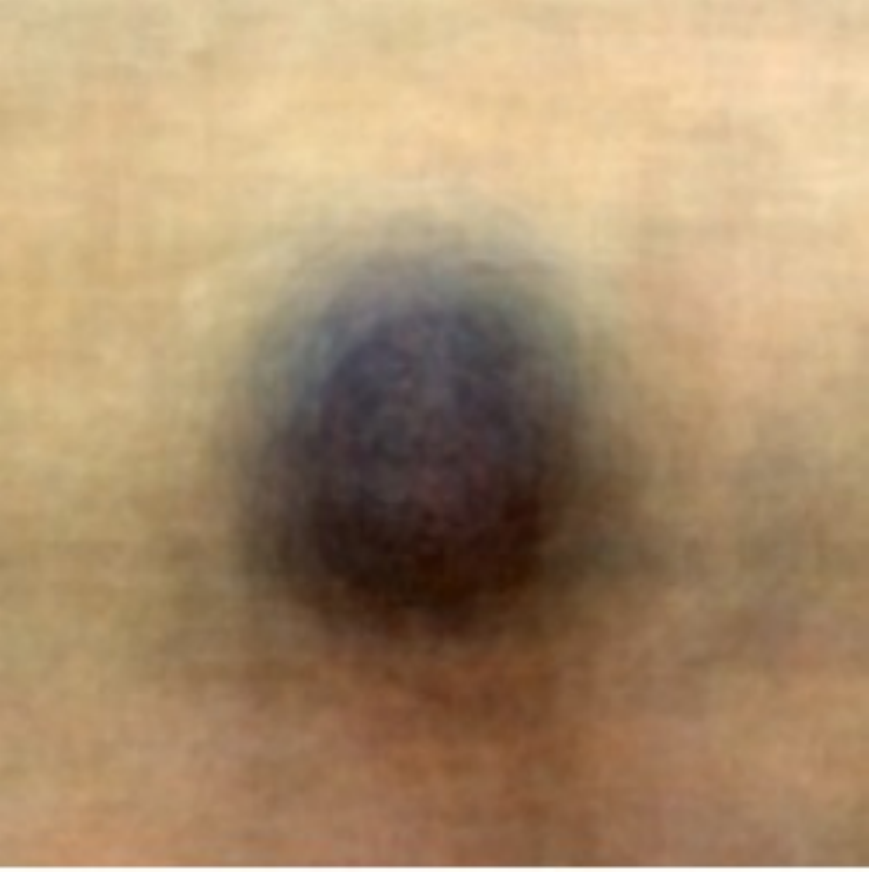} &
\includegraphics[width=0.08\textwidth]{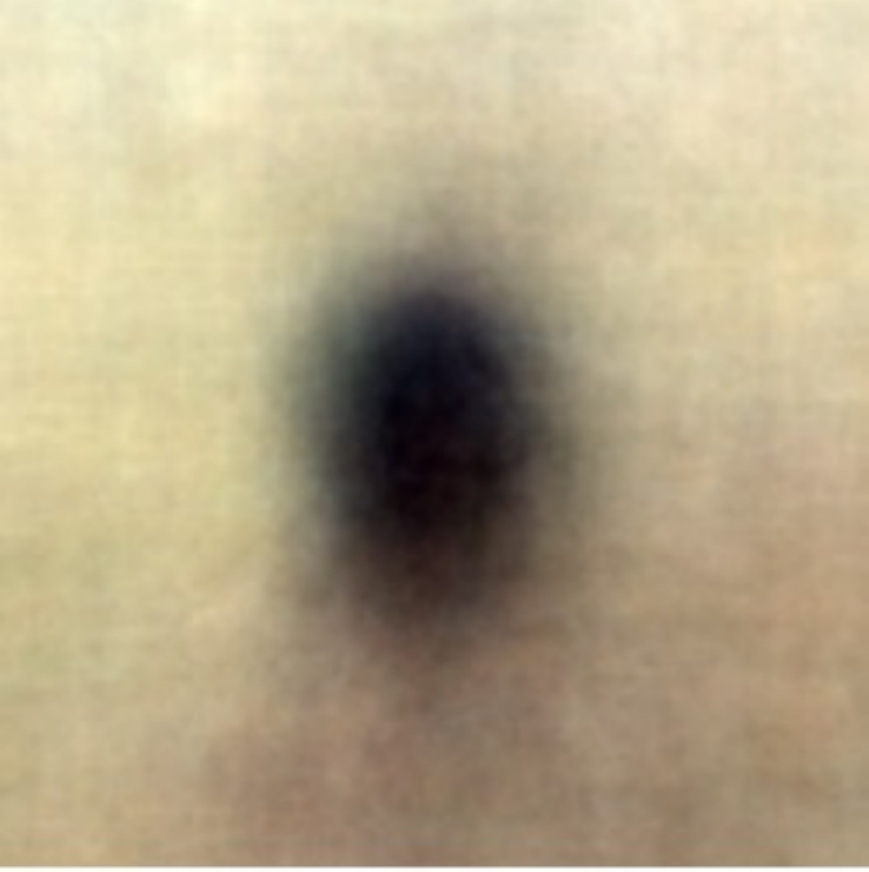} &
\includegraphics[width=0.08\textwidth]{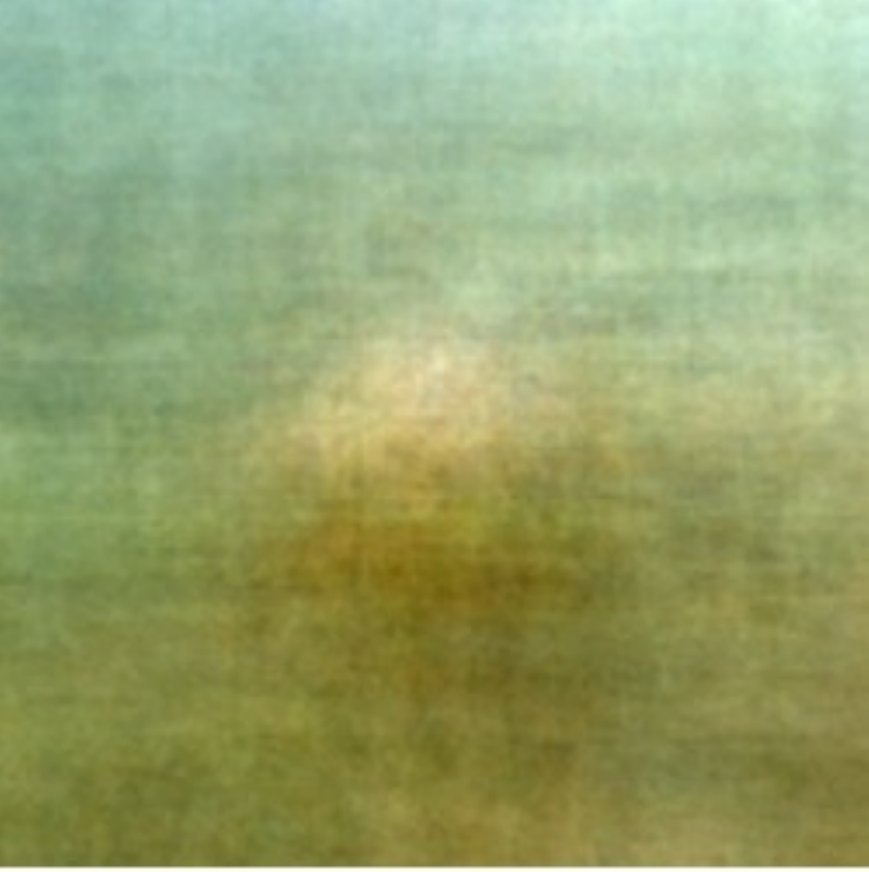} &
\includegraphics[width=0.08\textwidth]{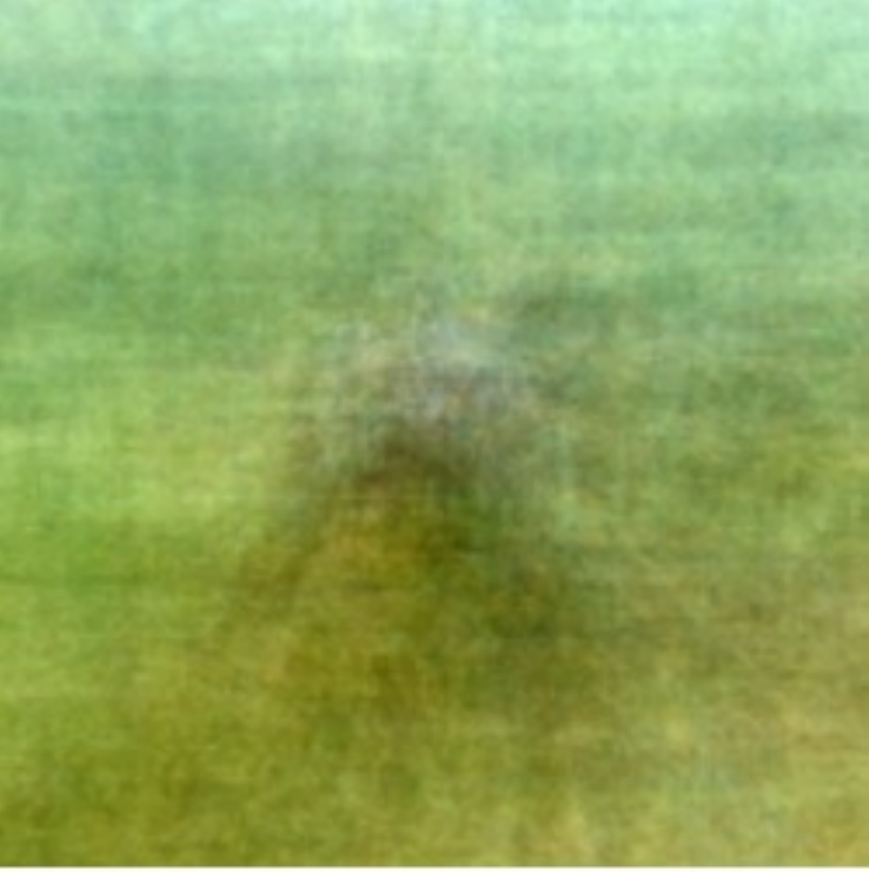} &
\includegraphics[width=0.08\textwidth]{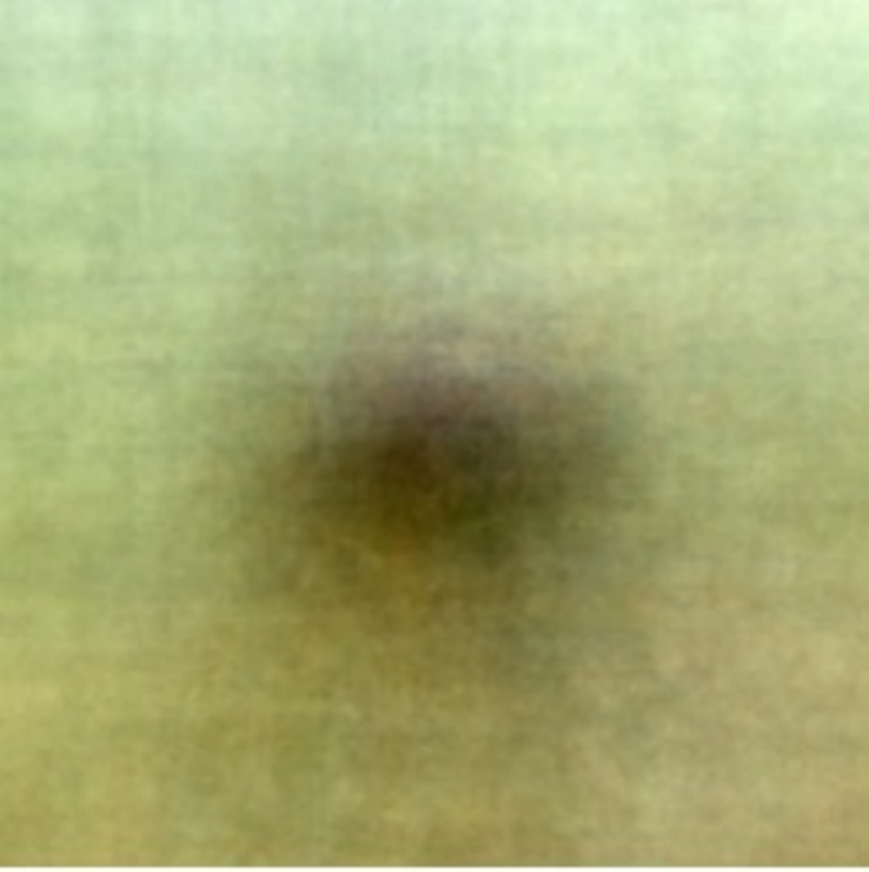} &
\includegraphics[width=0.08\textwidth]{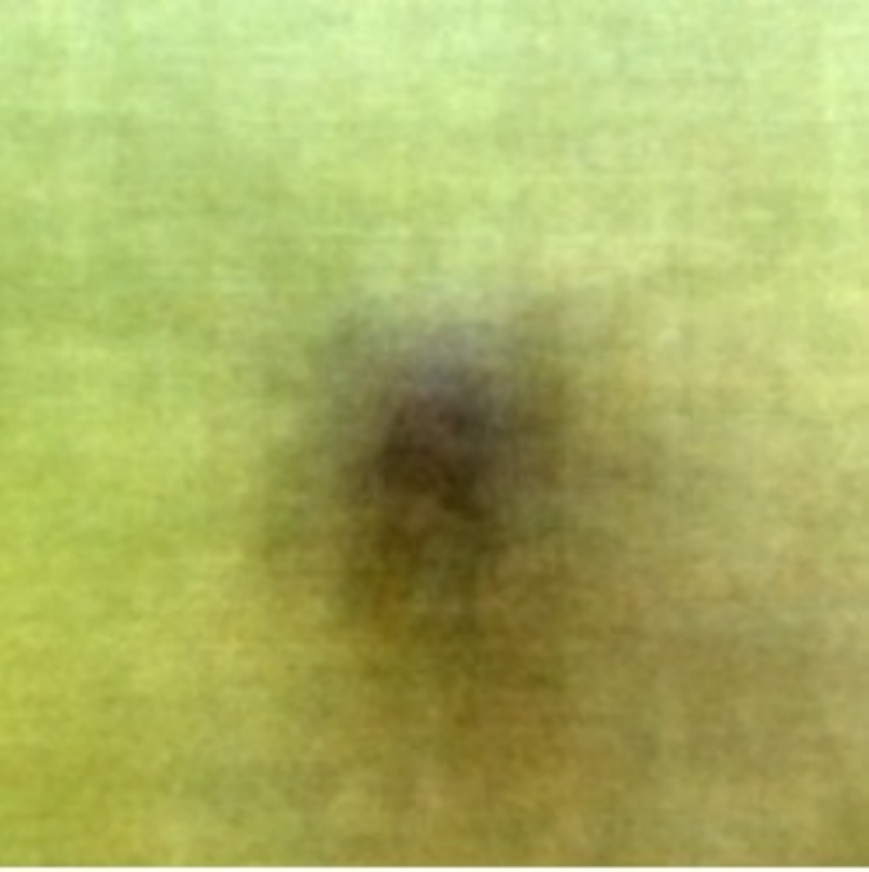}
\end{tabular}
}
\caption{
Similar convnet features tend to have similar receptive field centers.
Starting from a randomly selected seed patch occupying one rf in
\texttt{conv}3, 4, or 5, we find the nearest $k$ neighbor features computed on a
database of natural images, and average together the corresponding receptive
fields.
The contrast of each image has been expanded after averaging.
(Note that since each layer is computed with a stride of 16, there is an
upper bound on the quality of alignment that can be witnessed here.)
}
\label{fig:rfavg}
\end{figure}

In this section and Figures \ref{fig:patches} and \ref{fig:rfavg}, we provide a
novel visual investigation of the effective pooling regions of convnet features.

In Figure \ref{fig:patches}, we perform a nonparametric reconstruction of images
from features in the spirit of HOGgles \cite{hoggles}.
Rather than paired dictionary learning, however, we simply replace patches with
averages of their top-$k$ nearest neighbors in a convnet feature space.
To do so, we first compute all features at a particular layer, resulting in an
2d grid of feature vectors.
We associate each feature vector with a patch in the original image at the
center of the corresponding receptive field and with size equal to the receptive
field stride.
(Note that the strides of the receptive fields are much smaller than the
receptive fields themselves, which overlap.
Refer to Table \ref{tab:rfs} above for specific numbers.)
We replace each such patch with an average over $k$ nearest neighbor patches
using a database of features densely computed on the images of PASCAL VOC 2011.
Our database contains at least one million patches for every layer.
Features are matched by cosine similarity.

Even though the feature rfs cover large regions of the source images, the
specific resemblance of the resulting images shows that information is not
spread uniformly throughout those regions.
Notable features (e.g., the tires of the bicycle and the facial features of the
cat) are replaced in their corresponding locations.
Also note that replacement appears to become more semantic and less visually
specific as the layer deepens: the eyes and nose of the cat get replaced with
differently colored or shaped eyes and noses, and the fur gets replaced with
various animal furs, with the diversity increasing with layer number.

Figure \ref{fig:rfavg} gives a feature-centric rather than image-centric view of
feature locality.
For each column, we first pick a random seed feature vector (computed from
a PASCAL image), and find $k$ nearest neighbor features, again by cosine
similarity.
Instead of averaging only the centers, we average the entire receptive fields of
the neighbors.
The resulting images show that similar features tend to respond to similar
colors specifically in the centers of their receptive fields.

\section{Intraclass alignment}
\label{sec:flow}
We conjecture that category learning implicitly aligns instances by pooling over
a discriminative mid-level representation.
If this is true, then such features should be useful for post-hoc alignment in a
similar fashion to conventional features.
To test this, we use convnet features for the task of aligning different instances
of the same class.
We approach this difficult task in the style of SIFT flow \cite{sift-flow}: we
retrieve near neighbors using a coarse similarity measure, and then compute
dense correspondences on which we impose an MRF smoothness prior which finally
allows all images to be warped into alignment.

Nearest neighbors are computed using \texttt{fc}7 features.
Since we are specifically testing the quality of alignment, we use the
same nearest neighbors for convnet or conventional features, and we compute both
types of features at the same locations, the grid of convnet rf centers in the
response to a single image.

Alignment is determined by solving an MRF formulated on this grid of feature
locations.
Let $p$ be a point on this grid, let $f_s(p)$ be the feature vector of the
source image at that point, and let $f_t(p)$ be the feature vector of the target
image at that point.
For each feature grid location $p$ of the source image, there is a vector $w(p)$
giving the displacement of the corresponding feature in the target image.
We use the energy function
\[
E(w) = \sum_p \| f_s(p) - f_t(p + w(p)) \|_2 +
\beta \sum_{(p, q) \in \mathcal E} \| w(p) - w(q) \|_2^2,
\]
where $\mathcal E$ are the edges of a 4-neighborhood graph and $\beta$ is the
regularization parameter.
Optimization is performed using belief propagation, with the techniques
suggested in \cite{FHBP}.
Message passing is performed efficiently using the squared Euclidean distance
transform \cite{FHDT}.
(Unlike the $L_1$ regularization originally used by SIFT flow \cite{sift-flow},
this formulation maintains rotational invariance of $w$.)

Based on its performance in the next section, we use \texttt{conv}4 as our
convnet feature, and SIFT with descriptor radius 20 as our conventional feature.
From validation experiments, we set $\beta = 3 \cdot 10^{-3}$ for both
\texttt{conv}4 and SIFT features (which have a similar scale).

\begin{figure}[t]
\centering
\renewcommand{\tabcolsep}{1pt}
\scalebox{0.95}{
\begin{tabular}{crccccc}
target image & & \multicolumn{5}{c}{five nearest neighbors} \\
%\includegraphics[width=0.16\textwidth]{figures/align_cat889_keypoints_nn} &
%\rotatebox{90}{\hspace{2em}transfer} &
%\includegraphics[width=0.16\textwidth]{figures/align_cat889_warp1_nn} &
%\includegraphics[width=0.16\textwidth]{figures/align_cat889_warp2_nn} &
%\includegraphics[width=0.16\textwidth]{figures/align_cat889_warp3_nn} &
%\includegraphics[width=0.16\textwidth]{figures/align_cat889_warp4_nn} &
%\includegraphics[width=0.16\textwidth]{figures/align_cat889_warp5_nn} \\
\includegraphics[width=0.16\textwidth]{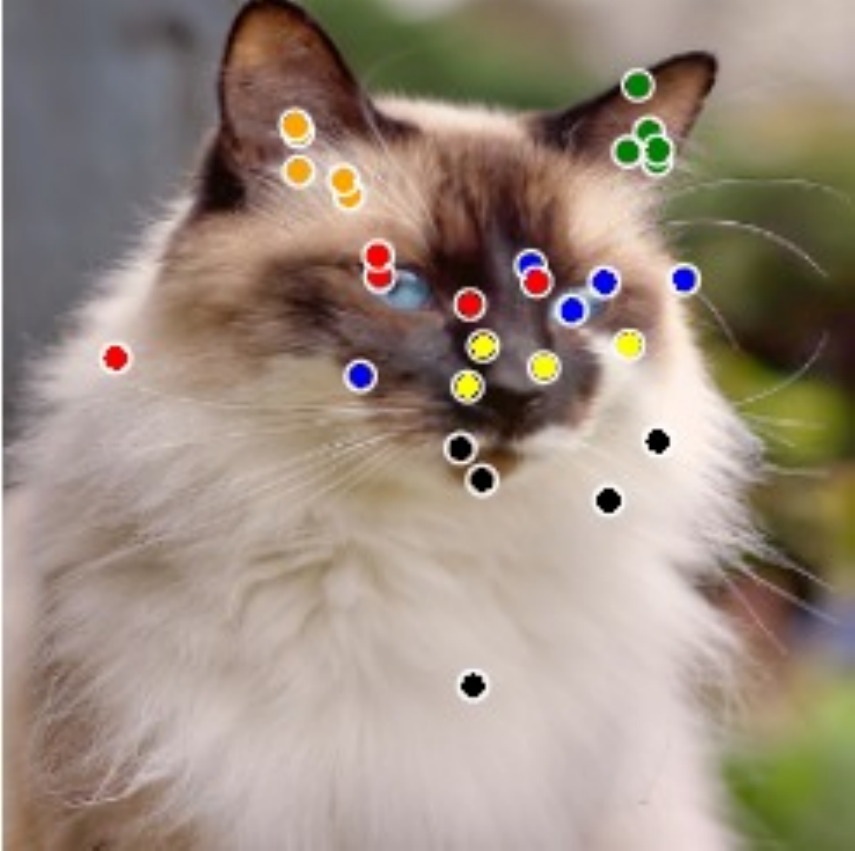} &
\rotatebox{90}{\hspace{1em}\texttt{conv}4 flow} &
\includegraphics[width=0.16\textwidth]{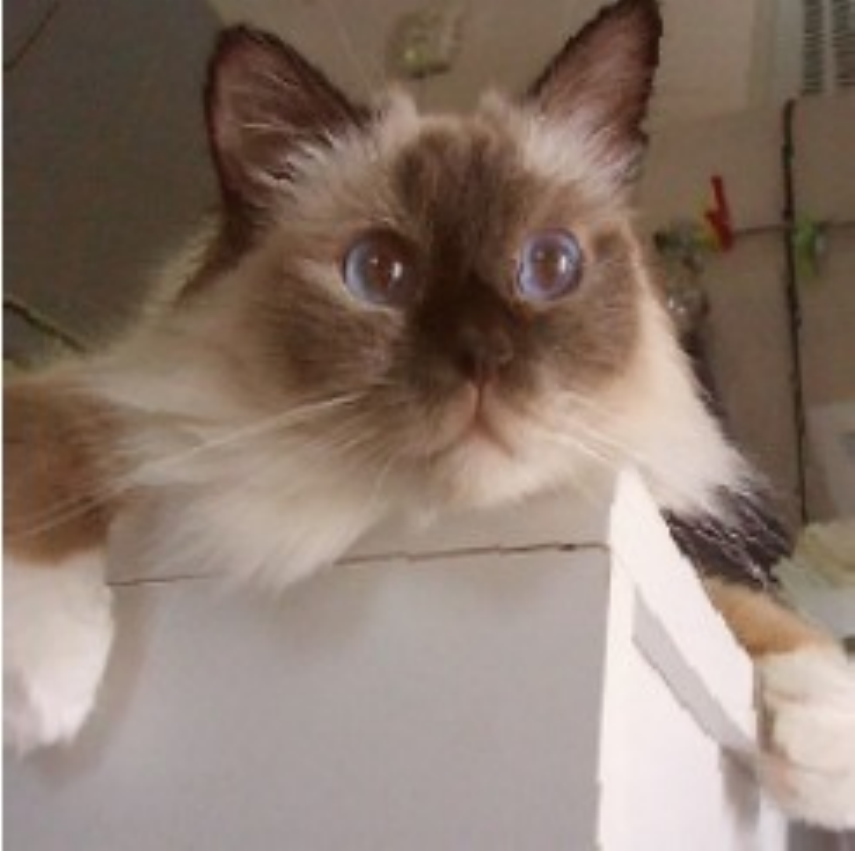} &
\includegraphics[width=0.16\textwidth]{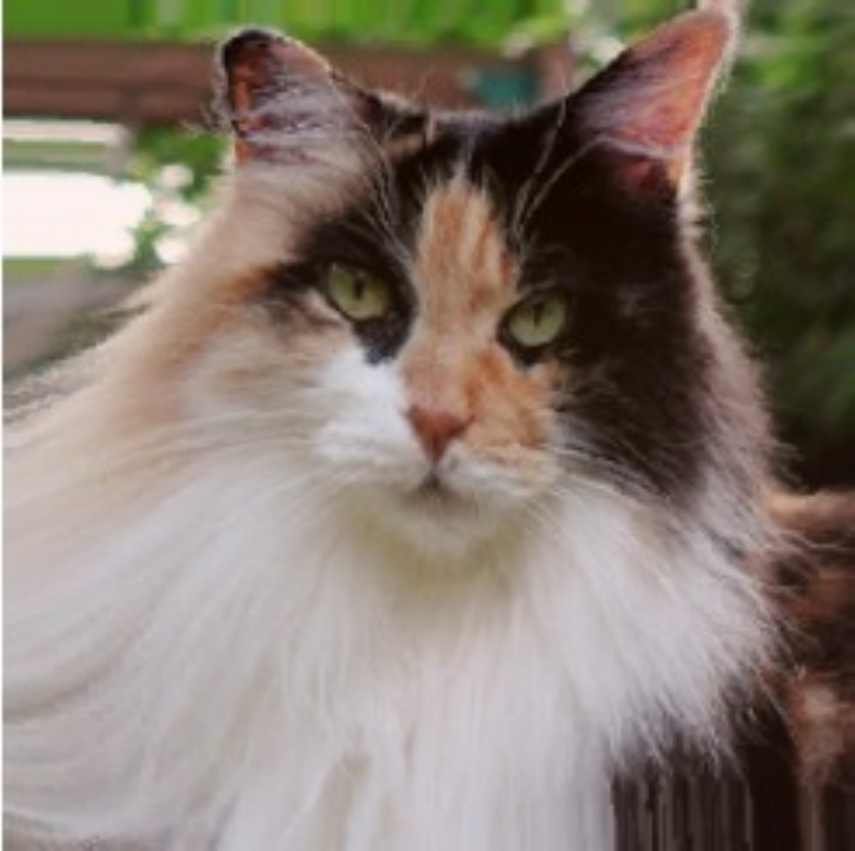} &
\includegraphics[width=0.16\textwidth]{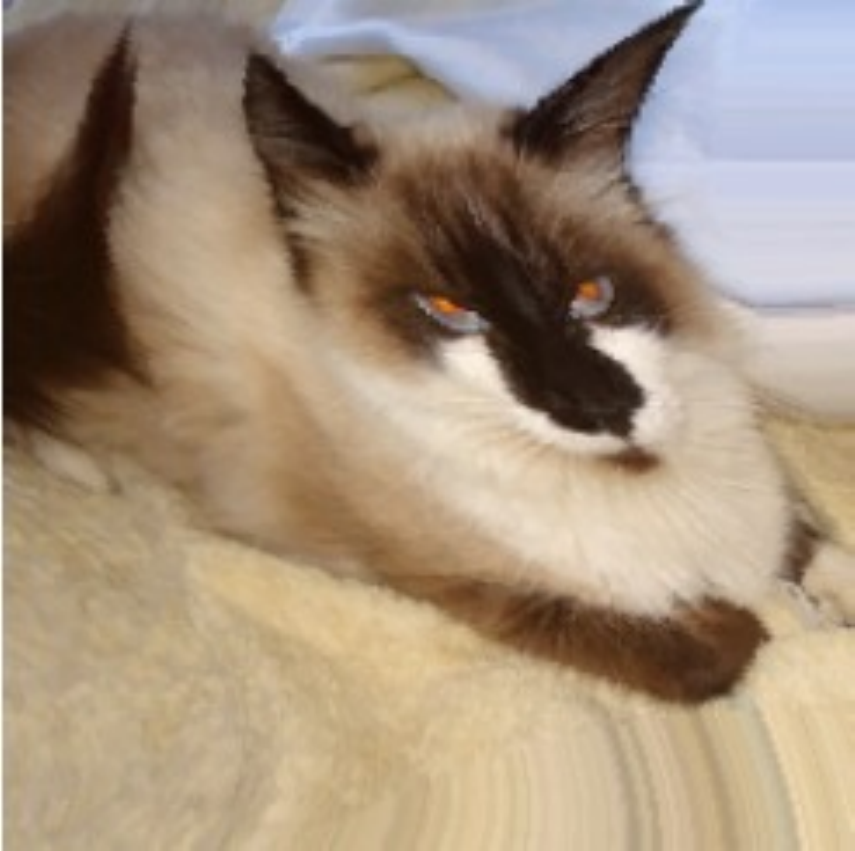} &
\includegraphics[width=0.16\textwidth]{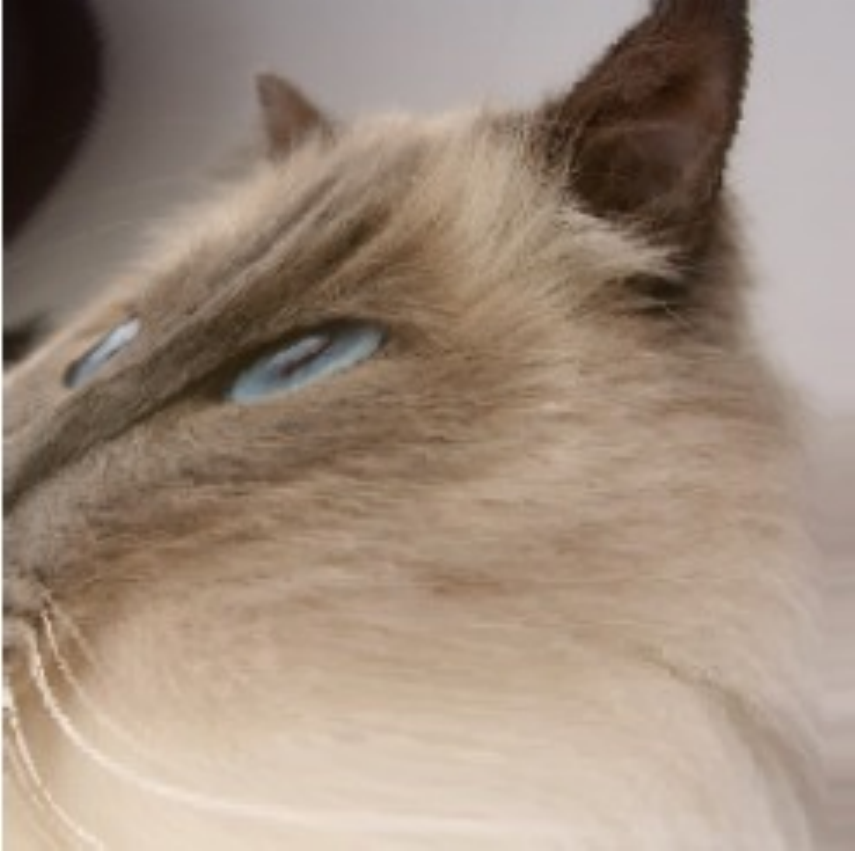} &
\includegraphics[width=0.16\textwidth]{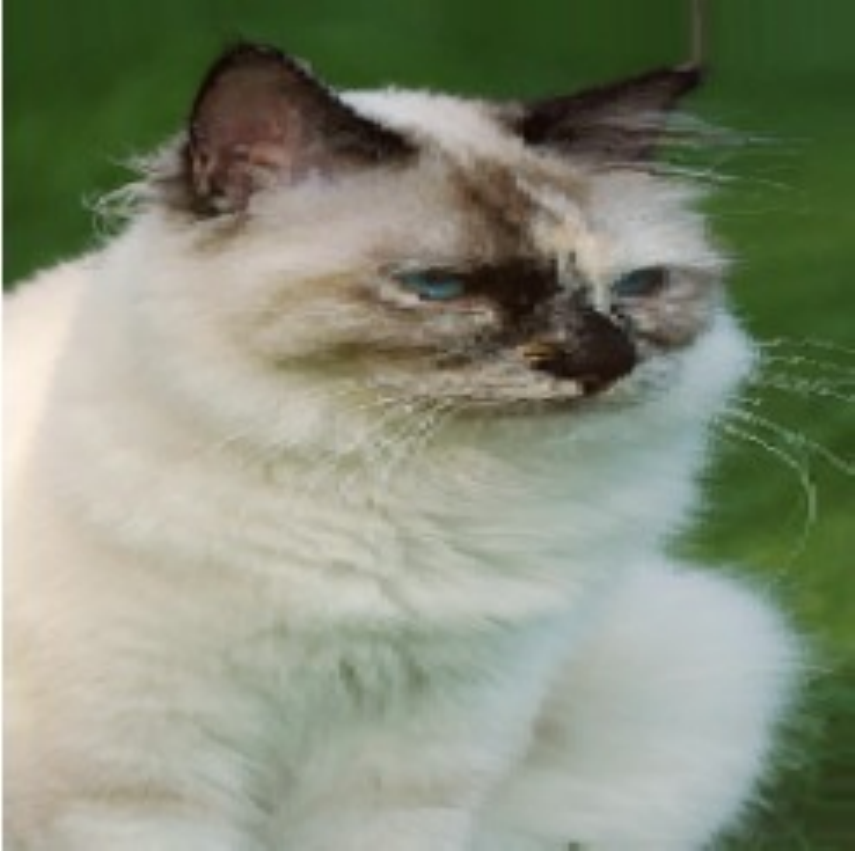} \\
\includegraphics[width=0.16\textwidth]{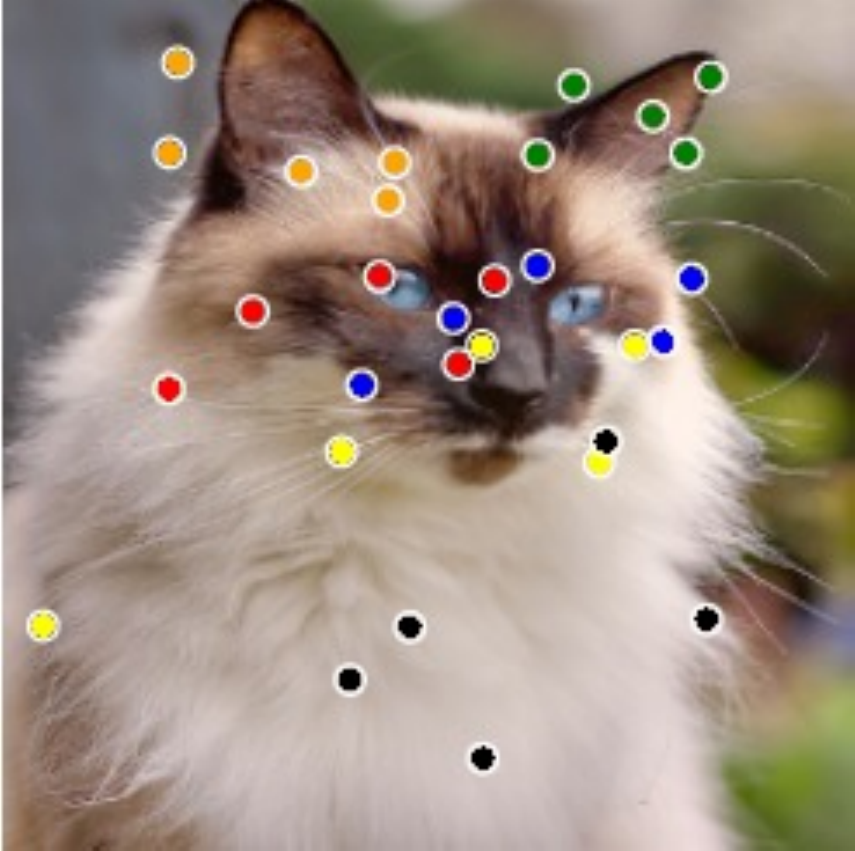} &
\rotatebox{90}{\hspace{1.5em}SIFT flow} &
\includegraphics[width=0.16\textwidth]{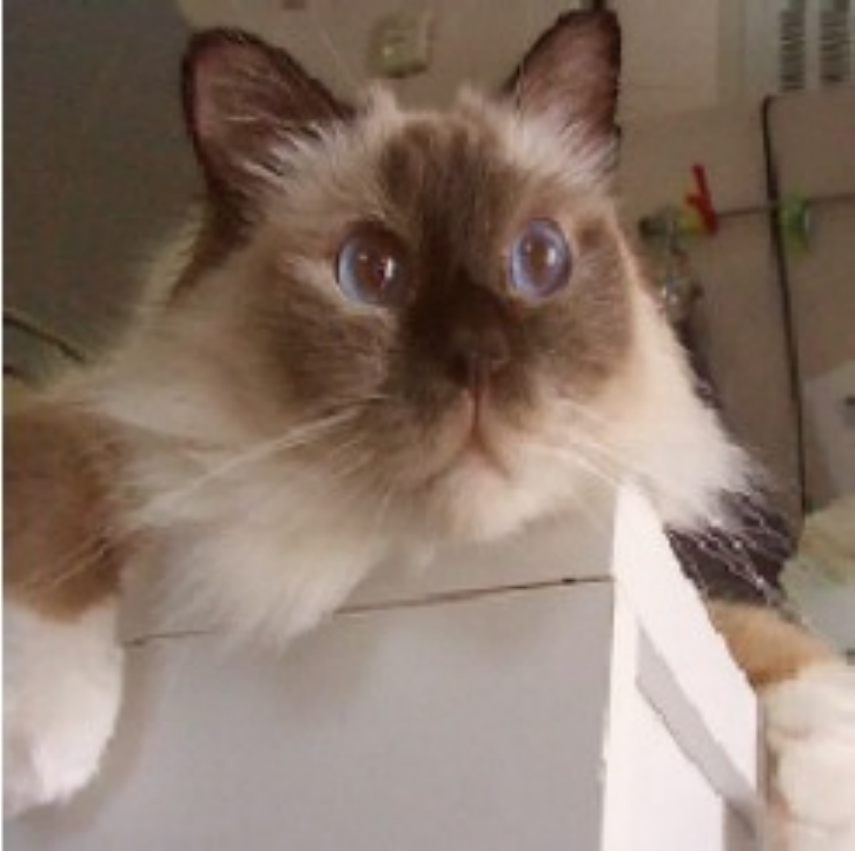} &
\includegraphics[width=0.16\textwidth]{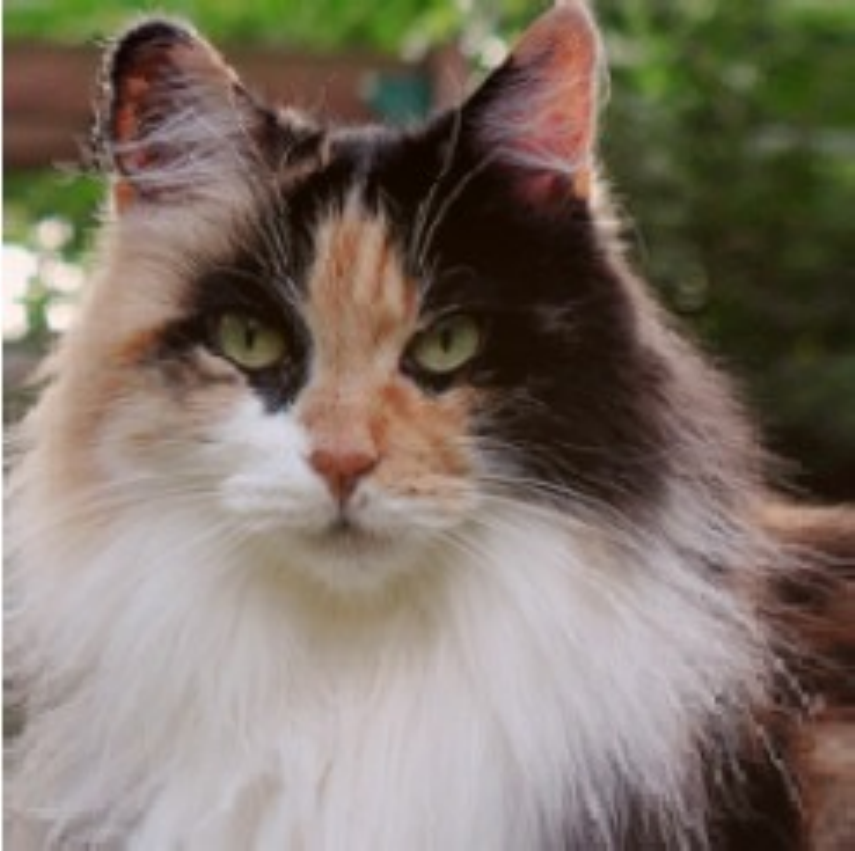} &
\includegraphics[width=0.16\textwidth]{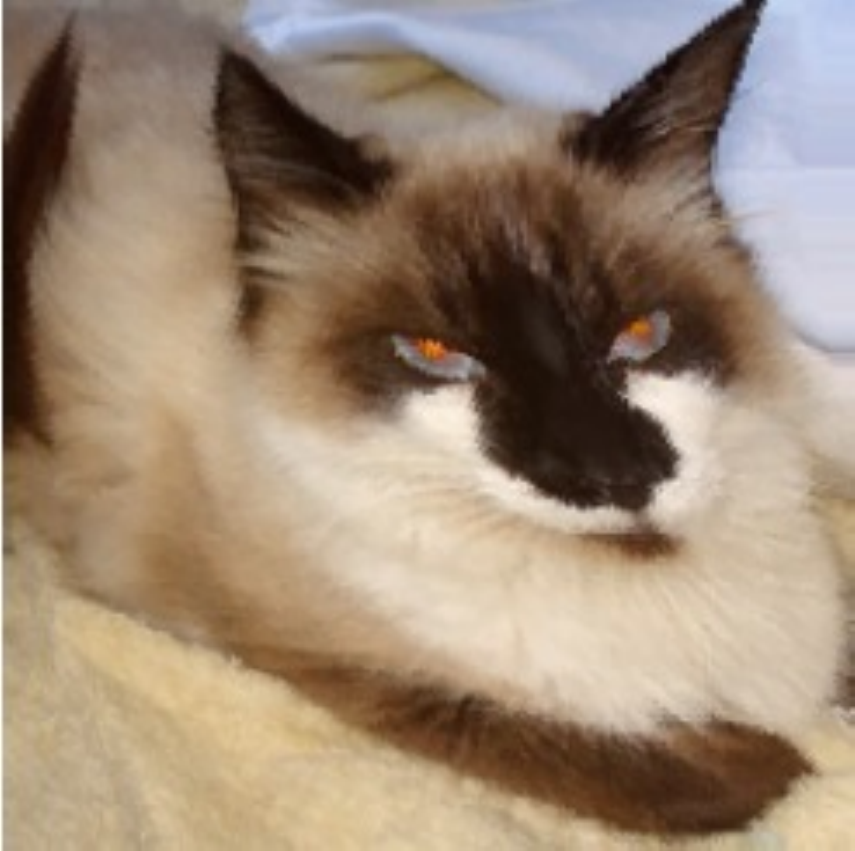} &
\includegraphics[width=0.16\textwidth]{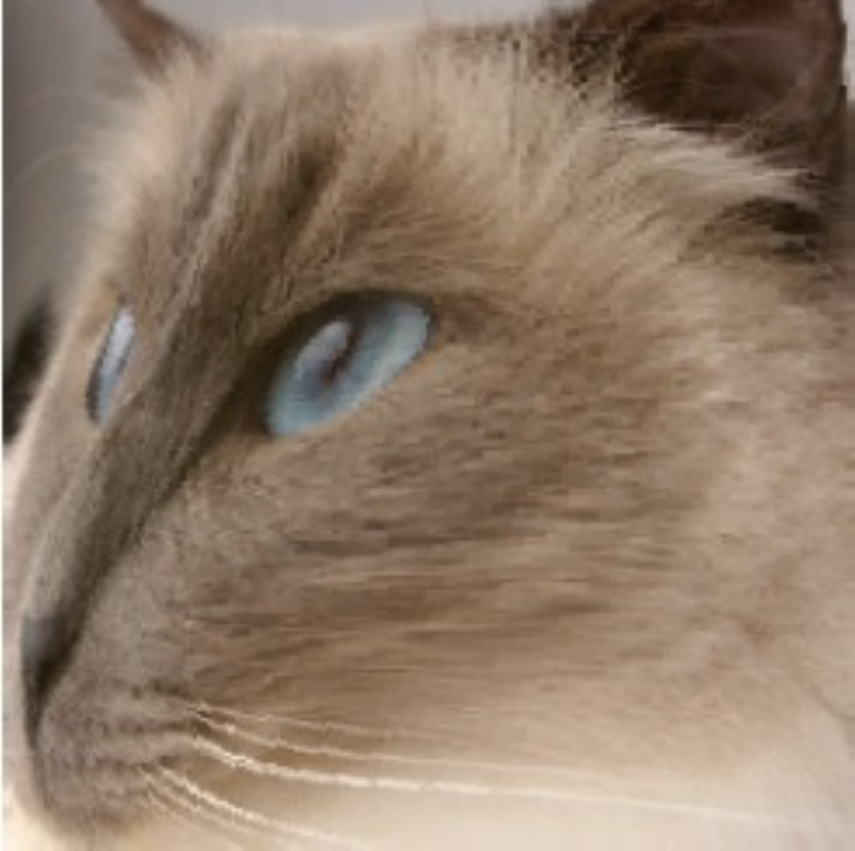} &
\includegraphics[width=0.16\textwidth]{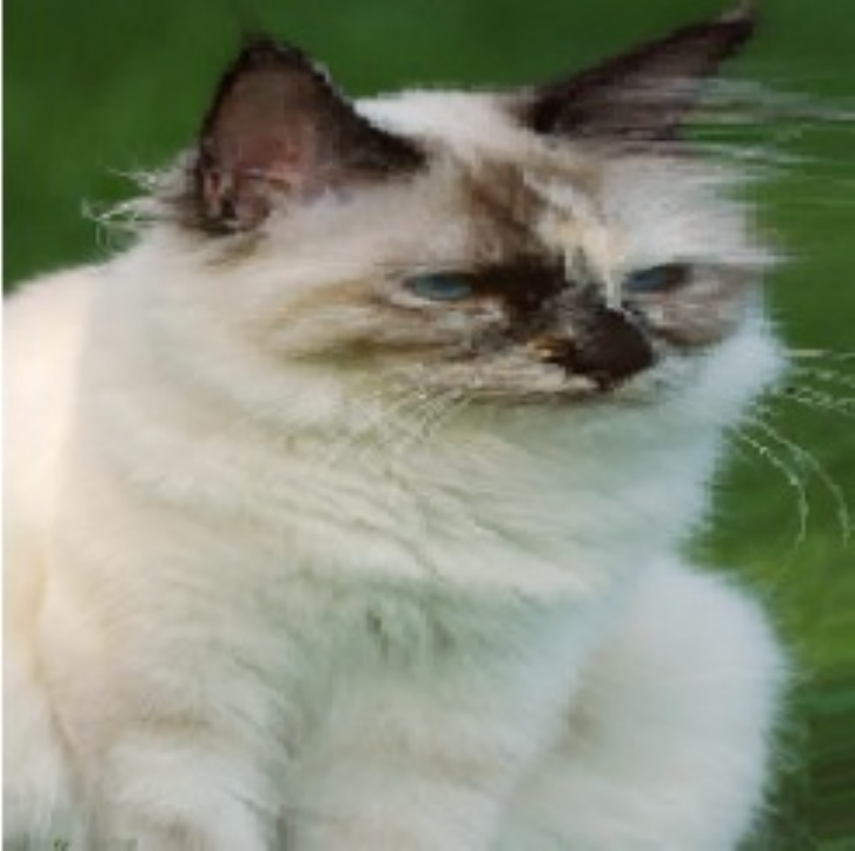} \\
%\includegraphics[width=0.16\textwidth]{figures/align_car190_keypoints_nn} &
%\rotatebox{90}{\hspace{2em}transfer} &
%\includegraphics[width=0.16\textwidth]{figures/align_car190_warp1_nn} &
%\includegraphics[width=0.16\textwidth]{figures/align_car190_warp2_nn} &
%\includegraphics[width=0.16\textwidth]{figures/align_car190_warp3_nn} &
%\includegraphics[width=0.16\textwidth]{figures/align_car190_warp4_nn} &
%\includegraphics[width=0.16\textwidth]{figures/align_car190_warp5_nn} \\
\includegraphics[width=0.16\textwidth]{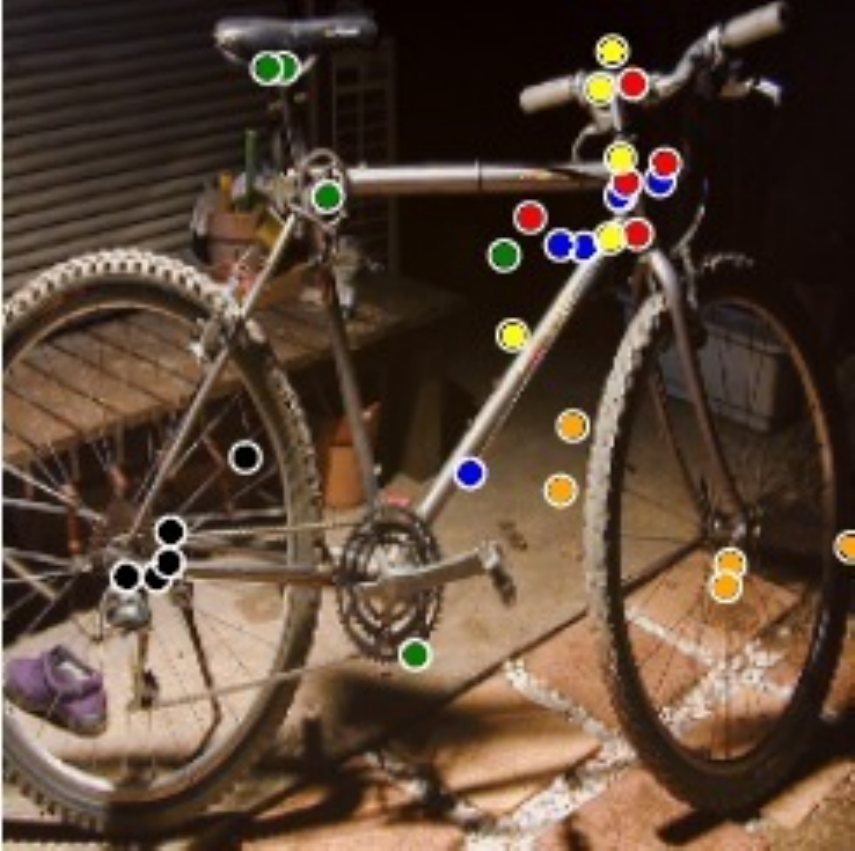} &
\rotatebox{90}{\hspace{1em}\texttt{conv}4 flow} &
\includegraphics[width=0.16\textwidth]{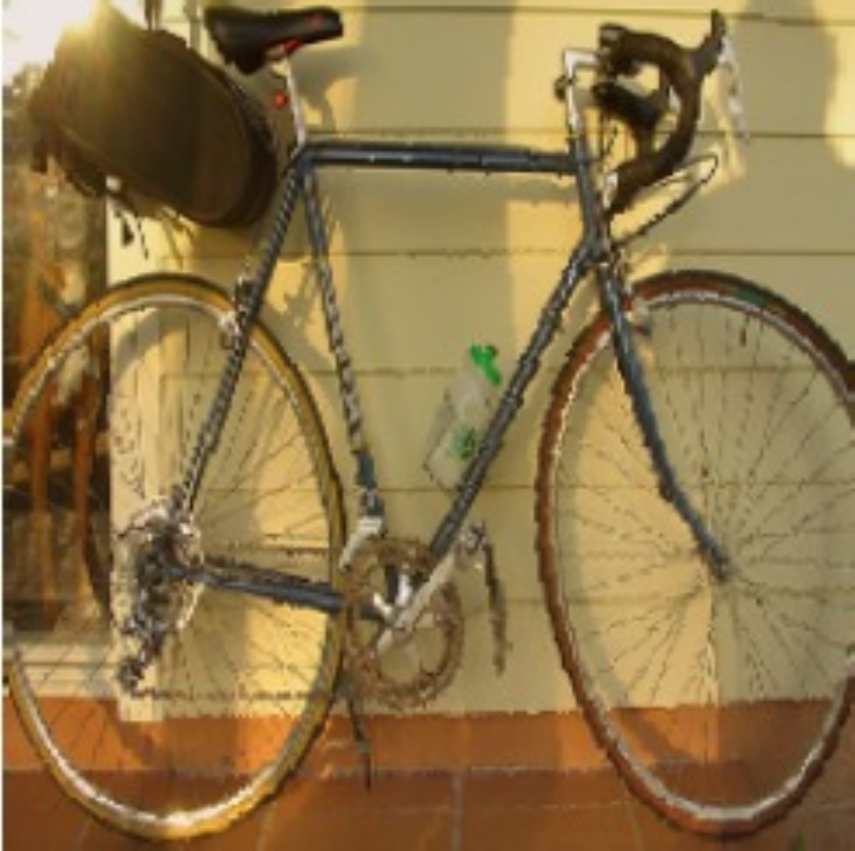} &
\includegraphics[width=0.16\textwidth]{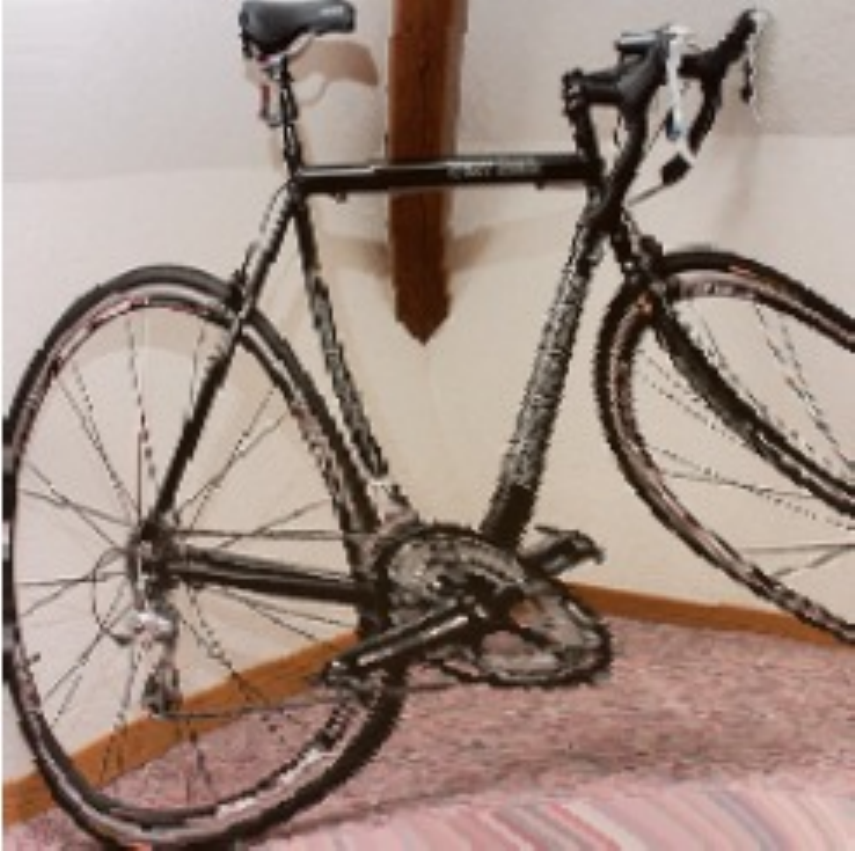} &
\includegraphics[width=0.16\textwidth]{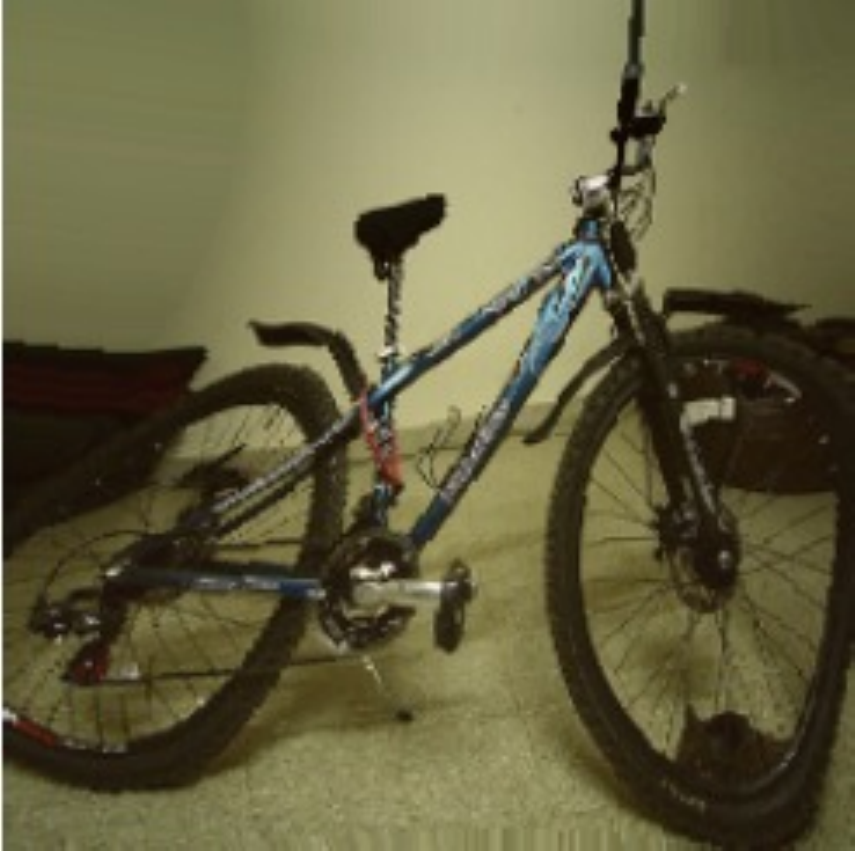} &
\includegraphics[width=0.16\textwidth]{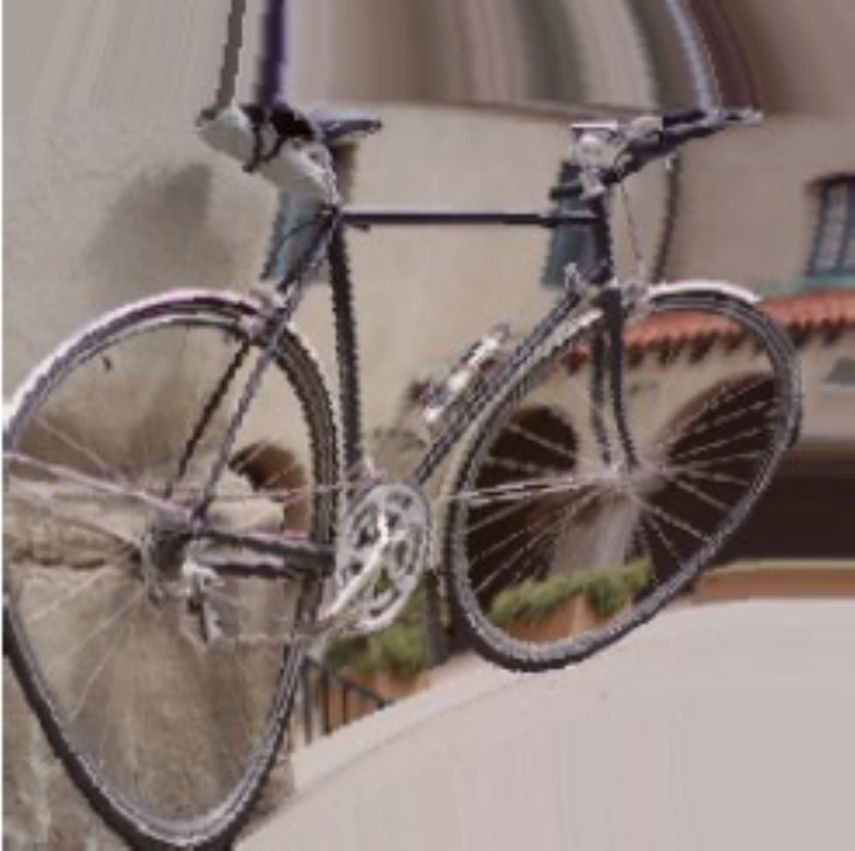} &
\includegraphics[width=0.16\textwidth]{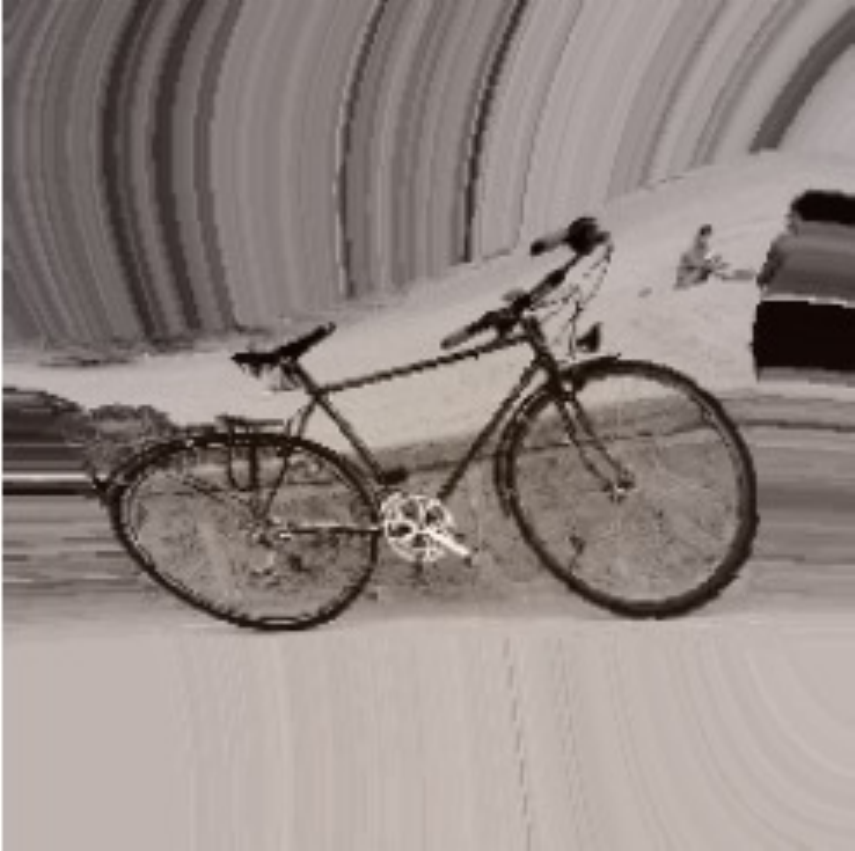} \\
\includegraphics[width=0.16\textwidth]{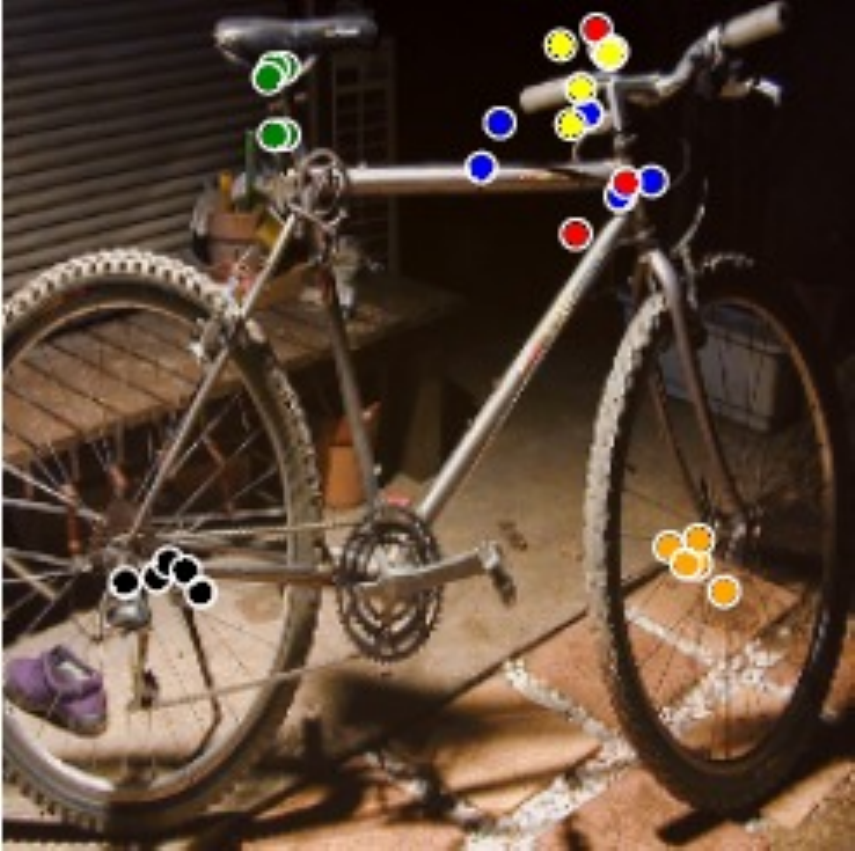} &
\rotatebox{90}{\hspace{1.5em}SIFT flow} &
\includegraphics[width=0.16\textwidth]{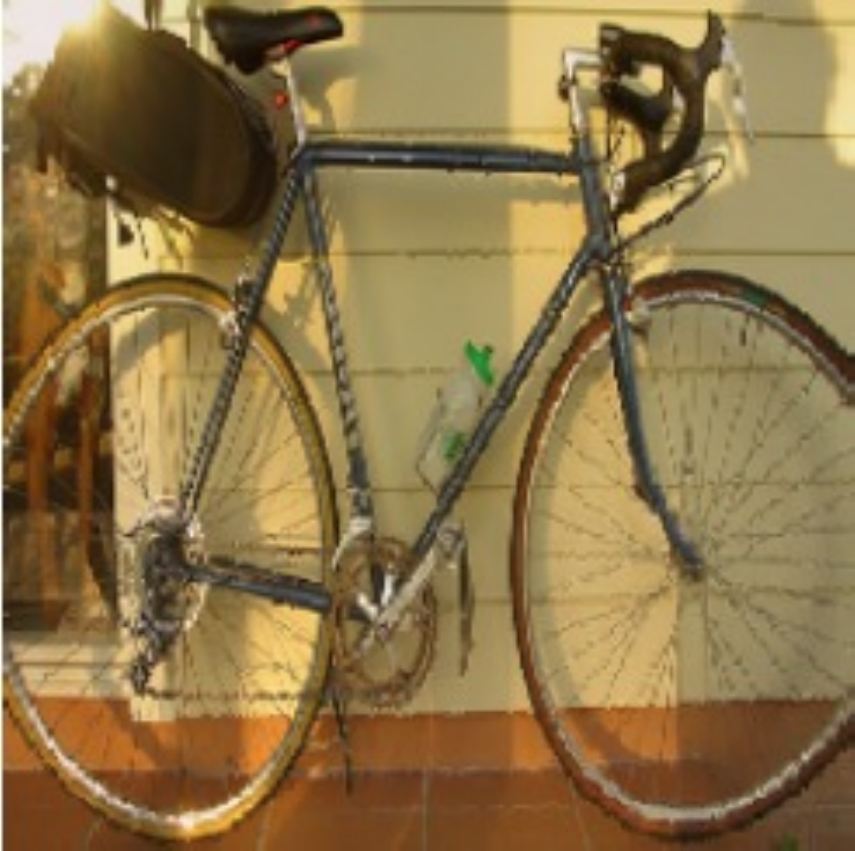} &
\includegraphics[width=0.16\textwidth]{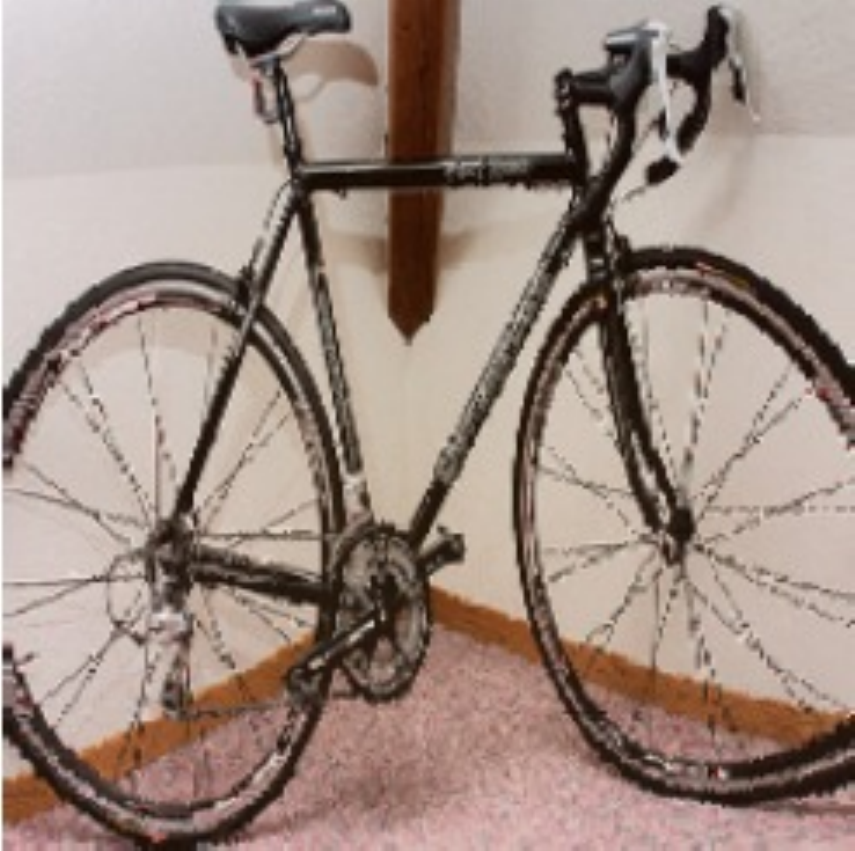} &
\includegraphics[width=0.16\textwidth]{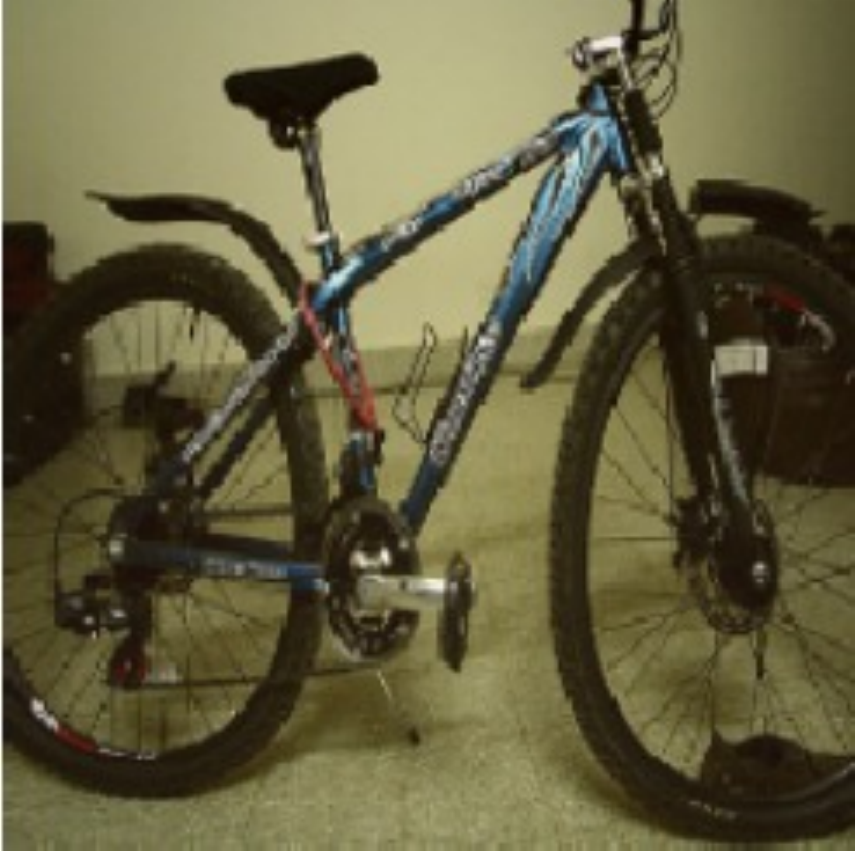} &
\includegraphics[width=0.16\textwidth]{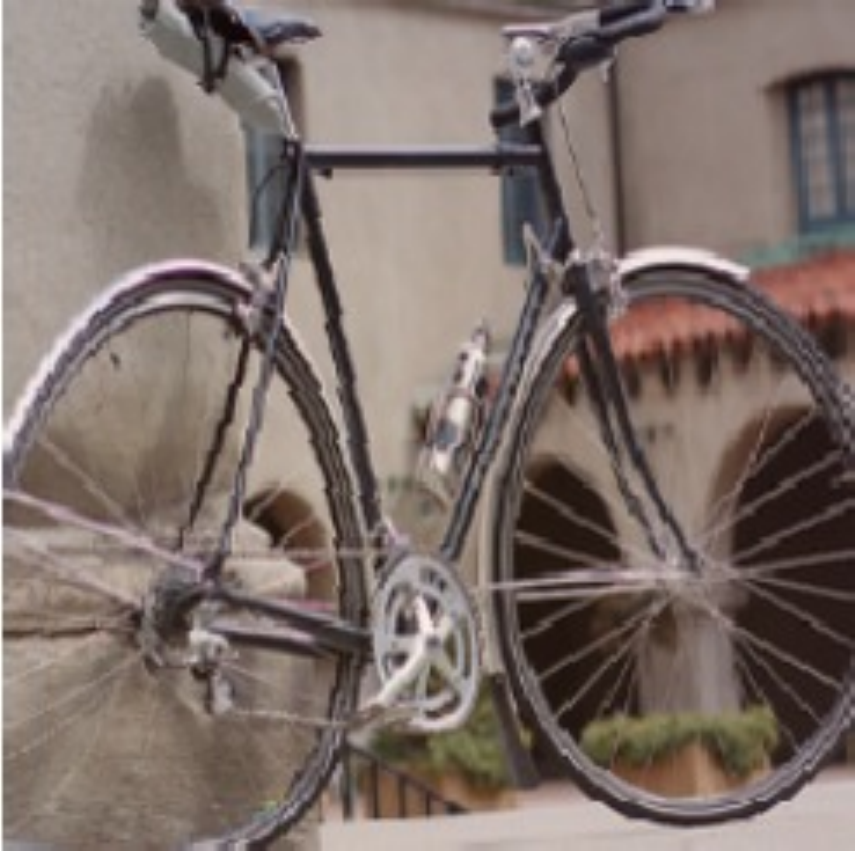} &
\includegraphics[width=0.16\textwidth]{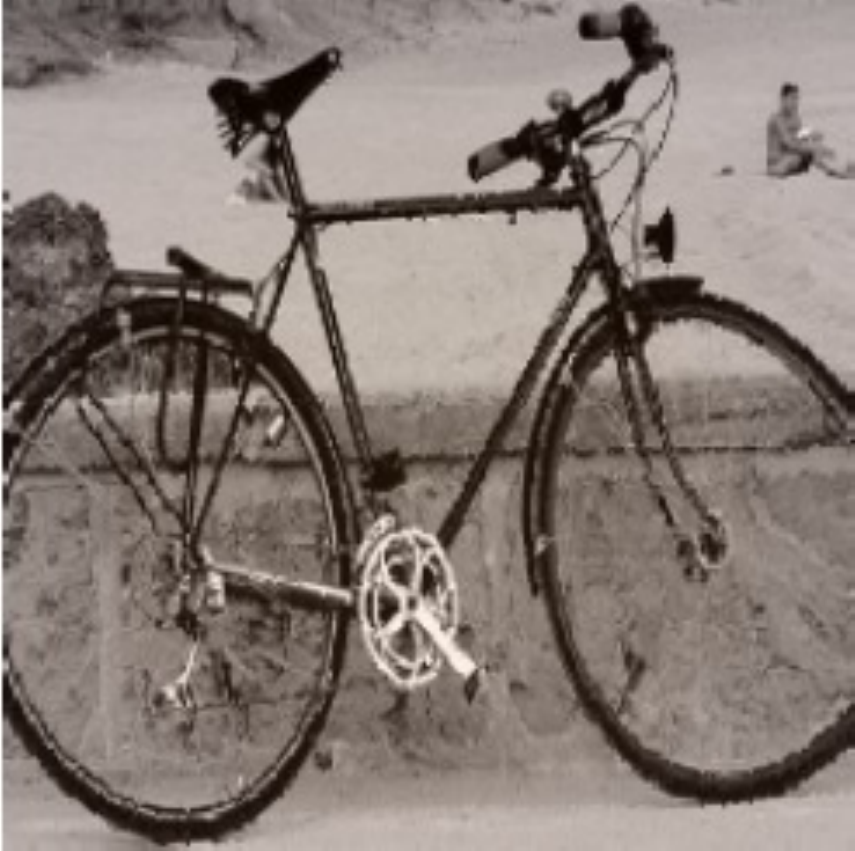}
\end{tabular}
}
\caption{
Convnet features can bring different instances of the same class into good
alignment at least as well (on average) as traditional features.
%For each target image (left column), we show five nearest neighbor images (first
%row), warped versions aligned by \texttt{conv}4 flow (second row), and warped
%versions aligned with SIFT flow \cite{sift-flow} (third row).
For each target image (left column), we show warped versions of five nearest
neighbor images aligned with \texttt{conv}4 flow (first row), and warped
versions aligned with SIFT flow \cite{sift-flow} (second row).
Keypoints from the warped images are shown copied to the target image.
The cat shows a case where convnet features perform better, while the bicycle shows
a case where SIFT features perform better.
(Note that each instance is warped to a square bounding box before alignment.
Best viewed in color.)
\trevor{Mention flipping? or flip in figure? Otherwise the second on looks like an error.}}
\label{fig:align}
\end{figure}

Given the alignment field $w$, we warp target to source using bivariate spline
interpolation (implemented in SciPy \cite{scipy}).
Figure \ref{fig:align} gives examples of alignment quality for a few different
seed images, using both SIFT and convnet features.
We show five warped nearest neighbors as well as keypoints transferred
from those neighbors.

We quantitatively assess the alignment by measuring the accuracy of predicted
keypoints. To obtain good predictions, we warp 25 nearest neighbors for each
target image, and order them from smallest to greatest deformation energy
(we
found this method to outperform ordering using the data term).
We take the predicted keypoints to be the median points (coordinate-wise) of the
top five aligned keypoints according to this ordering.

We assess correctness using mean PCK \cite{PCP}.
We consider a ground truth keypoint to be correctly predicted if the prediction
lies within a Euclidean distance of $\alpha$ times the maximum of the
bounding box width and height, picking some $\alpha \in [0, 1]$.
We compute the overall accuracy for each type of keypoint,
and report the average over keypoint types.
We do not penalize predicted keypoints that are not visible
in the target image.

Results are given in Table \ref{tab:align}. We show per category results using
$\alpha = 0.1$, and mean results for $\alpha = 0.1$, $0.05$, and $0.025$. 
Indeed, convnet learned features are at least as capable as SIFT at
alignment, and better than might have been expected 
given the size of their receptive fields.

\begin{table}
\centering
\caption{
Keypoint transfer accuracy using convnet flow, SIFT flow, and simple copying
from nearest neighbors. Accuracy (PCK) is shown per category using $\alpha = 0.1$ (see
text) and means are also shown for the stricter values $\alpha = 0.05$ and
$0.025$. On average, convnet flow performs as well as SIFT flow, and performs a
bit better for stricter tolerances.
}
\scalebox{0.75}{
\renewcommand{\tabcolsep}{2pt}
\begin{tabular}{rccccccccccccccccccccc}
& aero & bike & bird & boat & bttl & bus & car & cat & chair & cow & table & dog & horse & mbike & prsn & plant & sheep & sofa & train & tv & mean \\
\hline
\texttt{conv}4 flow
& 28.2 & 34.1 & 20.4 & 17.1 & 50.6 & 36.7 & 20.9 & 19.6 & 15.7 & 25.4 & 12.7 & 18.7 & 25.9 & 23.1 & 21.4 & 40.2 & 21.1 & 14.5 & 18.3 & 33.3 & 24.9 \\
SIFT flow
& 27.6 & 30.8 & 19.9 & 17.5 & 49.4 & 36.4 & 20.7 & 16.0 & 16.1 & 25.0 & 16.1 & 16.3 & 27.7 & 28.3 & 20.2 & 36.4 & 20.5 & 17.2 & 19.9 & 32.9 & 24.7 \\
NN transfer
& 18.3 & 24.8 & 14.5 & 15.4 & 48.1 & 27.6 & 16.0 & 11.1 & 12.0 & 16.8 & 15.7 & 12.7 & 20.2 & 18.5 & 18.7 & 33.4 & 14.0 & 15.5 & 14.6 & 30.0 & 19.9
\end{tabular}
}
%\vspace{2em}
\scalebox{0.8}{
\begin{tabular}{rccc}
mean & $\alpha = 0.1$ & $\alpha = 0.05$ & $\alpha = 0.025$ \\
\hline
\texttt{conv}4 flow & 24.9 & 11.8 & 4.08 \\
SIFT flow & 24.7 & 10.9 & 3.55 \\
NN transfer & 19.9 & 7.8 & 2.35 
\end{tabular}
}
\label{tab:align}
\end{table}

\section{Keypoint classification}

In this section, we specifically address the ability of convnet features to
understand semantic information at the scale of parts.
As an initial test, we consider the task of \emph{keypoint classification}:
given an image and the coordinates of a keypoint on that image, can we train a classifier to label
the keypoint?

\begin{figure}[t]
\centering
\begin{minipage}{0.4\textwidth}
\centering
\begin{subfigure}{0.45\textwidth}
    \includegraphics[width=\textwidth]{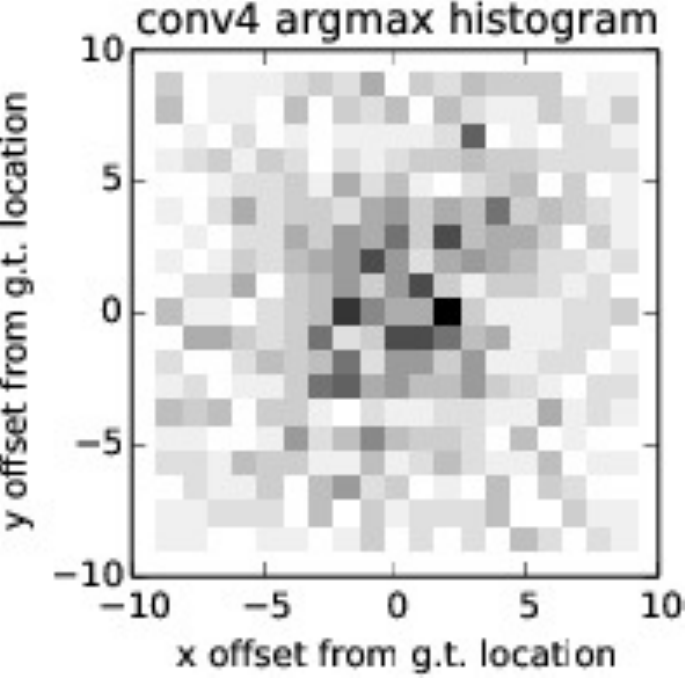}

\includegraphics[width=\textwidth]{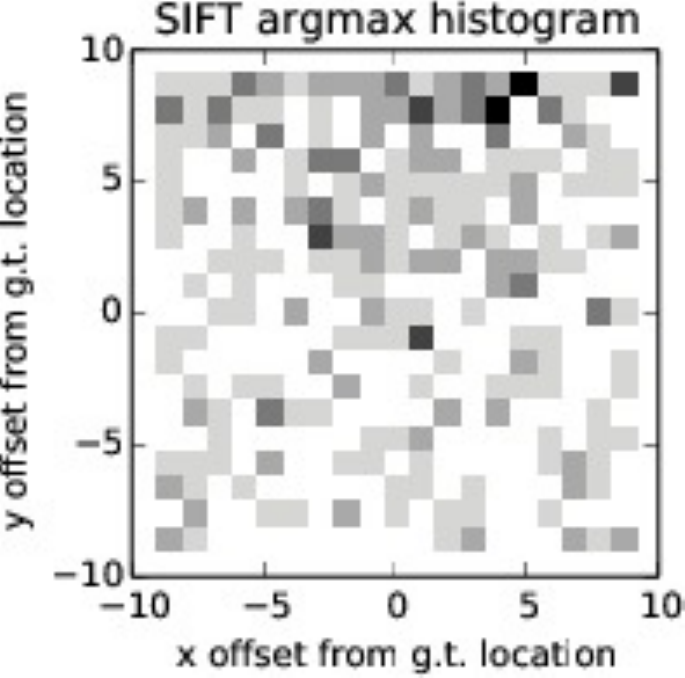}
\caption{cat left eye}
\end{subfigure}
\hspace{1em}
\begin{subfigure}{0.45\textwidth}
\includegraphics[width=\textwidth]{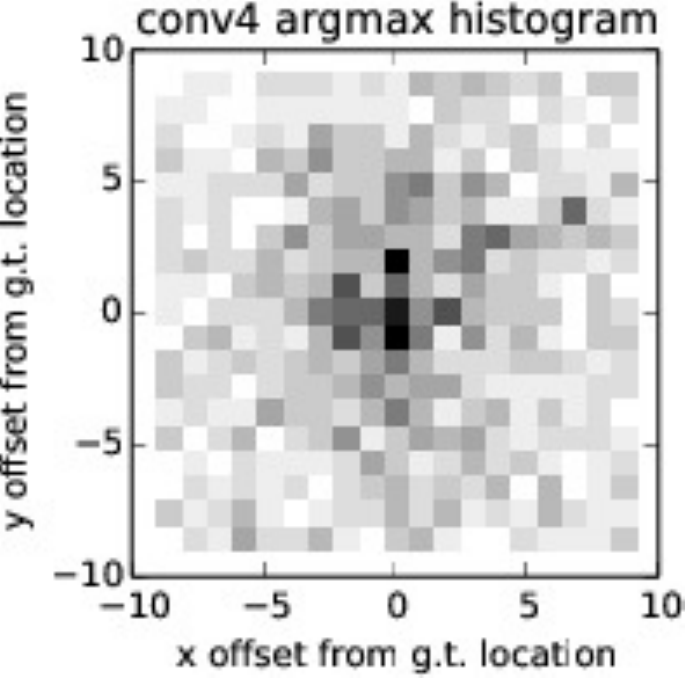}

\includegraphics[width=\textwidth]{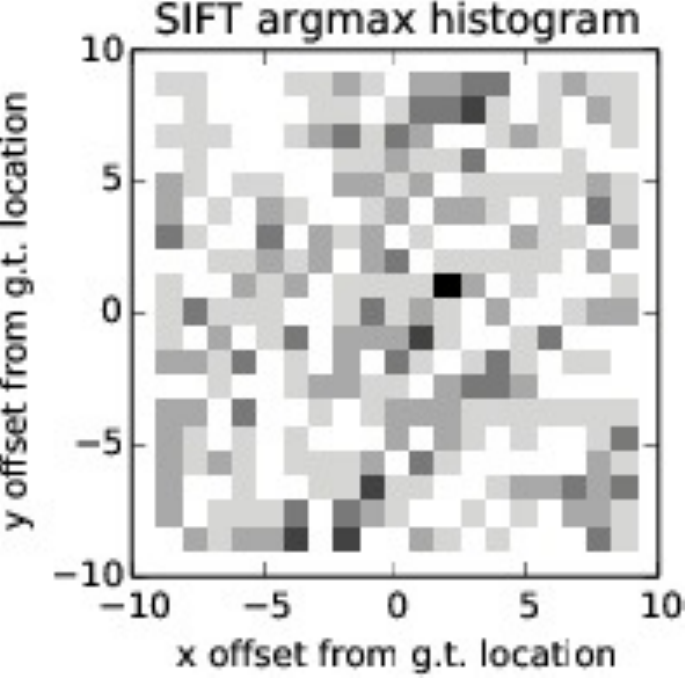}
\caption{cat nose}
\end{subfigure}
\captionof{figure}{
Convnet features show fine localization ability, even beyond their stride and in
cases where SIFT features do not perform as well.
Each plot is a 2D histogram of the locations of the maximum responses of a
classifer in a 21 by 21 pixel rectangle taken around a ground truth keypoint.
}
\label{fig:precision}
\end{minipage}
\hspace{2em}
\begin{minipage}{0.4\textwidth}
\centering
\begin{subfigure}{0.45\textwidth}
\includegraphics[width=\textwidth]{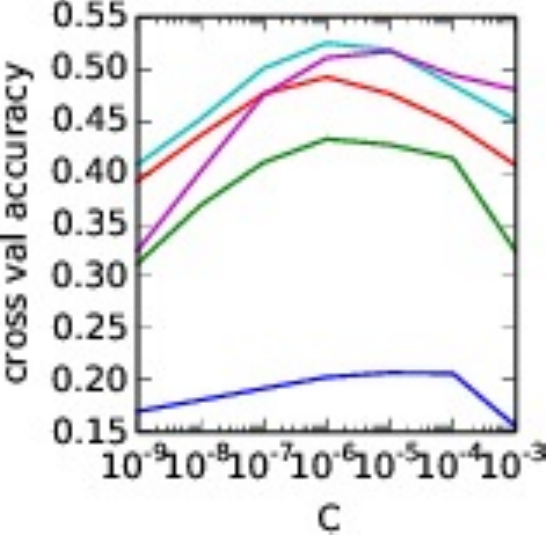}
\caption{}
\end{subfigure}
\begin{subfigure}{0.45\textwidth}
\includegraphics[width=\textwidth]{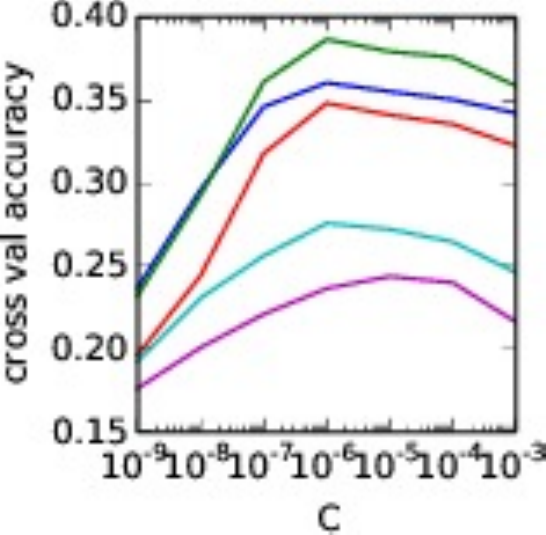}
\caption{}
\end{subfigure}
\captionof{figure}{
Cross validation scores for cat keypoint classification as a function of the
SVM parameter $C$.
In (a), we plot mean accuracy against $C$ for five different convnet features; in (b) we plot
the same for SIFT features of different sizes.
We use $C = 10^{-6}$ for all experiments in Table \ref{tab:class}.
}
\label{fig:cv}
\end{minipage}
\end{figure}

For this task we use keypoint data \cite{BourdevMalikICCV09} on the twenty classes
of PASCAL VOC 2011 \cite{pascal}.
We extract features at each keypoint using SIFT \cite{SIFT} and using the column
of each convnet layer whose rf center lies closest to the keypoint. (Note that the
SIFT  features will be more precisely placed as a result of this approximation.)
We trained one-vs-all linear SVMs on the train set using SIFT at five different
radii and each of the five convolutional layer activations
as features (in general, we found pooling and normalization layers to have lower
performance).
We set the SVM parameter $C = 10^{-6}$ for all experiments based on five-fold cross validation on
the training set (see Figure \ref{fig:cv}).

Table \ref{tab:class} gives the resulting accuracies on the val set.
We find features from convnet layers consistently perform at least as well as
and often better than SIFT at this task, with the highest performance coming
from layers \texttt{conv}4 and \texttt{conv}5. 
Note that we are specifically testing convnet features trained only for
classification; the same net could be expected to achieve even higher performance
if trained for this task.

\begin{table}
\centering
\caption{
Keypoint classification accuracies, in percent, on the twenty categories of
PASCAL 2011 val, trained with SIFT or convnet features. The best SIFT
and convnet scores are bolded in each category.
}
\renewcommand{\tabcolsep}{2pt}
\scalebox{0.8}{
\begin{tabular}{rrccccccccccccccccccccc}
& & aero & bike & bird & boat & bttl & bus & car & cat & chair & cow & table & dog
& horse & mbike & prsn & plant & sheep & sofa & train & tv & mean \\
\hline
SIFT & 10 & 36 & 42 & 36 & 32 & 67 & 64 & 40 & 37 & 33 & 37 & 60 & 34 & 39 & 38 & 29 & 63
 & 37 & 42 & 64 & 75 & 45 \\
(radius) & 20 & \textbf{37} & 50 & \textbf{39} & 35 & 74 & 67 & \textbf{47} & \textbf{40} & 36
& \textbf{43} & 68 & \textbf{38} & \textbf{42} & 48 & \textbf{33} & \textbf{70}
& \textbf{44} & \textbf{52} & 68 & 77 & 50 \\
& 40 & 35 & \textbf{54} & 37 & 41 & \textbf{76} & \textbf{68} & \textbf{47} & 37 & 39
& 40 & 69 & 36 & \textbf{42} & 49 & 32 & 69 & 39 & \textbf{52} & \textbf{74} &
\textbf{78} & \textbf{51} \\
& 80 & 33 & 43 & 37 & \textbf{42} & 75 & 66 & 42 & 30 & \textbf{43} & 36 &
\textbf{70} & 31 & 36 & \textbf{51} & 27 & \textbf{70} & 35 & 49 & 69 & 77 & 48 \\
& 160 & 27 & 36 & 34 & 38 & 72 & 59 & 35 & 25 & 39 & 30 & 67 & 27 & 32 & 46 & 25 &
\textbf{70} & 29 & 48 & 66 & 76 & 44 \\
\hline
\texttt{conv} & 1 & 16 & 14 & 15 & 19 & 20 & 29 & 15 & 22 & 16 & 17 & 29 & 17 & 14 & 16 & 15 & 33
& 18 & 12 & 27 & 29 & 20 \\
(layer) & 2 & 37 & 43 & 40 & 35 & 69 & 63 & 38 & 44 & 35 & 40 & 61 & 38 & 40 & 44 & 34 & 65
& 39 & 41 & 63 & 72 & 47 \\
& 3 & 42 & 50 & 46 & 41 & 76 & 69 & \textbf{46} & 52 & 39 & 45 & 64 & 47 & 48 & 52 &
40 & 74 & 46 & 50 & 71 & \textbf{77} & 54 \\
& 4 & \textbf{44} & \textbf{53} & \textbf{49} & \textbf{42} & \textbf{78} &
\textbf{70} & 45 & \textbf{55} & \textbf{41} & \textbf{48} & \textbf{68} &
\textbf{51} & \textbf{51} & \textbf{53} & \textbf{41} & \textbf{76} &
\textbf{49} & \textbf{52} & \textbf{73} & 76 & \textbf{56} \\
& 5 & \textbf{44} & 51 & \textbf{49} & 41 & 77 & 68 & 44 & 53 & 39 & 45 & 63 & 50 &
49 & 52 & 39 & 73 & 47 & 47 & 71 & 75 & 54 \\
\end{tabular}
}
\label{tab:class}
\end{table}

Finally, we study the precise location understanding of our classifiers
by computing their responses with a single-pixel stride around
ground truth keypoint locations.
For two example keypoints (cat left eye and nose), we histogram the locations of
the maximum responses within a 21 pixel by 21 pixel rectangle around the
keypoint, shown in Figure \ref{fig:precision}.
We do not include maximum responses that lie on the boundary of this rectangle.
While the SIFT classifiers do not seem to be sensitive to the precise locations of
the keypoints, in many cases the convnet ones seem to be capable of localization
finer than their strides, not just their receptive field sizes. This observation 
motivates our final experiments to consider detection-based localization performance.

\section{Keypoint prediction}
\label{sec:pred}

We have seen that 
despite their large receptive field sizes, convnets work as well as the hand-engineered
feature SIFT for alignment and slightly better than SIFT for
keypoint classification. Keypoint prediction provides a natural follow-up test.
As in Section \ref{sec:flow}, we use keypoint annotations from PASCAL VOC 2011,
and we assume a ground truth bounding box.

%Suppose each keypoint is represented by $X = (x,y,v)$ where $x$, $y$ 
%are the coordinates and $v \in \{0,~1\}$ is the visible state of the keypoint.
% We refer $X$ as one part vector. Each image contains $P$ parts and all keypoints 
%are represented as $\mathbf{X} = \{X_1,\ldots, X_k, \ldots, X_P\}$.  A labeled image is represented as $(I, \mathbf{X})$, where $I$ stands for the image feature and
 %we will use both \texttt{conv}5 and SIFT for the image representation $I$ to compare the localization properties.

Inspired in part by \cite{RossJeff, overfeat, HumanPoseICLR}, we train sliding
window part detectors to predict keypoint locations independently.  R-CNN
\cite{RossJeff} and OverFeat \cite{overfeat} have both demonstrated the
effectiveness of deep convolutional networks on the generic object detection
task. However, neither of them have investigated the application of CNNs for
keypoint prediction.\footnote{But see works cited in Section 1.1 regarding
keypoint localization.} R-CNN starts from bottom-up region proposal
\cite{selsearch}, which tends to overlook the signal from small parts. OverFeat, on
the other hand, combines convnets trained for classification and
for regression and runs in multi-scale sliding window fashion.

We rescale each bounding box to $500 \times 500$ and compute \texttt{conv}5
(with a stride of 16 pixels).
Each cell of
\texttt{conv}5 contains one $256$-dimensional descriptor.
We concatenate \texttt{conv}5 descriptors from a local region of $3 \times 3$ 
cells, giving an overall receptive field size of $195 \times 195$ and feature
dimension of $2304$.
For each
keypoint, we train a linear SVM with hard negative mining.
We consider the ten closest features to each ground truth keypoint as positive
examples, and all the features whose rfs do not contain the keypoint as negative
examples.
We also train using dense SIFT descriptors for comparison.
We compute SIFT on a grid of stride eight
 and bin size of eight using VLFeat
\cite{vedaldi08vlfeat}.
For SIFT, we consider features within twice the bin
size from the ground truth keypoint to be positives,
while samples that are at least four times the bin size away are negatives.

We augment our SVM detectors with a spherical Gaussian prior over candidate locations
constructed by nearest neighbor matching.
The mean of each Gaussian is taken to be the location of the keypoint
in the nearest neighbor in the training set found using cosine similarity
on \texttt{pool}5 features, and we use a fixed standard deviation of 22 pixels.
Let $s(X_i)$ be the output score of our local detector for
keypoint $X_i$, and let $p(X_i)$ be the prior score.
We combine these to yield a final score $f(X_i) = s(X_i)^{1 - \eta}
p(X_i) ^ \eta$,
where $\eta \in [0,1]$ is a tradeoff parameter.
In our experiments, we set $\eta = 0.1$ by cross validation.
At test time, we predict the keypoint location as the highest scoring candidate
over all feature locations.

%Recent work by Jain et al.\ \cite{HumanPoseICLR} proposes a convolutional
%network architecture for human pose estimation. They train multiple small
%convnets on $64 \times 64$ patches from scratch for independent binary body-part
%detection and apply a chain model on top to enforce pose consistency while we
%use the pretrained ImageNet reference model and share mid-level feature
%representations among all parts. 
%%We don't have to retrain convnets to generalize to new classes.
%We also use a simpler nonparametric prior to
%filter out outliers. 

\begin{figure}[t]
\begin{minipage}{\textwidth}
\centering
\renewcommand{\tabcolsep}{2pt}
\captionof{table}{Keypoint prediction results on PASCAL VOC 2011.
The numbers give average accuracy of keypoint prediction using
the criterion described in Section \ref{sec:flow}, PCK with $\alpha = 0.1$.}
\scalebox{0.75}{
\begin{tabular}{cccccccccccccccccccccc}
& aero & bike & bird & boat & bttl & bus & car & cat & chair & cow & table
& dog & horse & mbike & prsn & plant & sheep & sofa & train & tv & mean \\
\hline
SIFT  & 17.9  &    16.5   &  15.3 &   15.6  &   25.7 &  21.7 &  22.0  &   12.6 &  11.3  &
  7.6  &  6.5 &  12.5 &  18.3 &  15.1  &  15.9  &  21.3  &  14.7  &  15.1 &  9.2 &  19.9  &  15.7 \\
SIFT+prior & 33.5  & 36.9 &  22.7 &  23.1 &   44.0 & 42.6 &   39.3  &  22.1 & 18.5 &  
  23.5 & 11.2 &  20.6 &  32.2 &  33.9 &  26.7 &  30.6 &  25.7 &  26.5 &   21.9 &  32.4 &   28.4\\
\texttt{conv}5 & 38.5 & 37.6 &  29.6 &  25.3  &  54.5 &   52.1 &   28.6  &  31.5  &  8.9 &   30.5 &
24.1 &  23.7 &  35.8 &  29.9 &   39.3 &   38.2 &  30.5 &   24.5 &  41.5 &  42.0 &  33.3 \\
\texttt{conv}5+prior & \textbf{50.9} &   \textbf{48.8} &  \textbf{35.1} &  \textbf{32.5} &   \textbf{66.1} &
  \textbf{62.0}  &  \textbf{45.7} &   \textbf{34.2} &  \textbf{21.4} &  \textbf{41.1} &  \textbf{27.2} &  
  \textbf{29.3} &  \textbf{46.8} &  \textbf{45.6} &  \textbf{47.1} &   \textbf{42.5} &  \textbf{38.8} &  
  \textbf{37.6} &  \textbf{50.7} &  \textbf{45.6}  & \textbf{42.5} \\
\end{tabular}
}
\label{table:keypoint_prediction}
\label{table:keypointprediction}
\end{minipage}
\vspace{1em}

\begin{minipage}{\textwidth}
\centering
\renewcommand{\tabcolsep}{2pt}
\scalebox{0.9}{
\begin{tabular}{ccccccc}
Groundtruth  &  SIFT+prior &\texttt{conv}5+prior& \hspace{1em} & Groundtruth  &  SIFT+prior &\texttt{conv}5+prior \\
\includegraphics[width=0.17\textwidth]{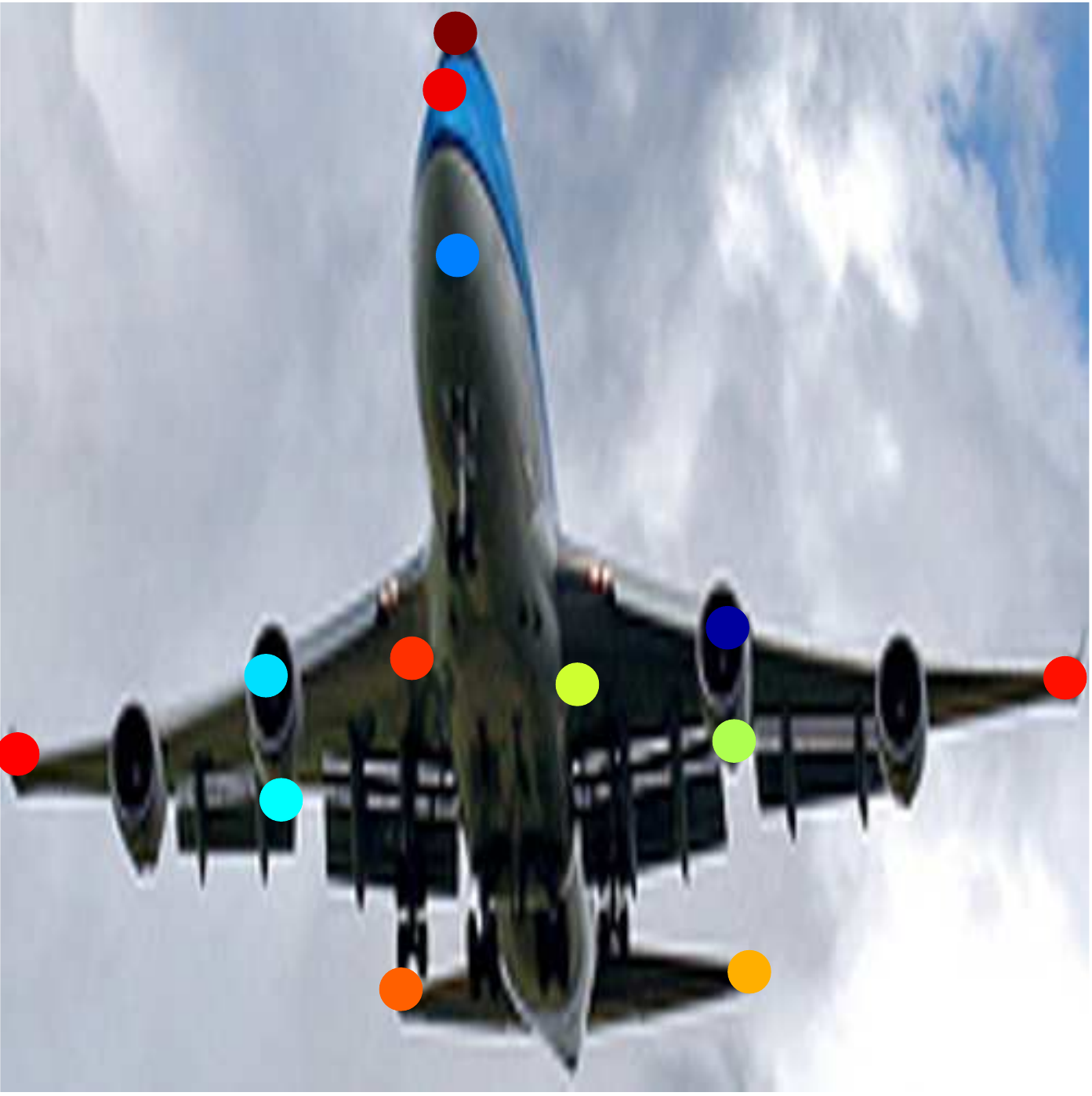} &
\includegraphics[width=0.17\textwidth]{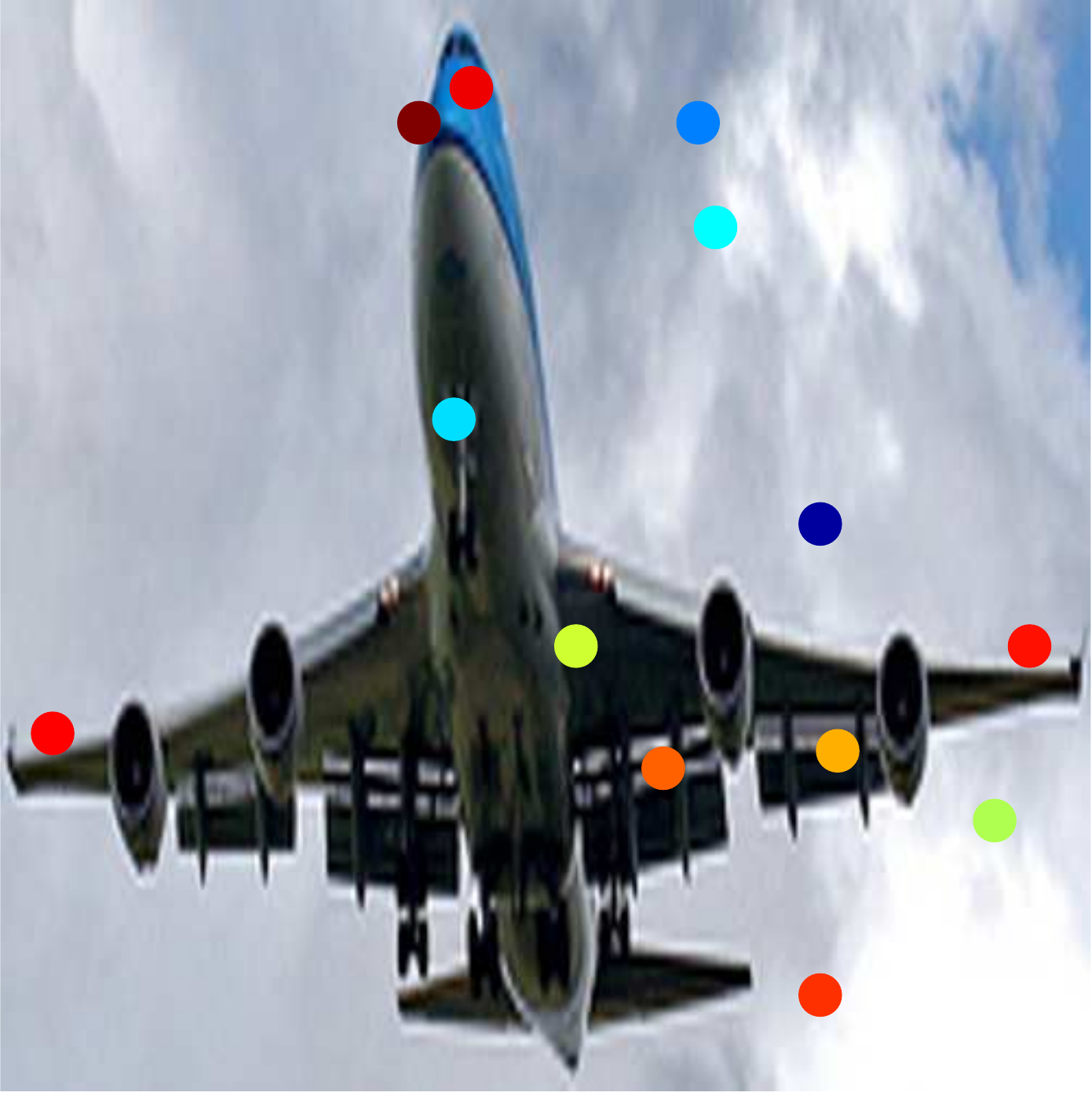} &
\includegraphics[width=0.17\textwidth]{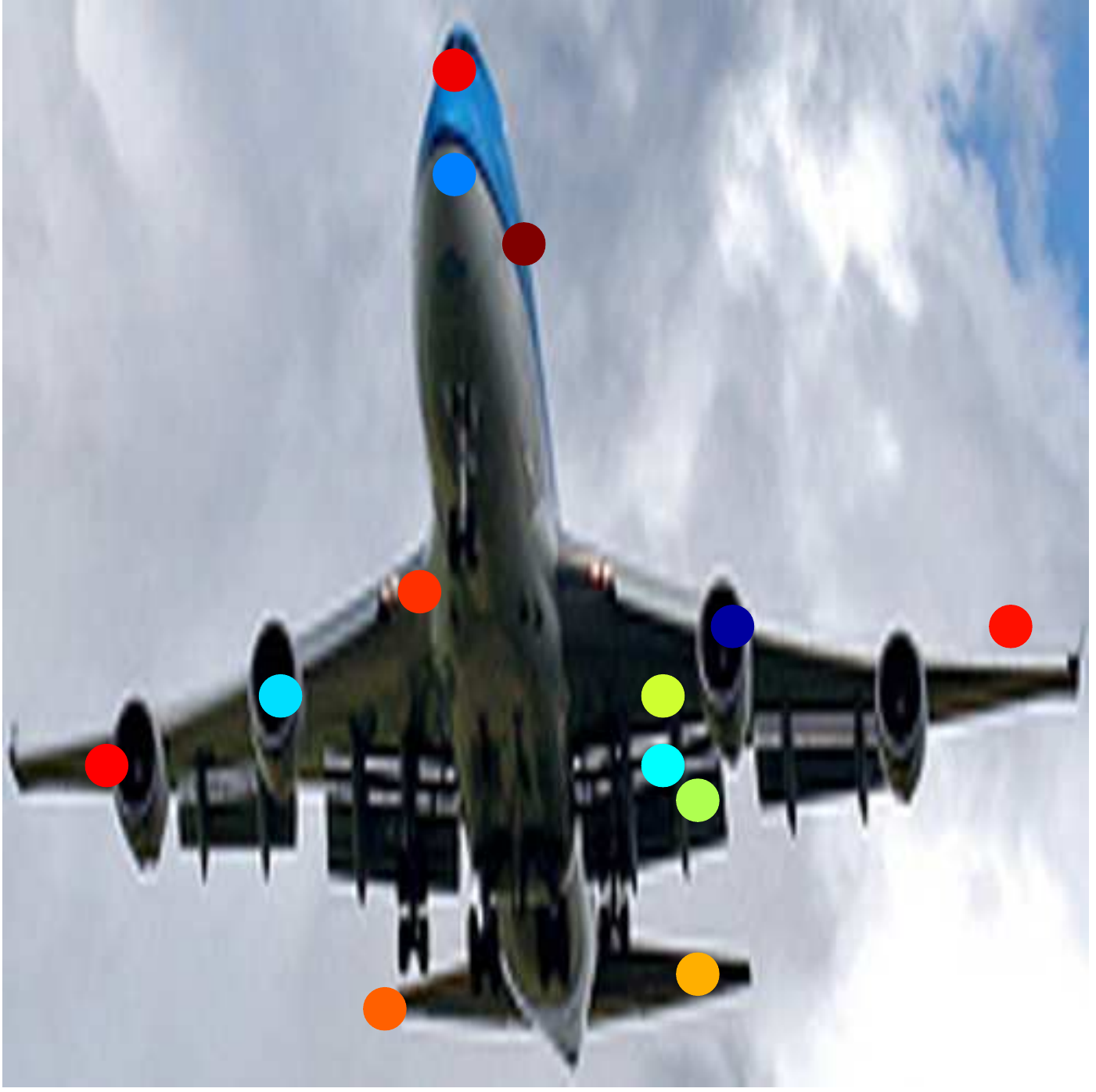} & &
\includegraphics[width=0.17\textwidth]{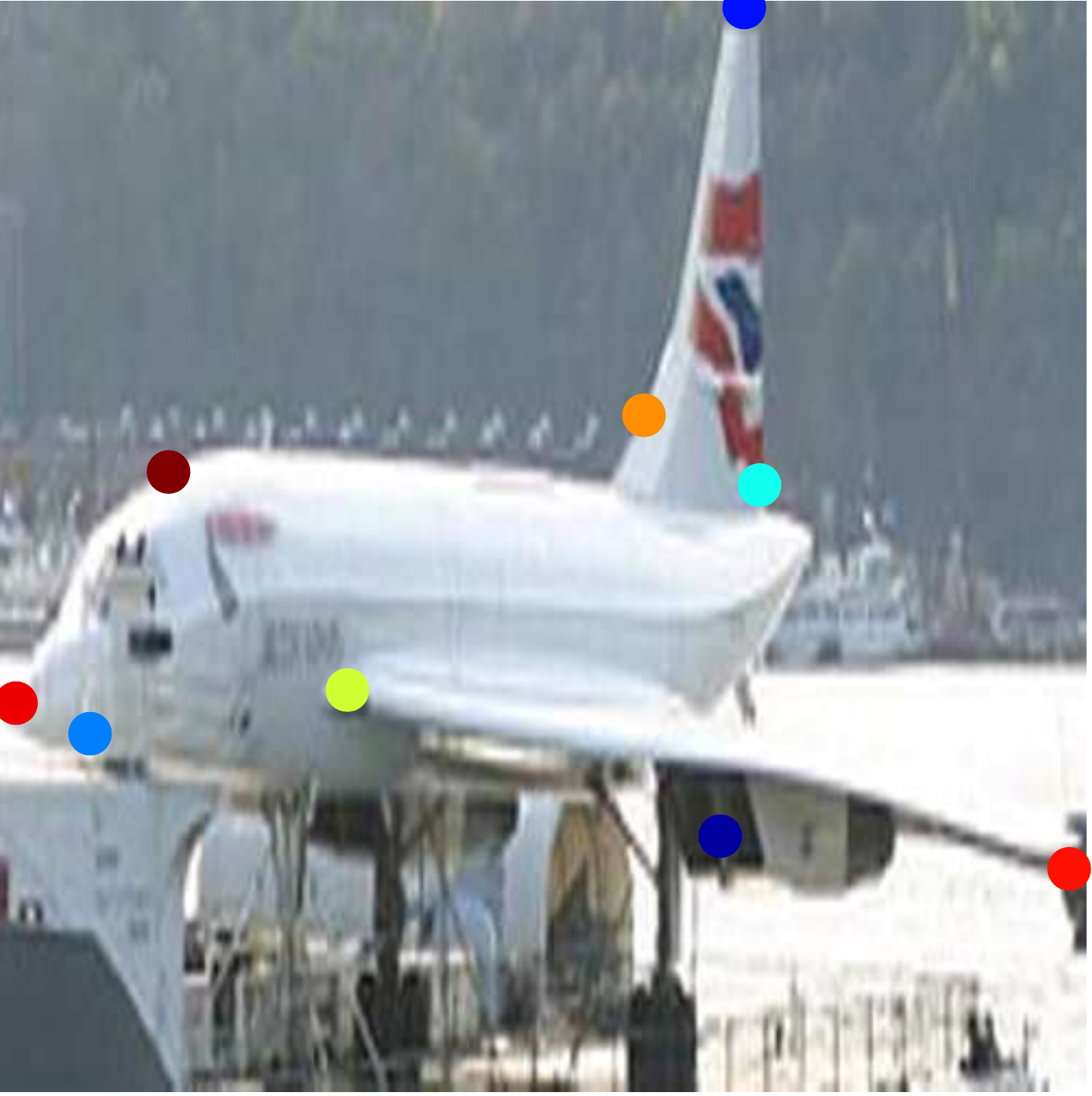} &
\includegraphics[width=0.17\textwidth]{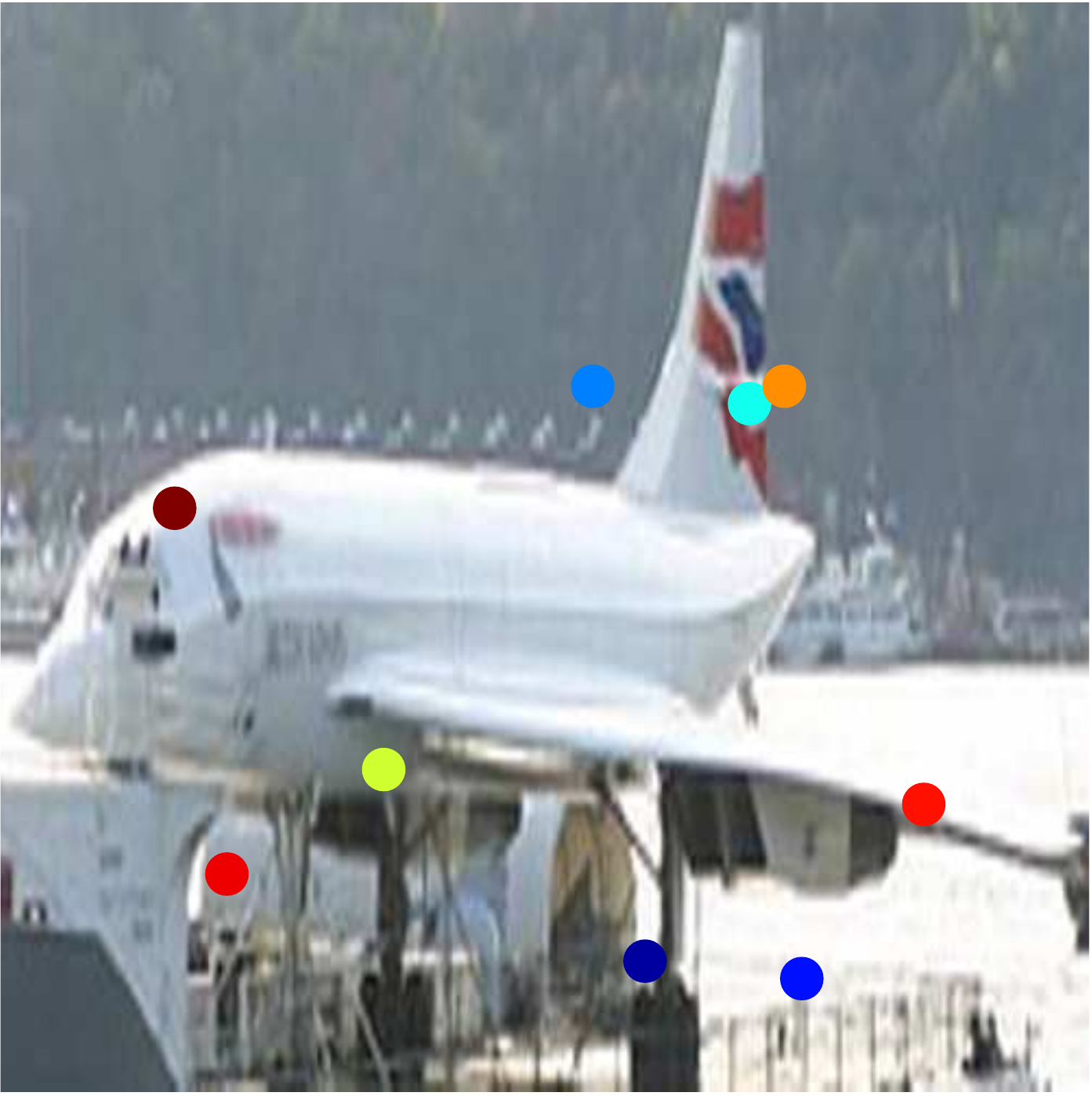} &
\includegraphics[width=0.17\textwidth]{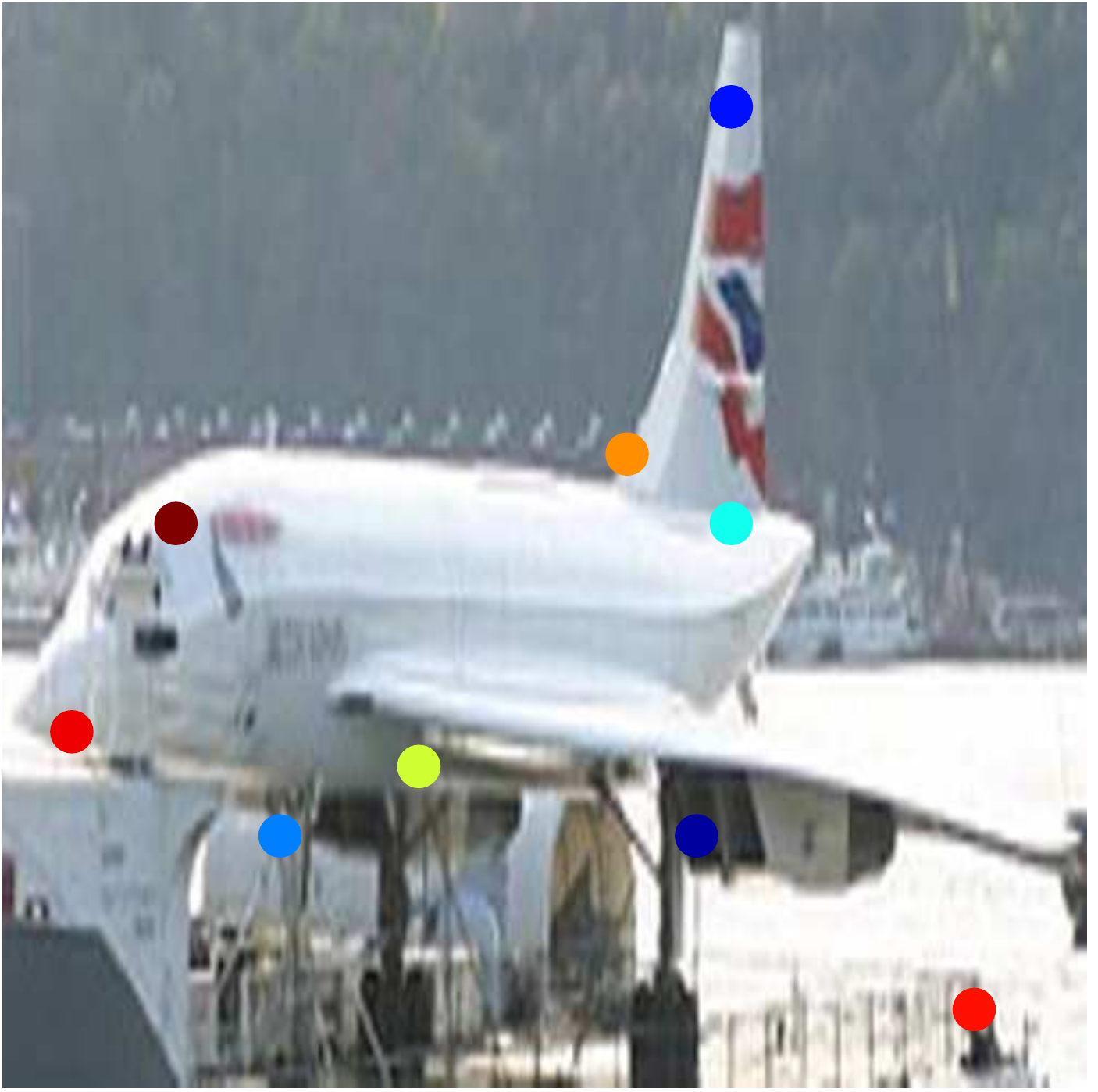} \\
\includegraphics[width=0.17\textwidth]{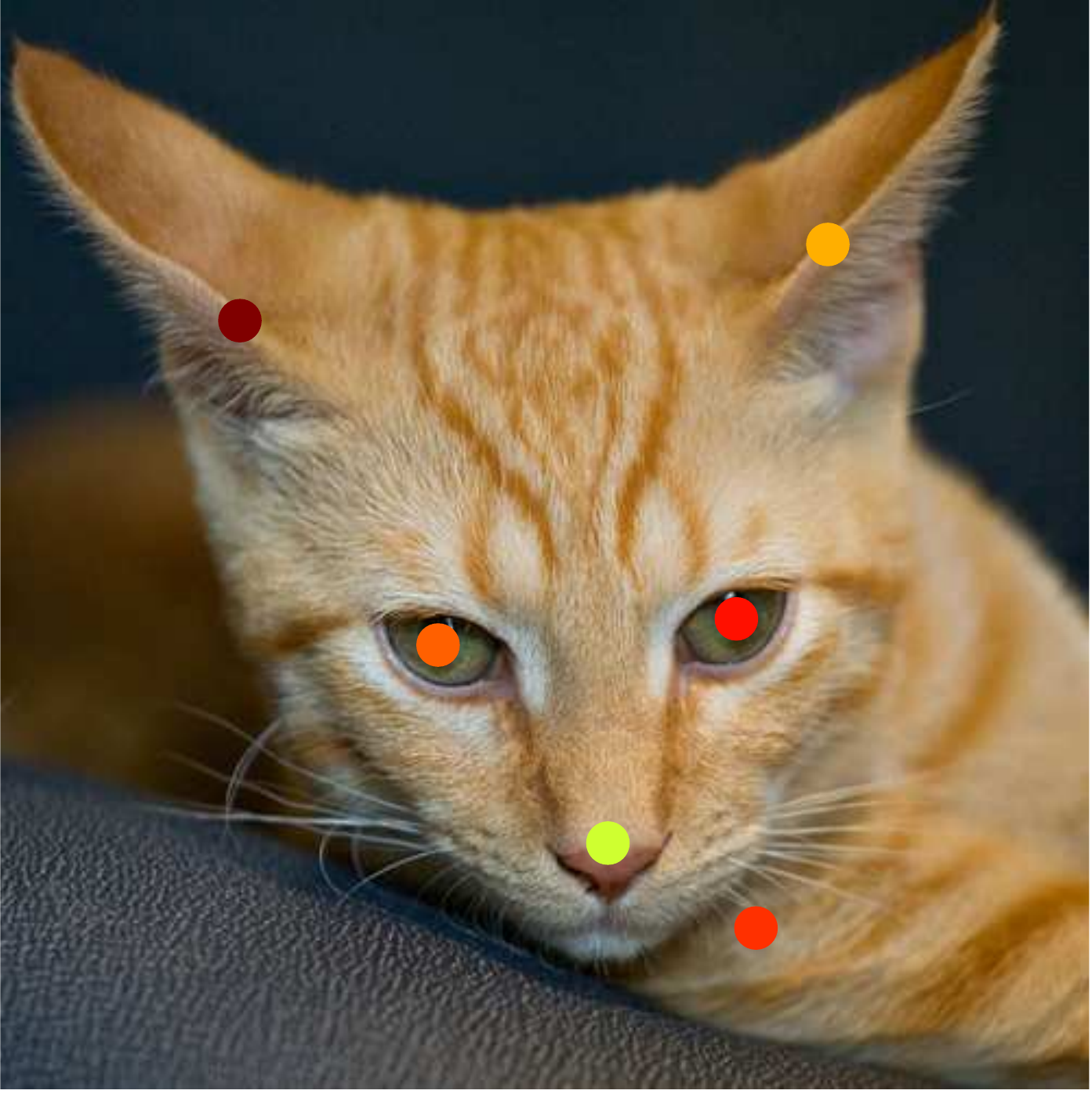} &
\includegraphics[width=0.17\textwidth]{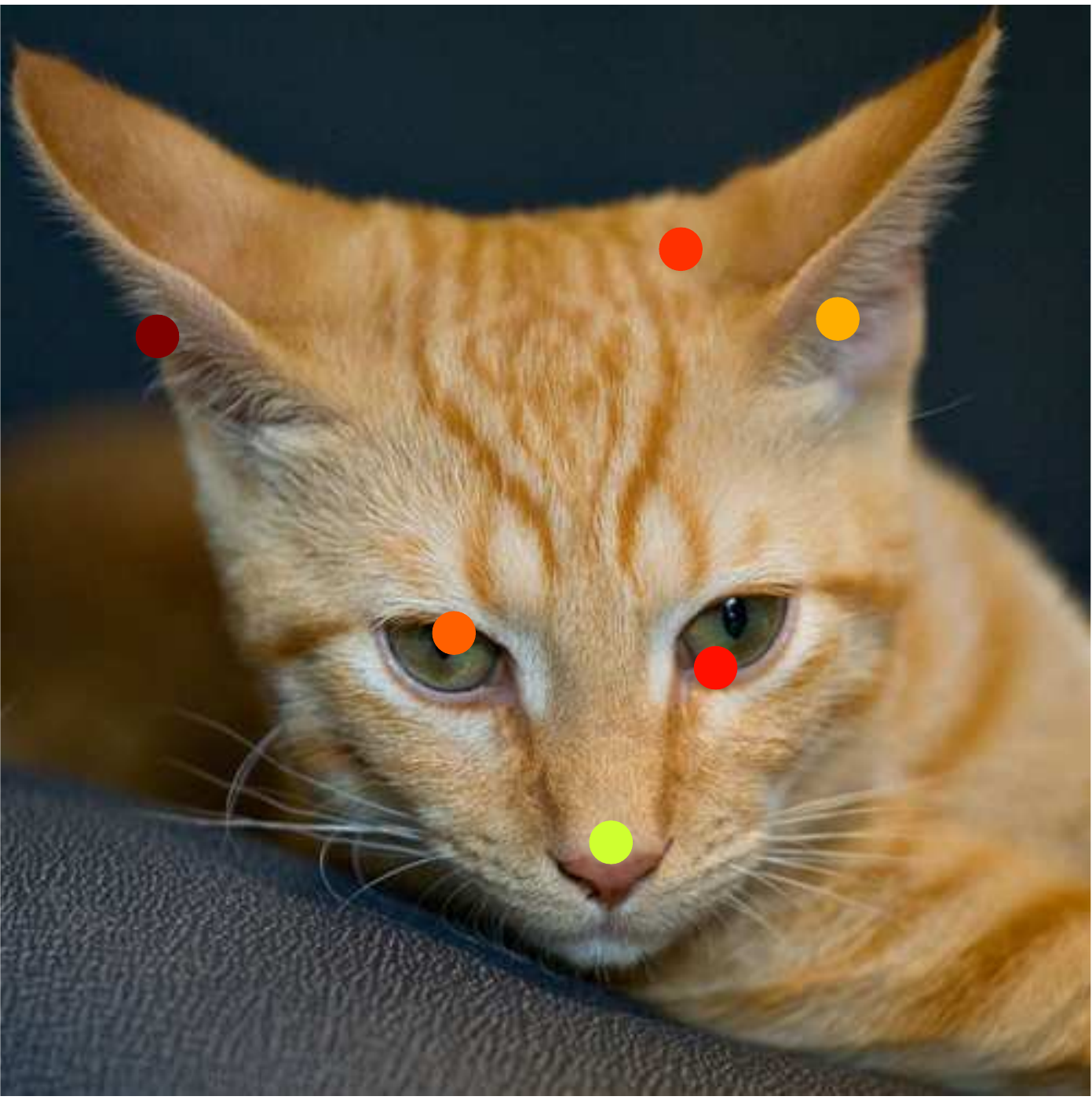} &
\includegraphics[width=0.17\textwidth]{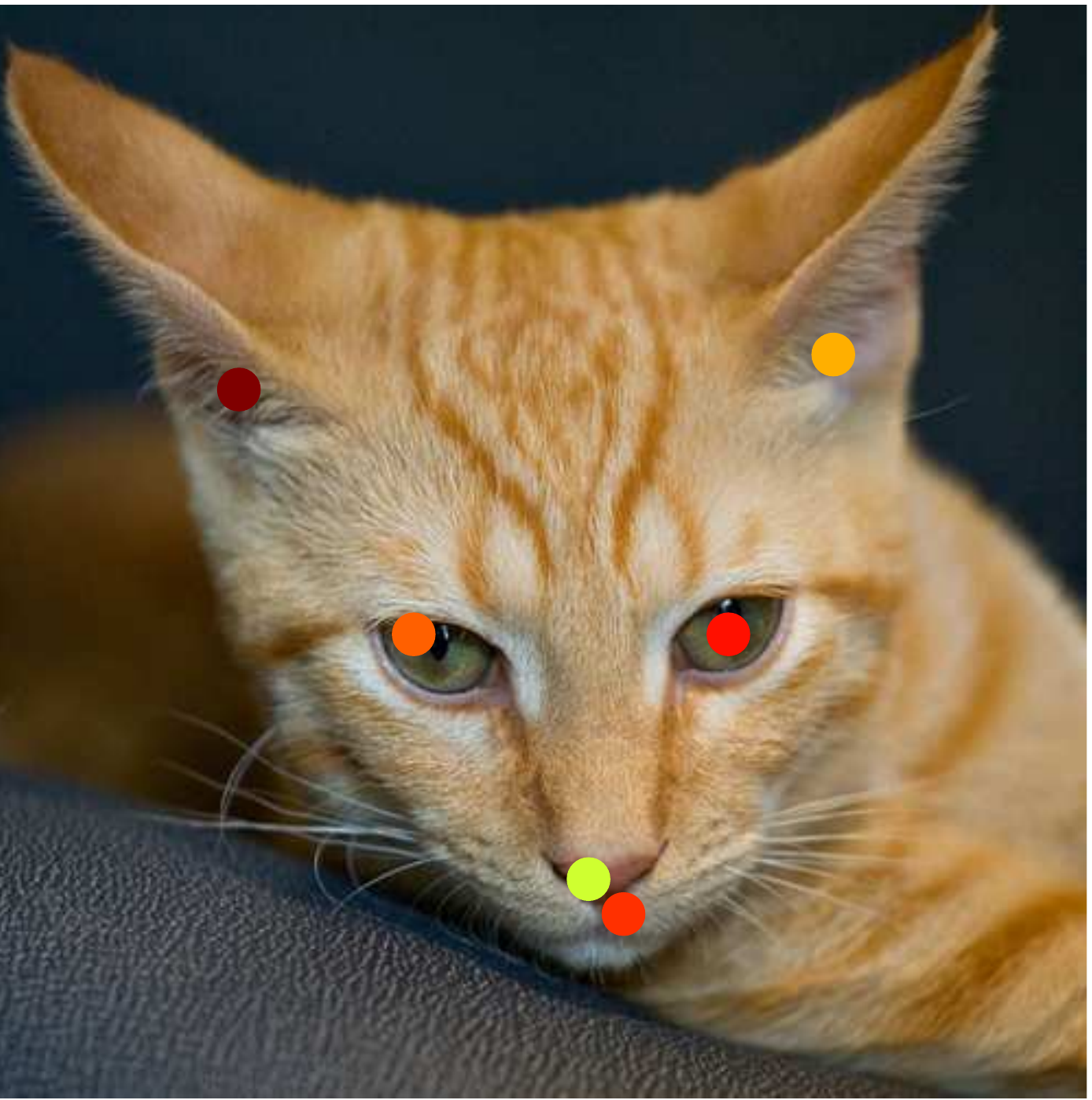} & &
\includegraphics[width=0.17\textwidth]{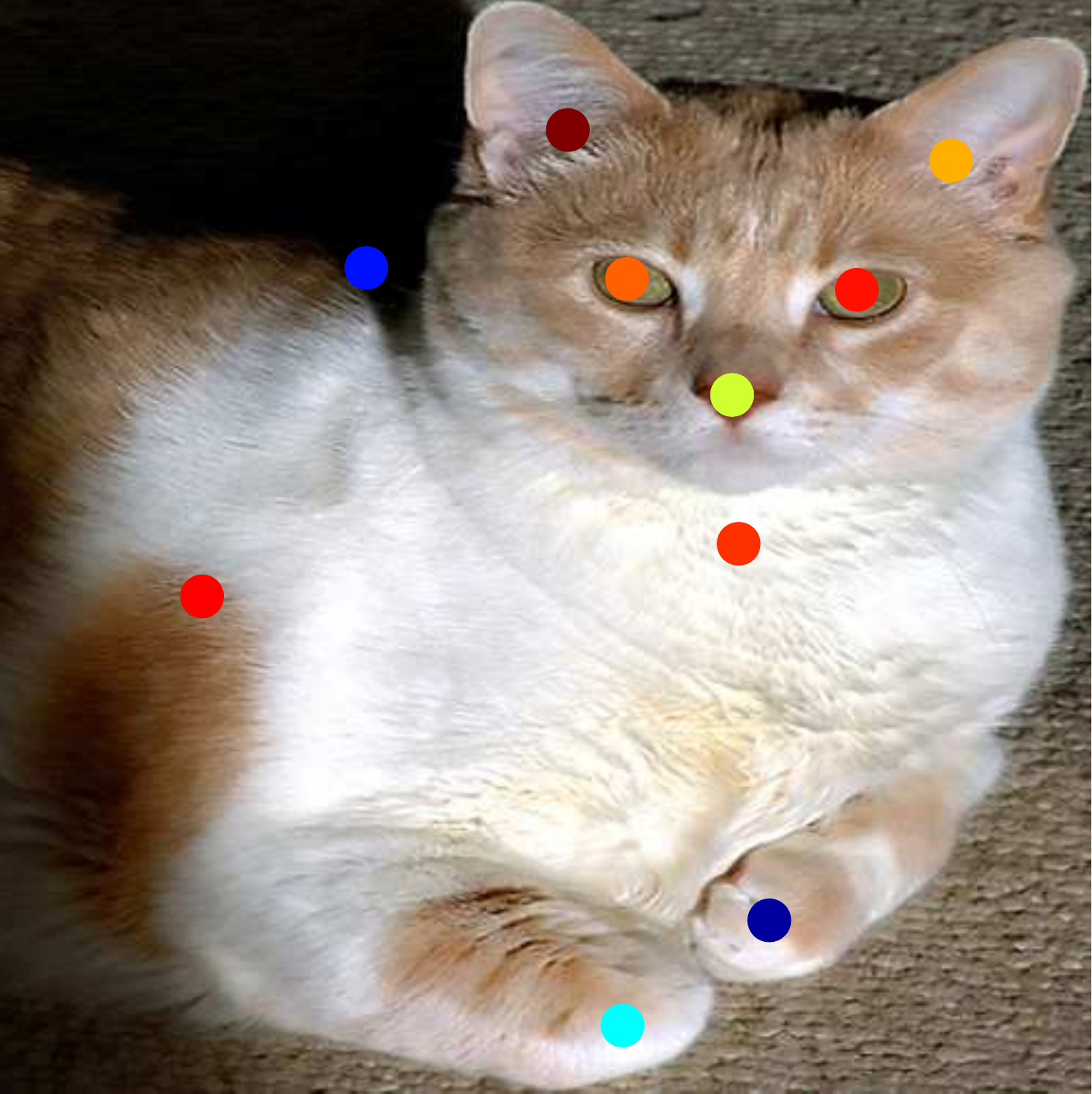} &
\includegraphics[width=0.17\textwidth]{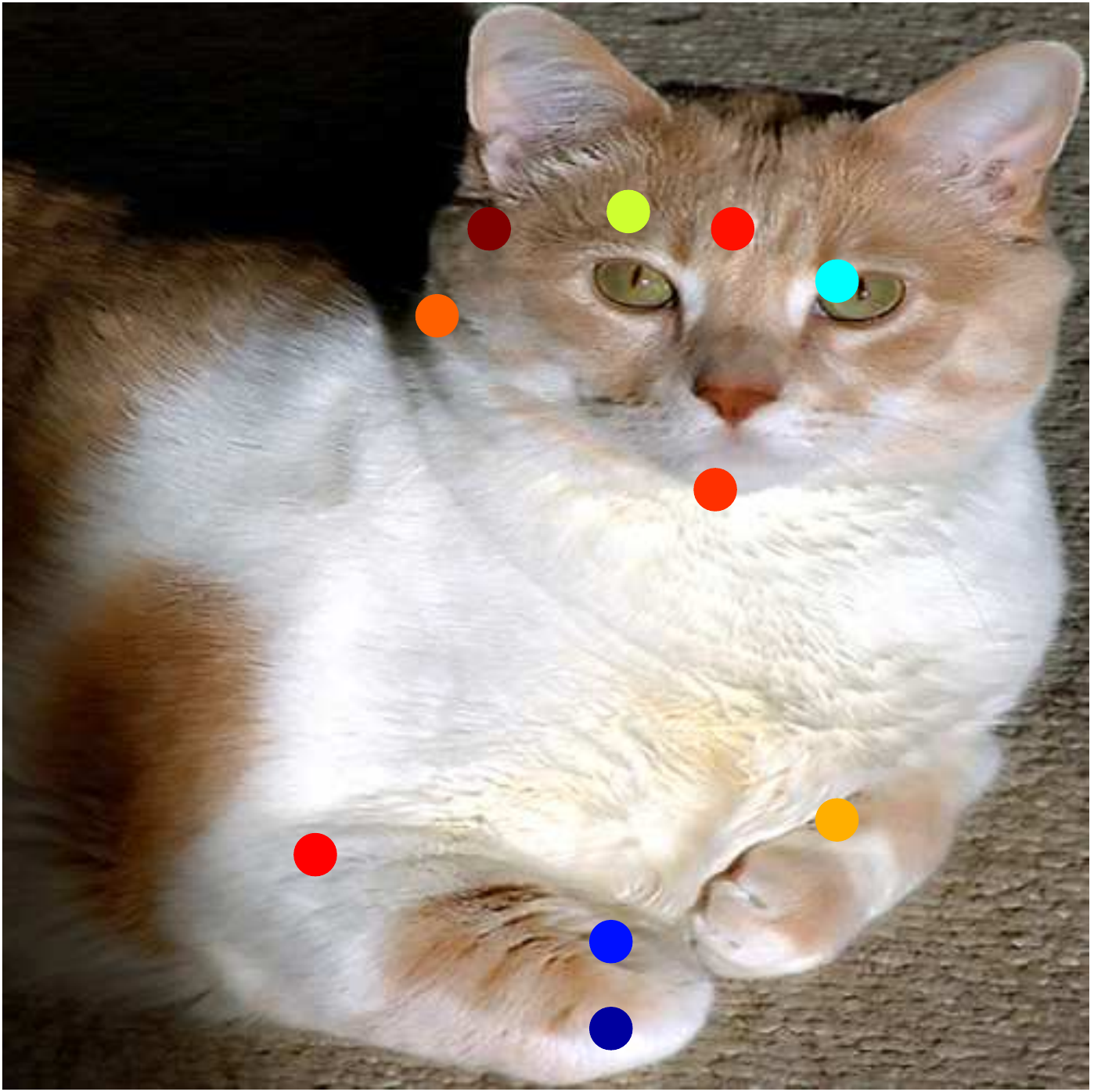} &
\includegraphics[width=0.17\textwidth]{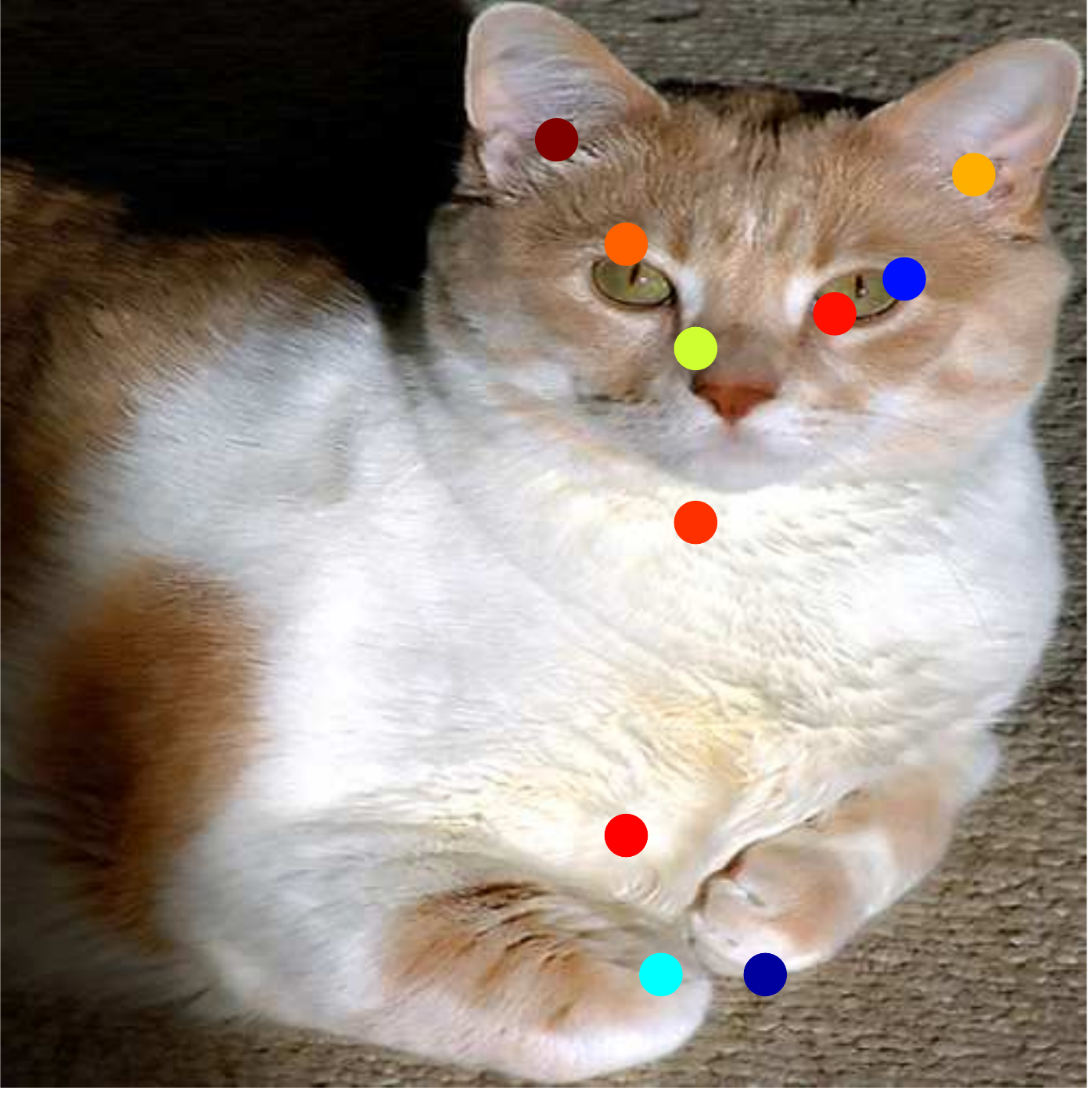} \\
\includegraphics[width=0.17\textwidth]{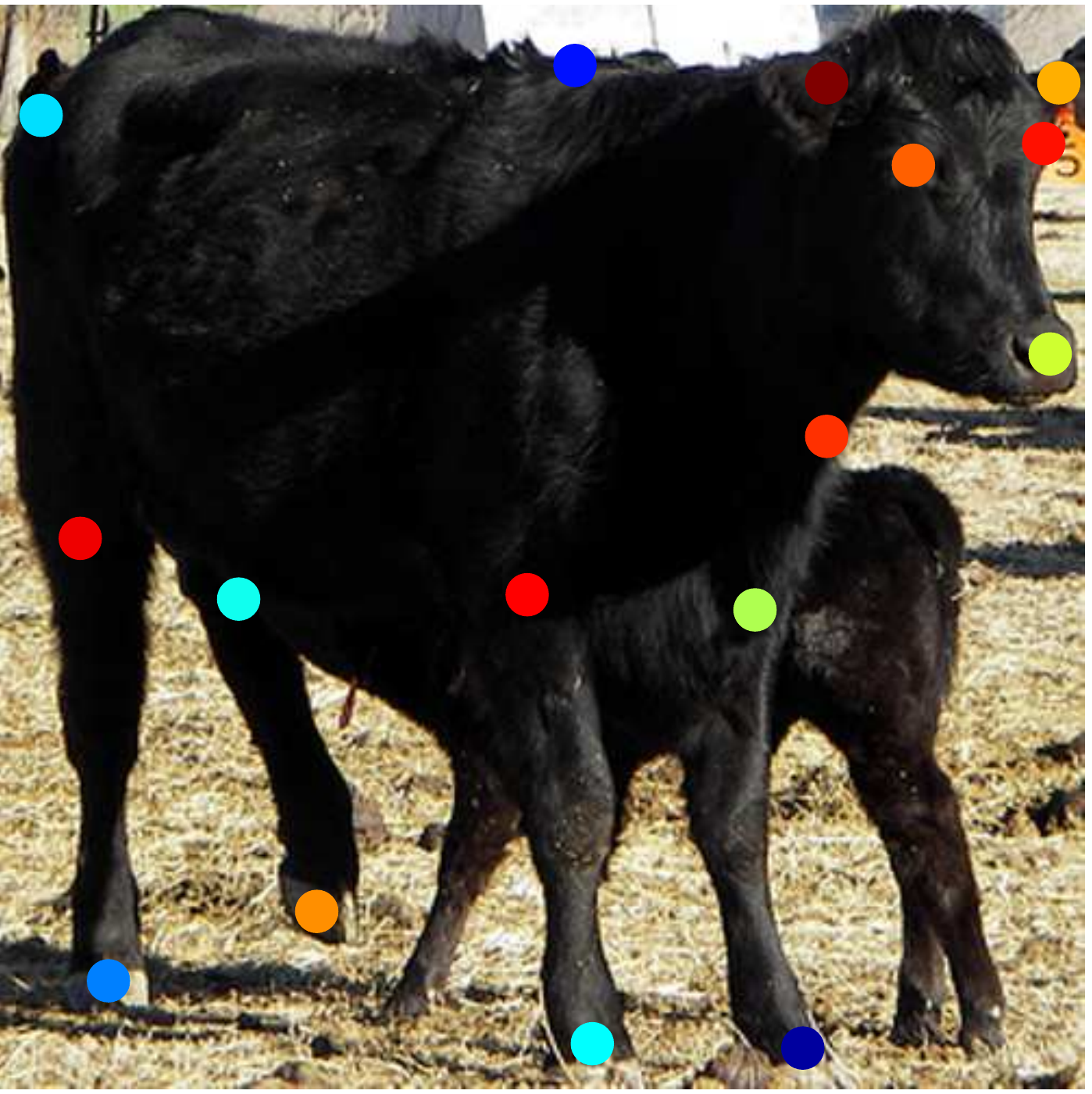} &
\includegraphics[width=0.17\textwidth]{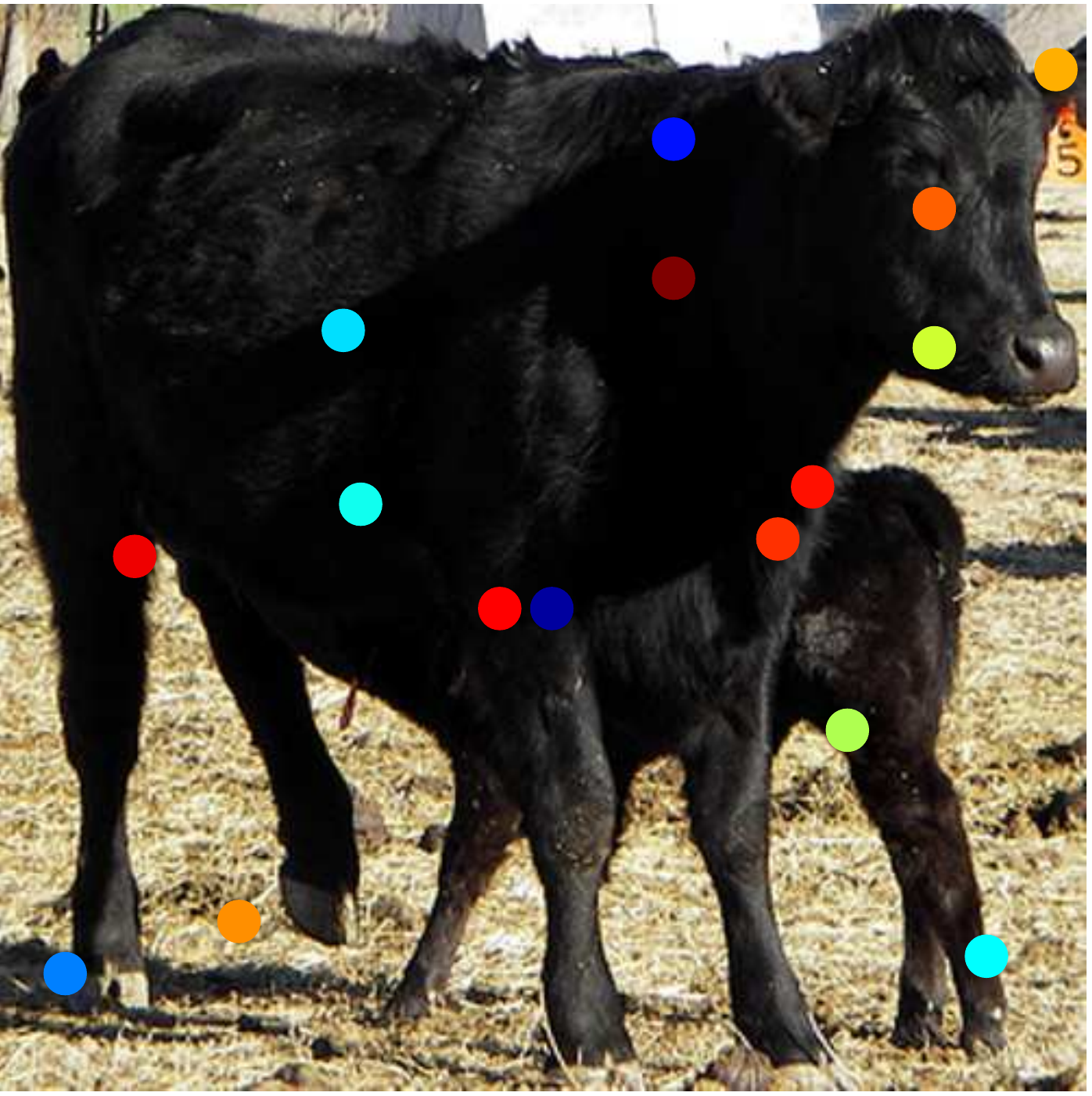} &
\includegraphics[width=0.17\textwidth]{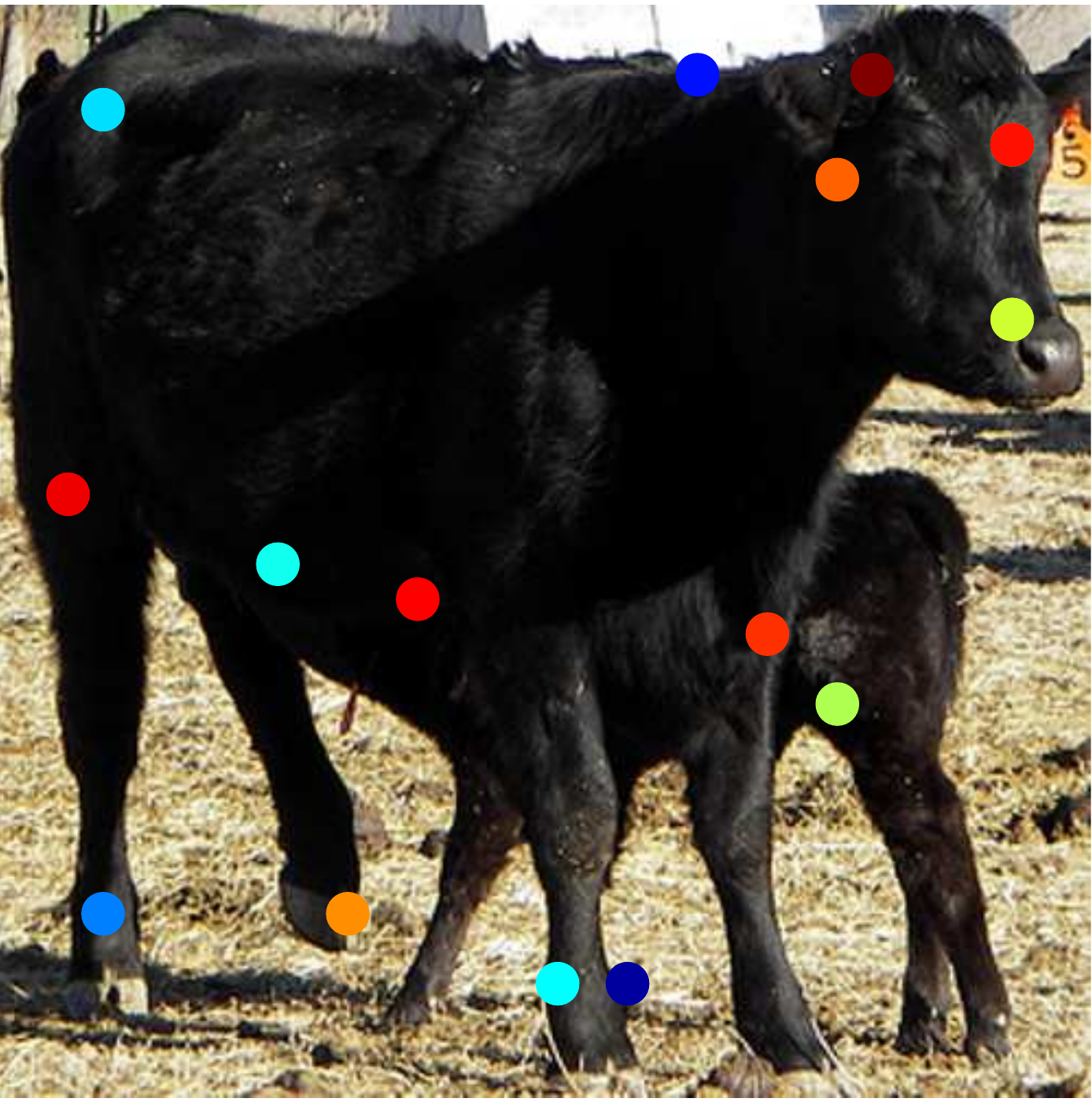} & &
\includegraphics[width=0.17\textwidth]{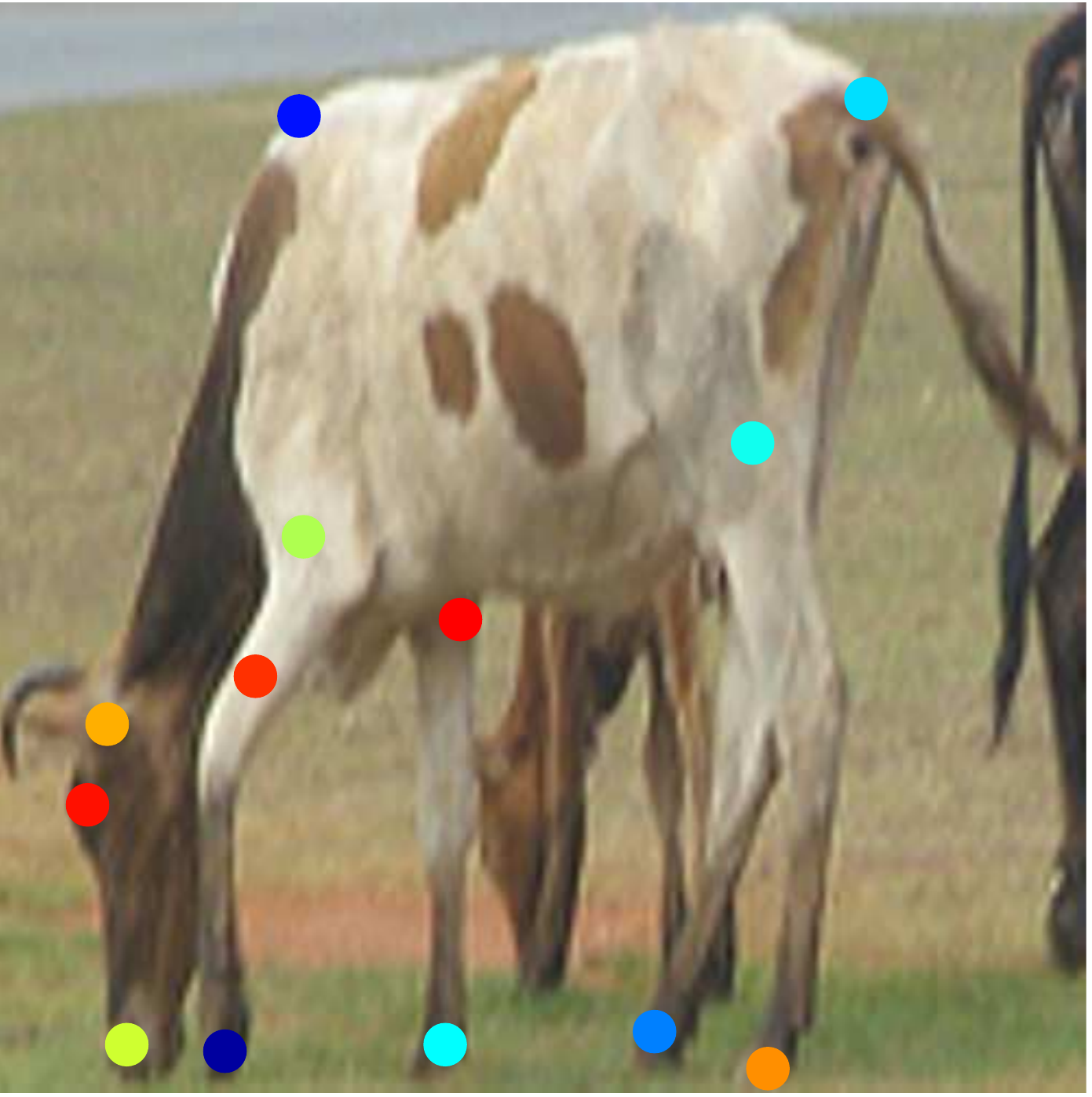} &
\includegraphics[width=0.17\textwidth]{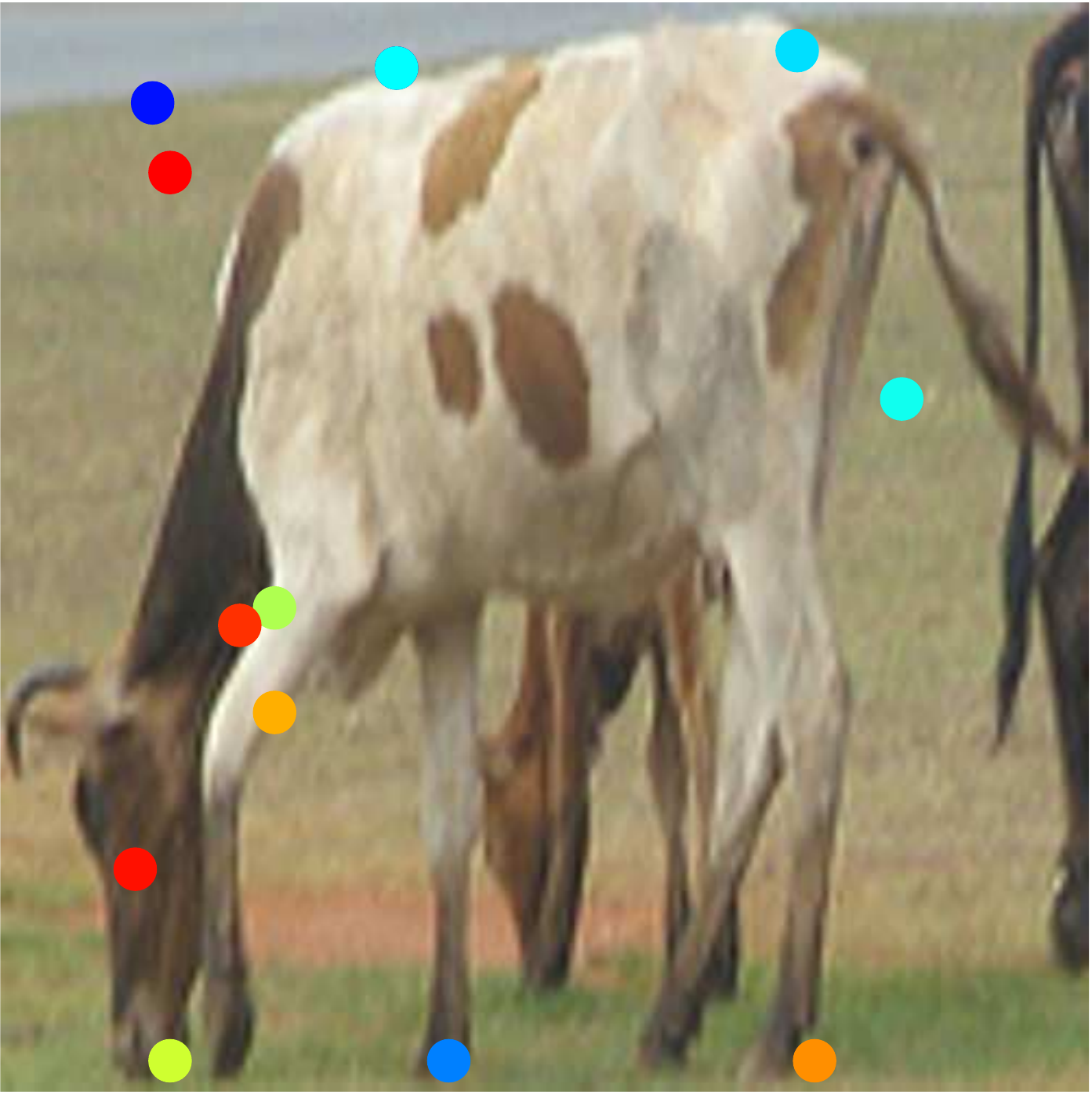} &
\includegraphics[width=0.17\textwidth]{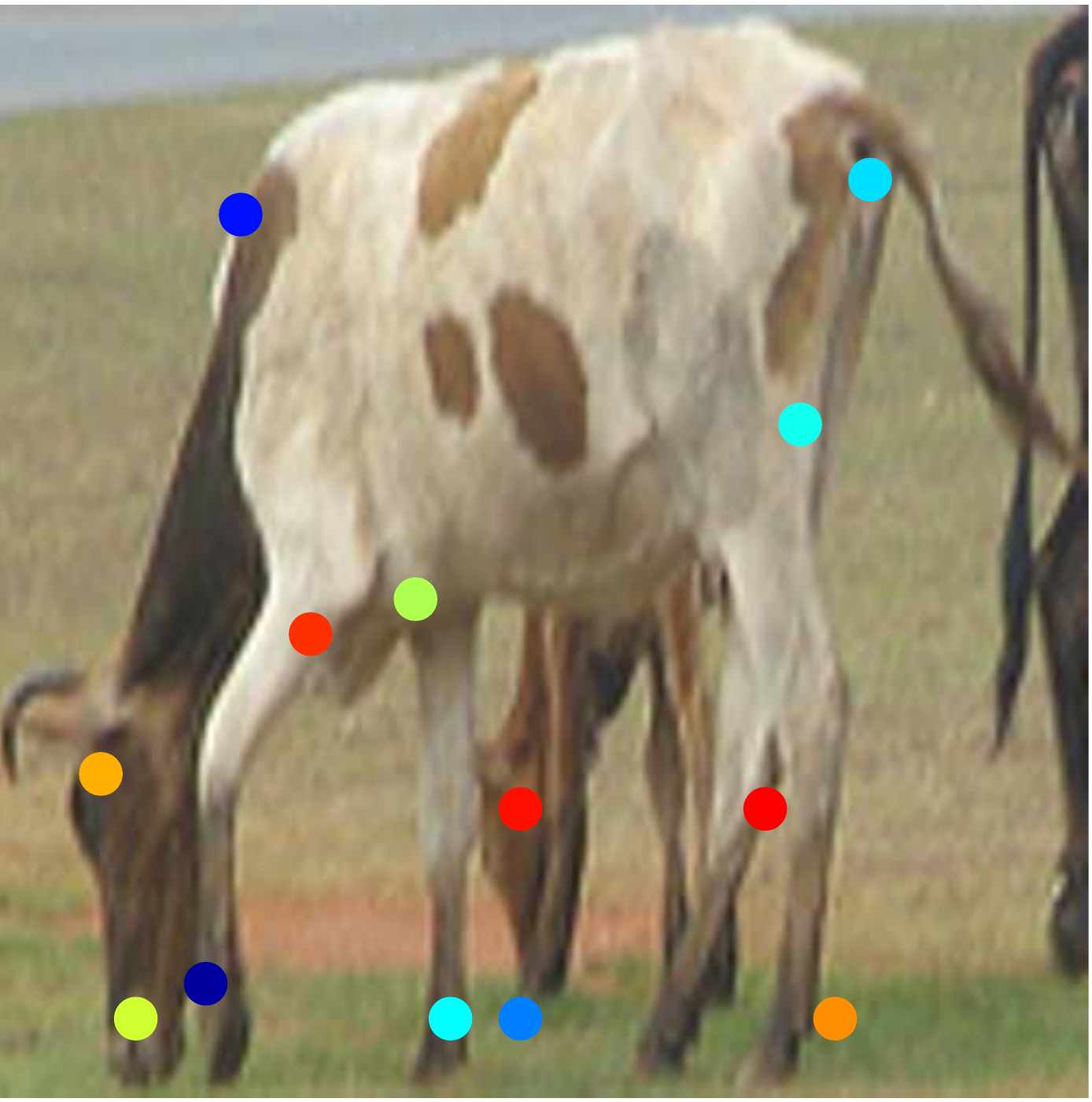} \\
\includegraphics[width=0.17\textwidth]{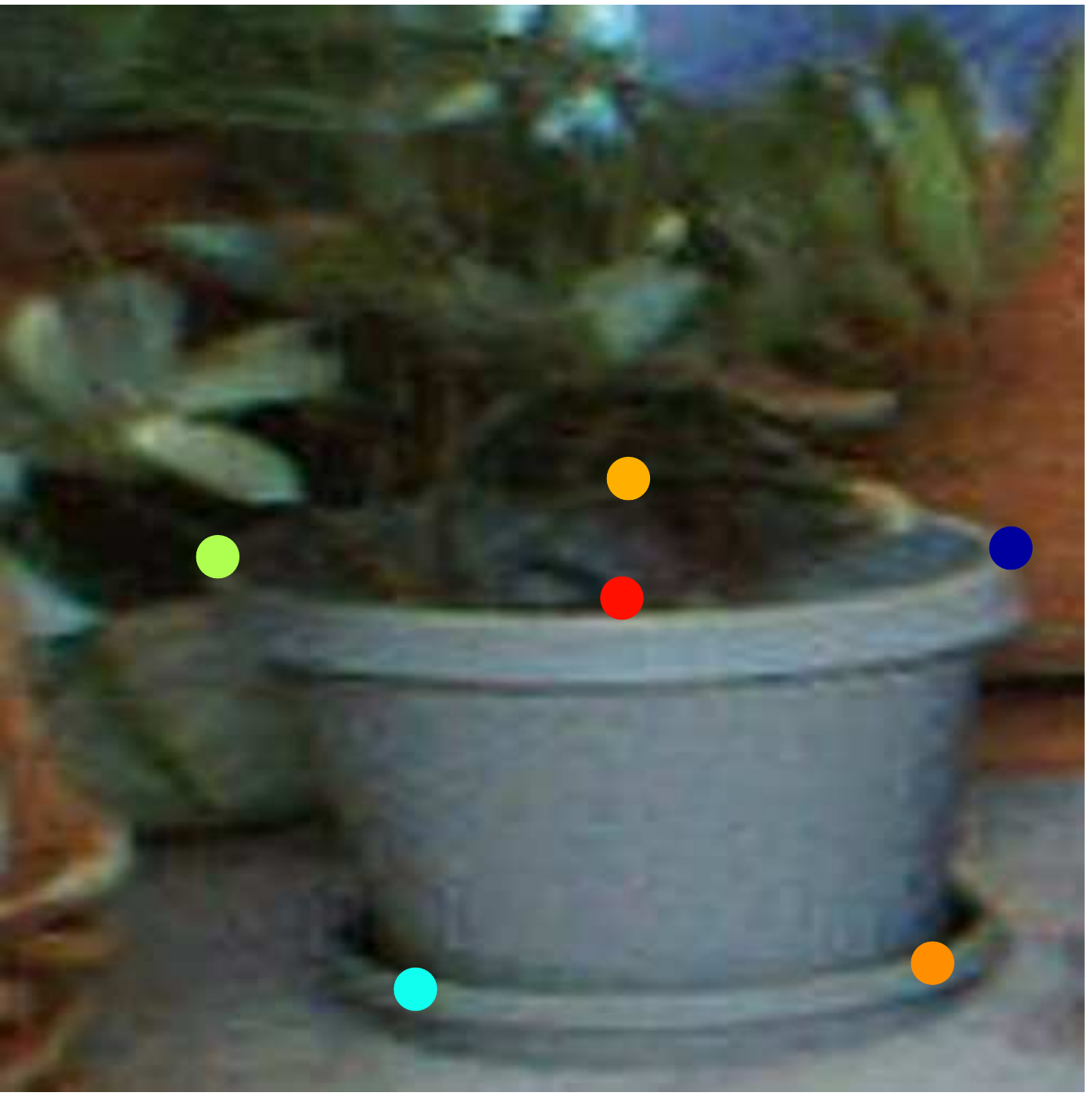} &
\includegraphics[width=0.17\textwidth]{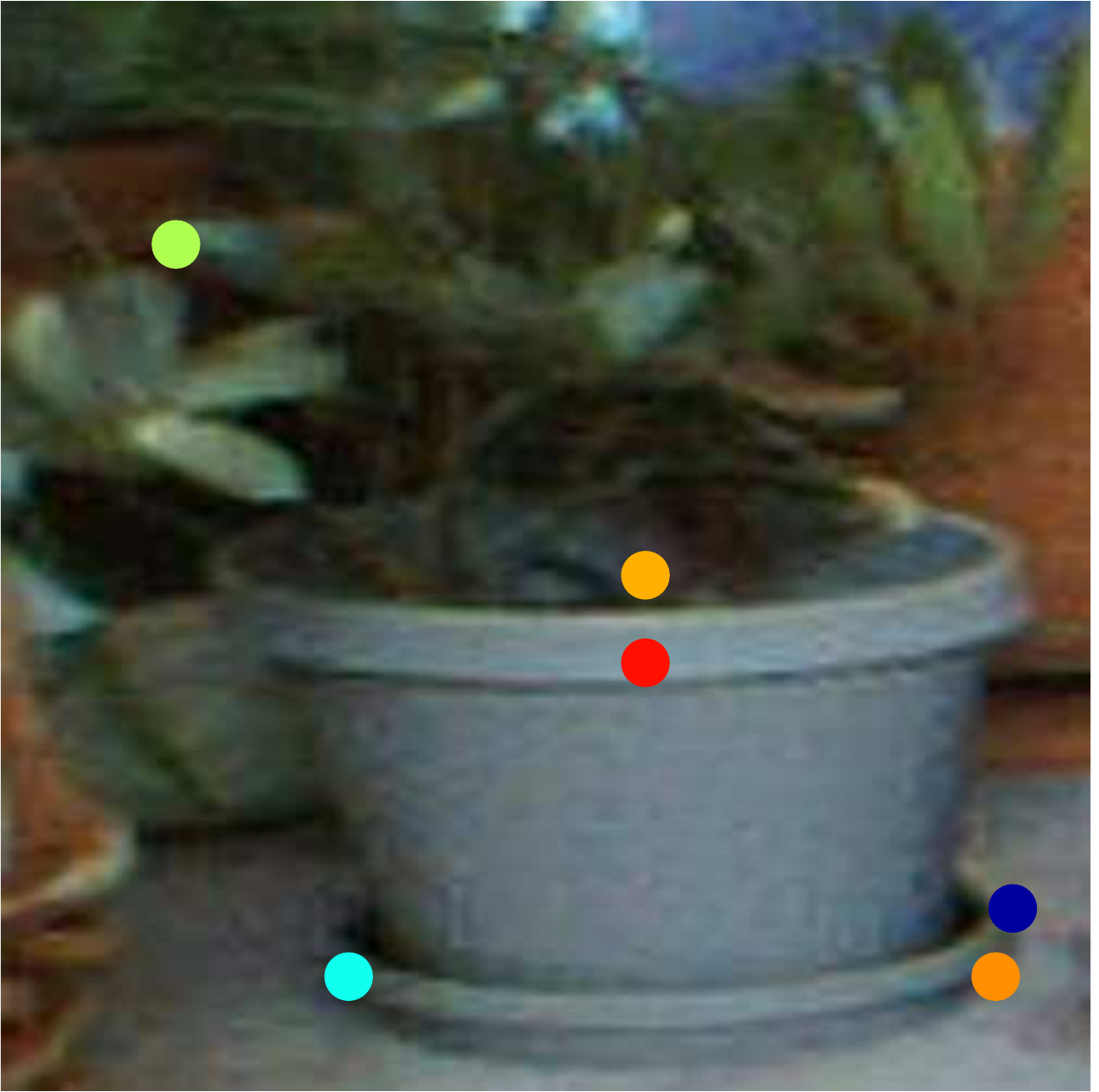} &
\includegraphics[width=0.17\textwidth]{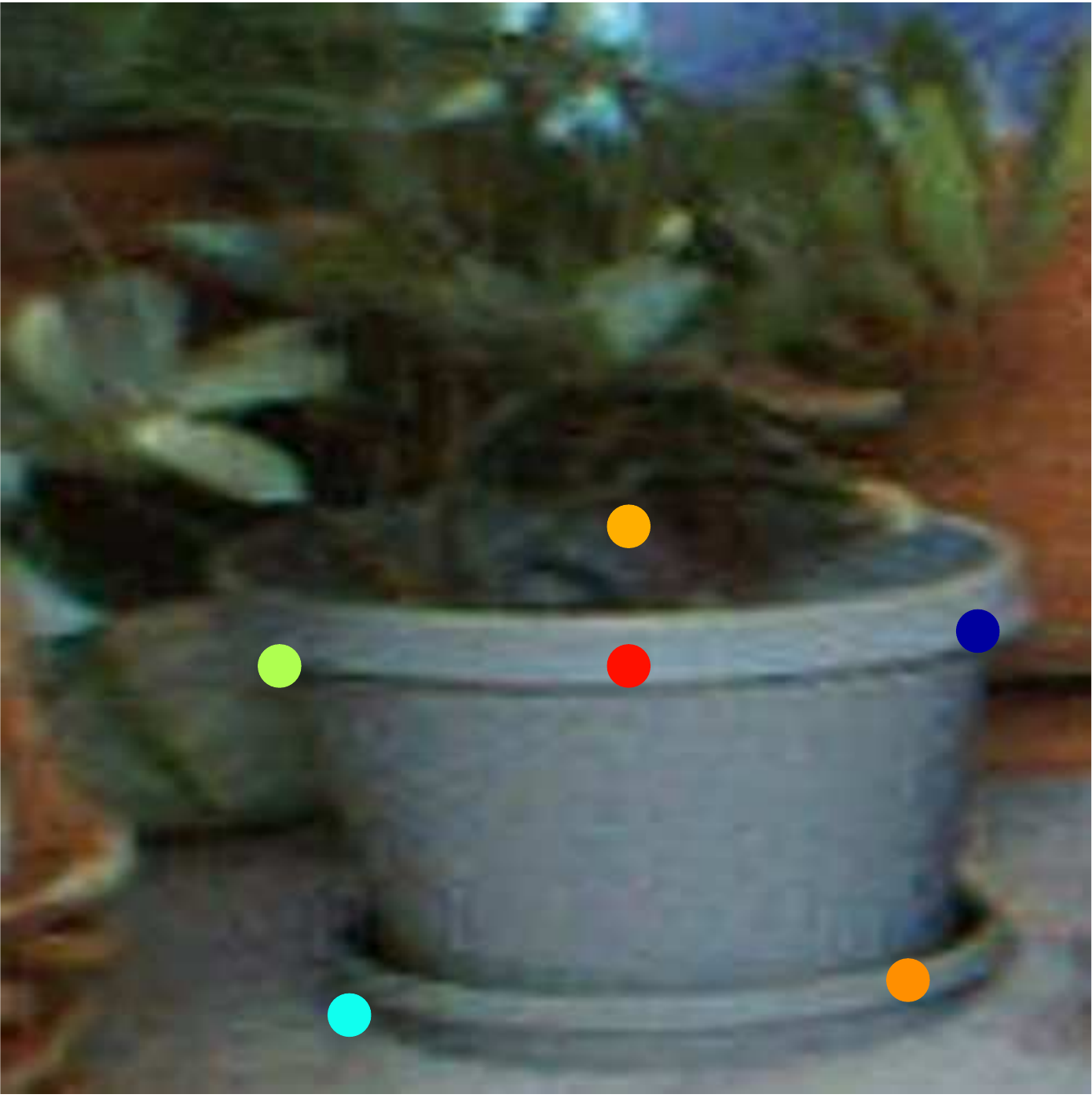} & &
\includegraphics[width=0.17\textwidth]{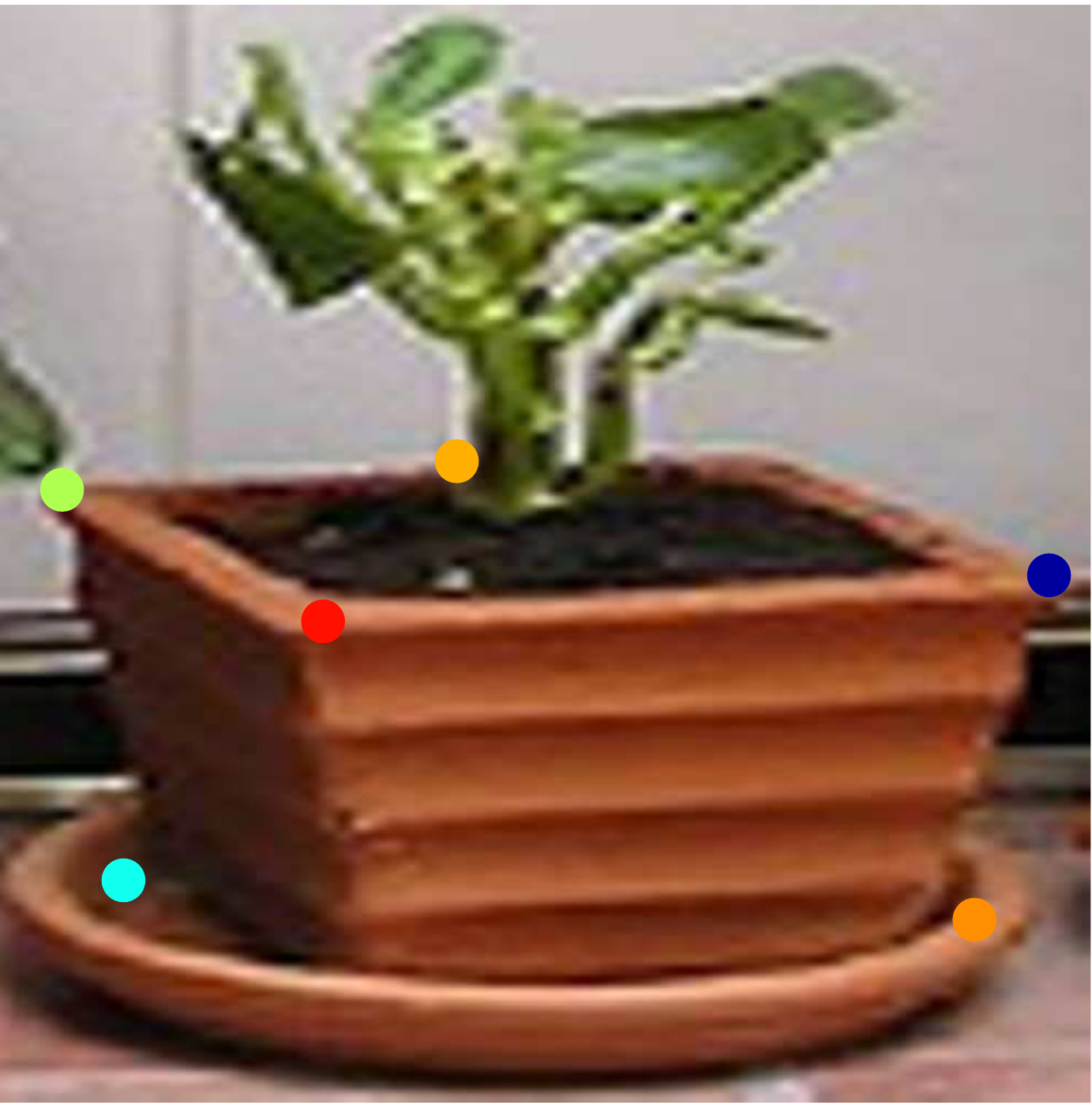} &
\includegraphics[width=0.17\textwidth]{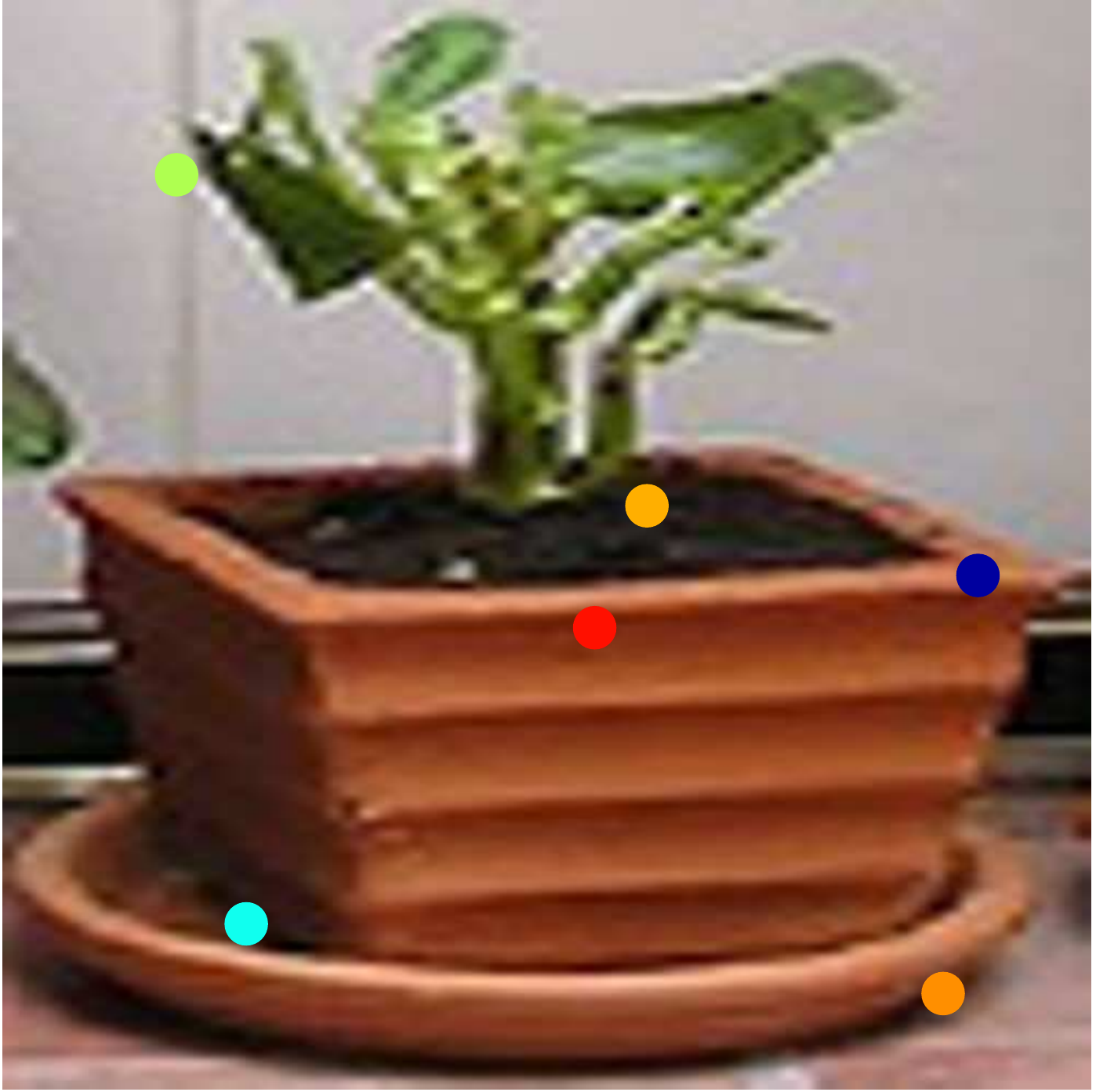} &
\includegraphics[width=0.17\textwidth]{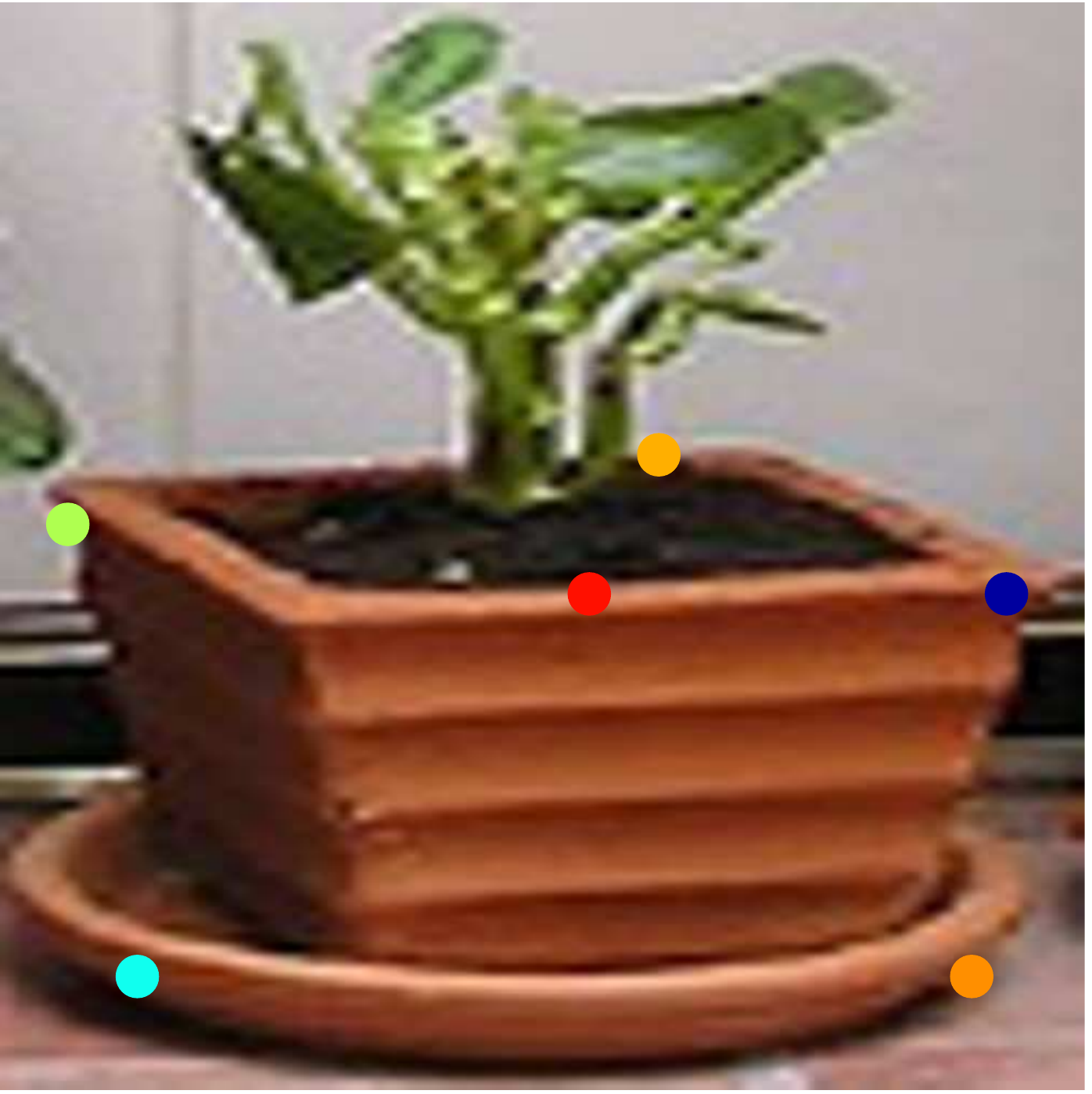} \\
\includegraphics[width=0.17\textwidth]{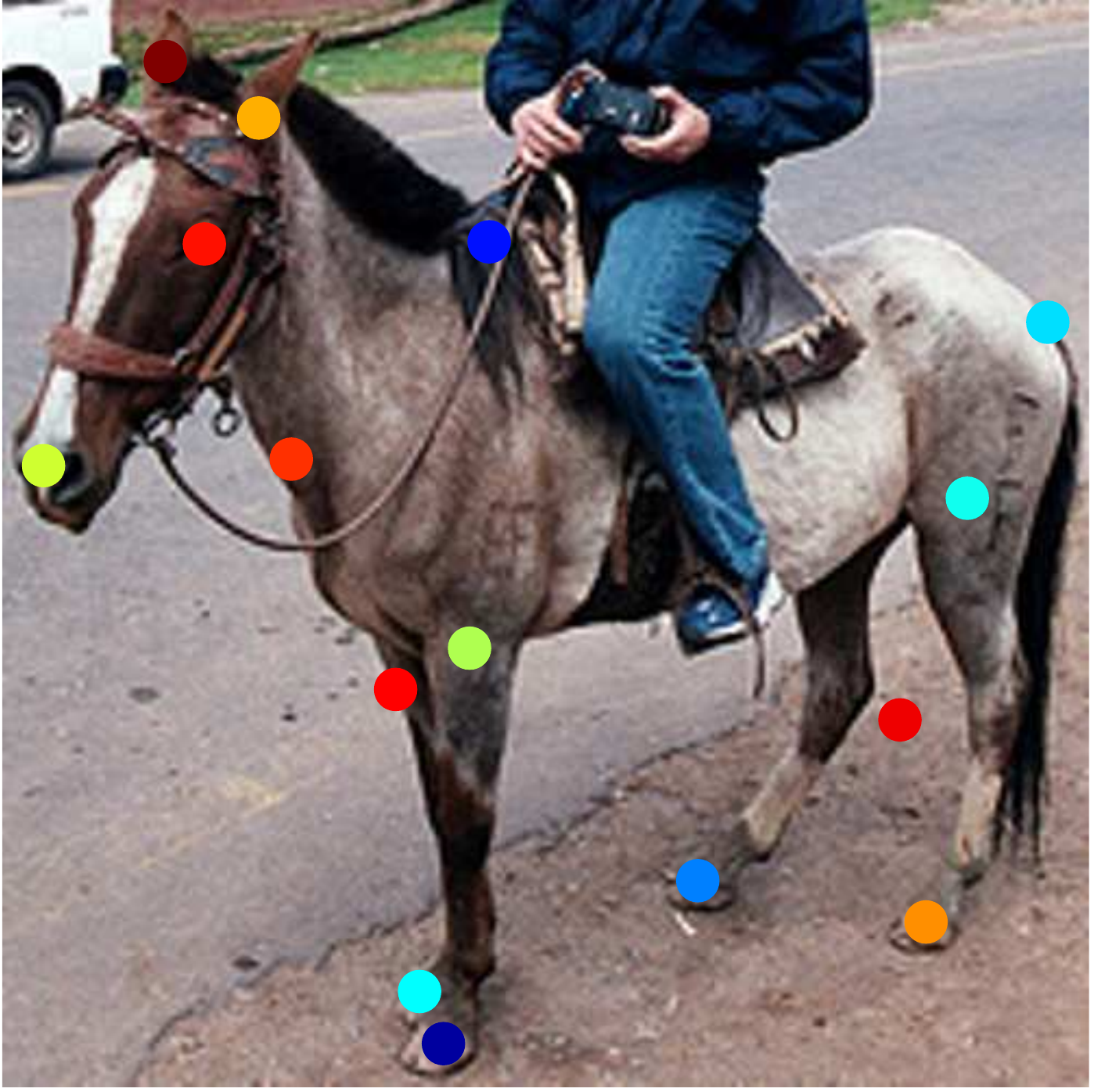} &
\includegraphics[width=0.17\textwidth]{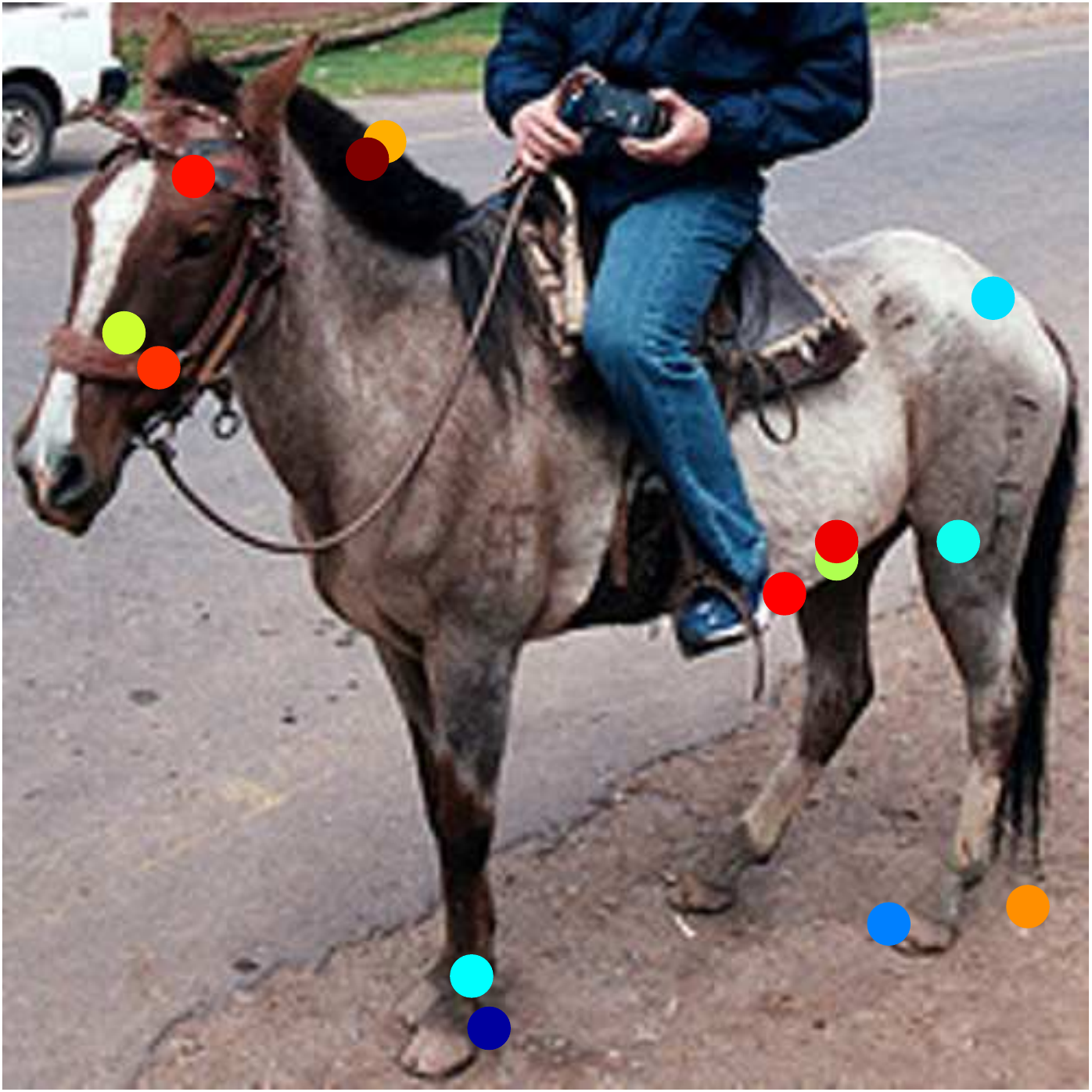} &
\includegraphics[width=0.17\textwidth]{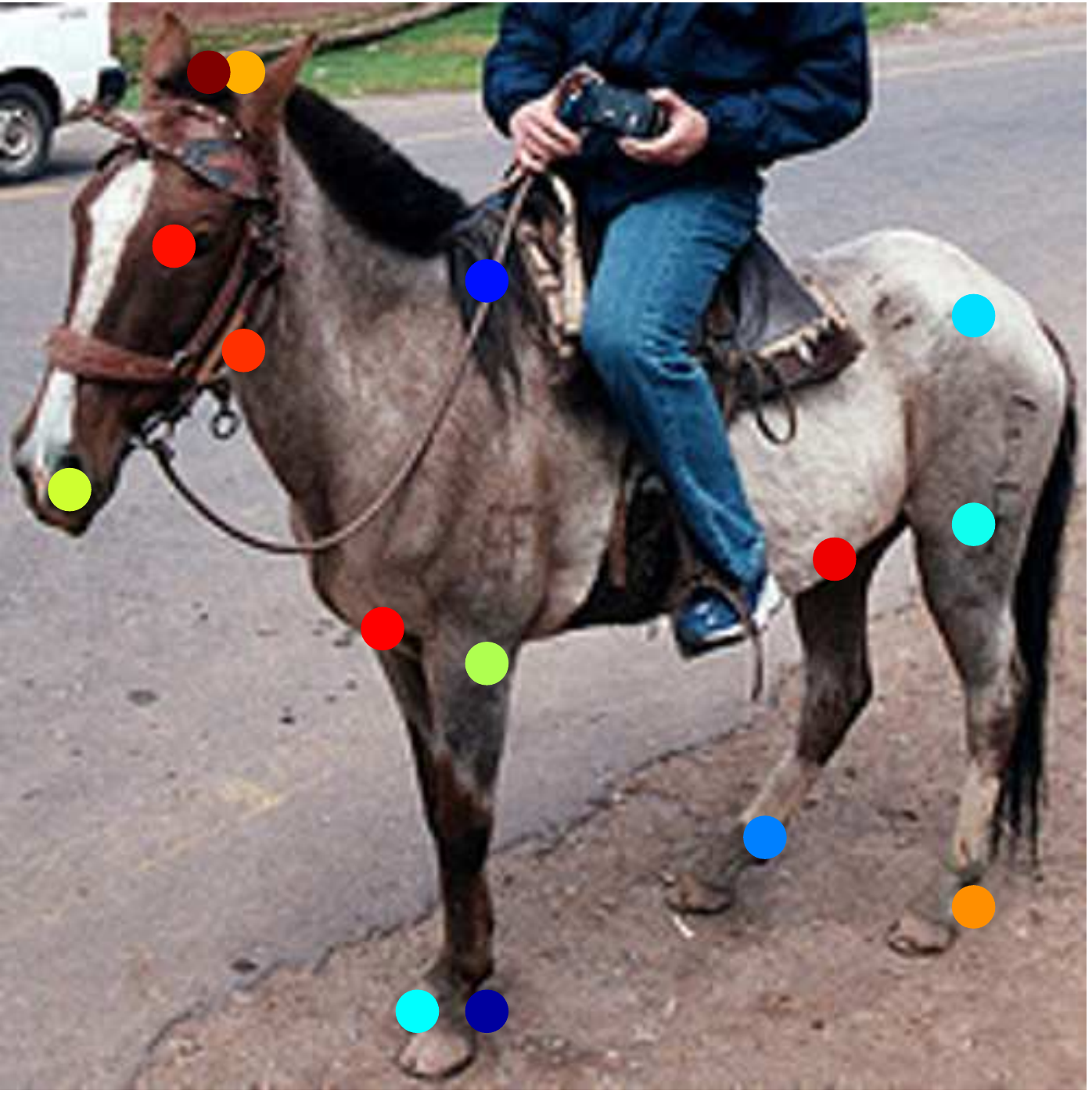} & &
\includegraphics[width=0.17\textwidth]{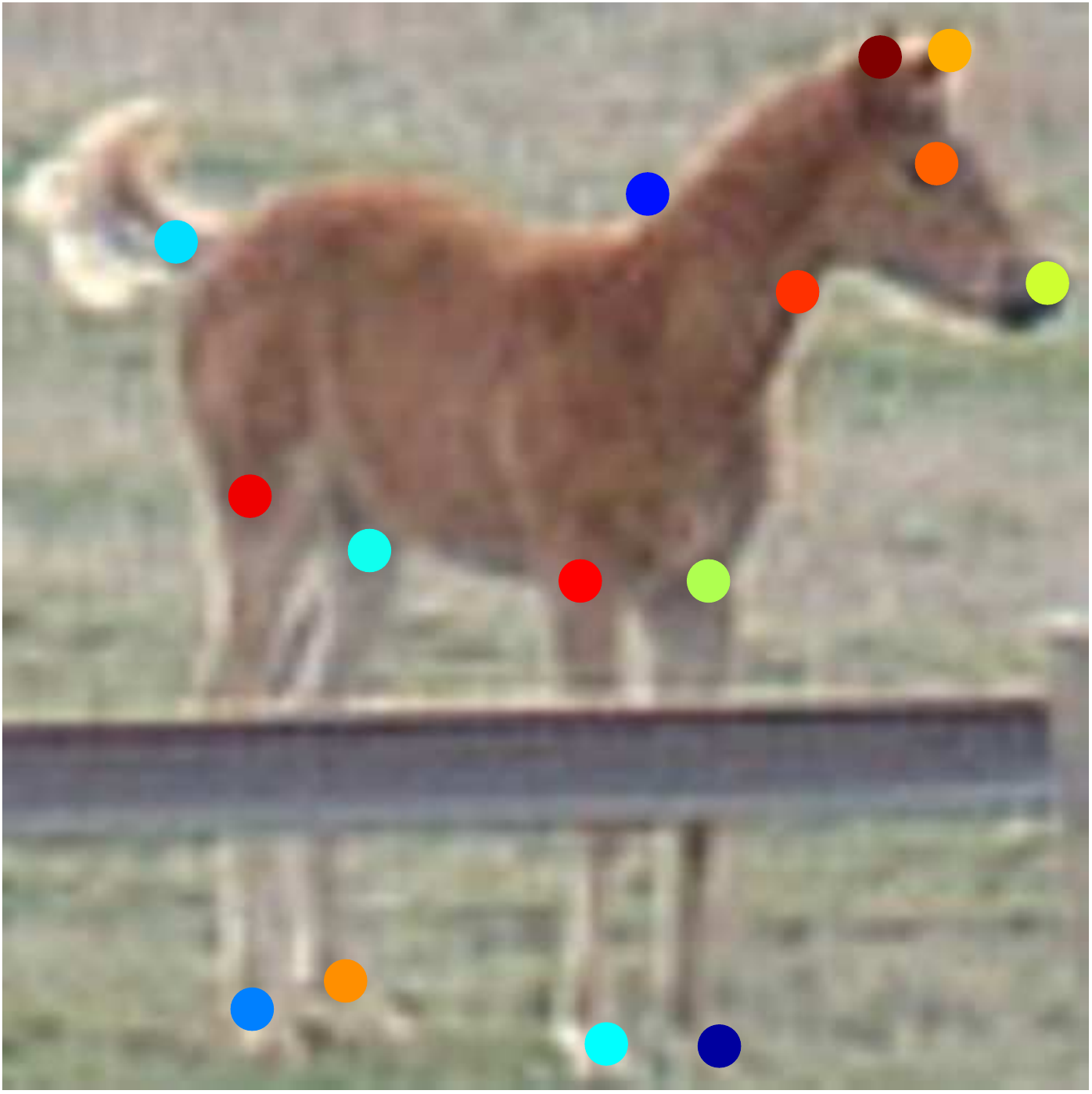} &
\includegraphics[width=0.17\textwidth]{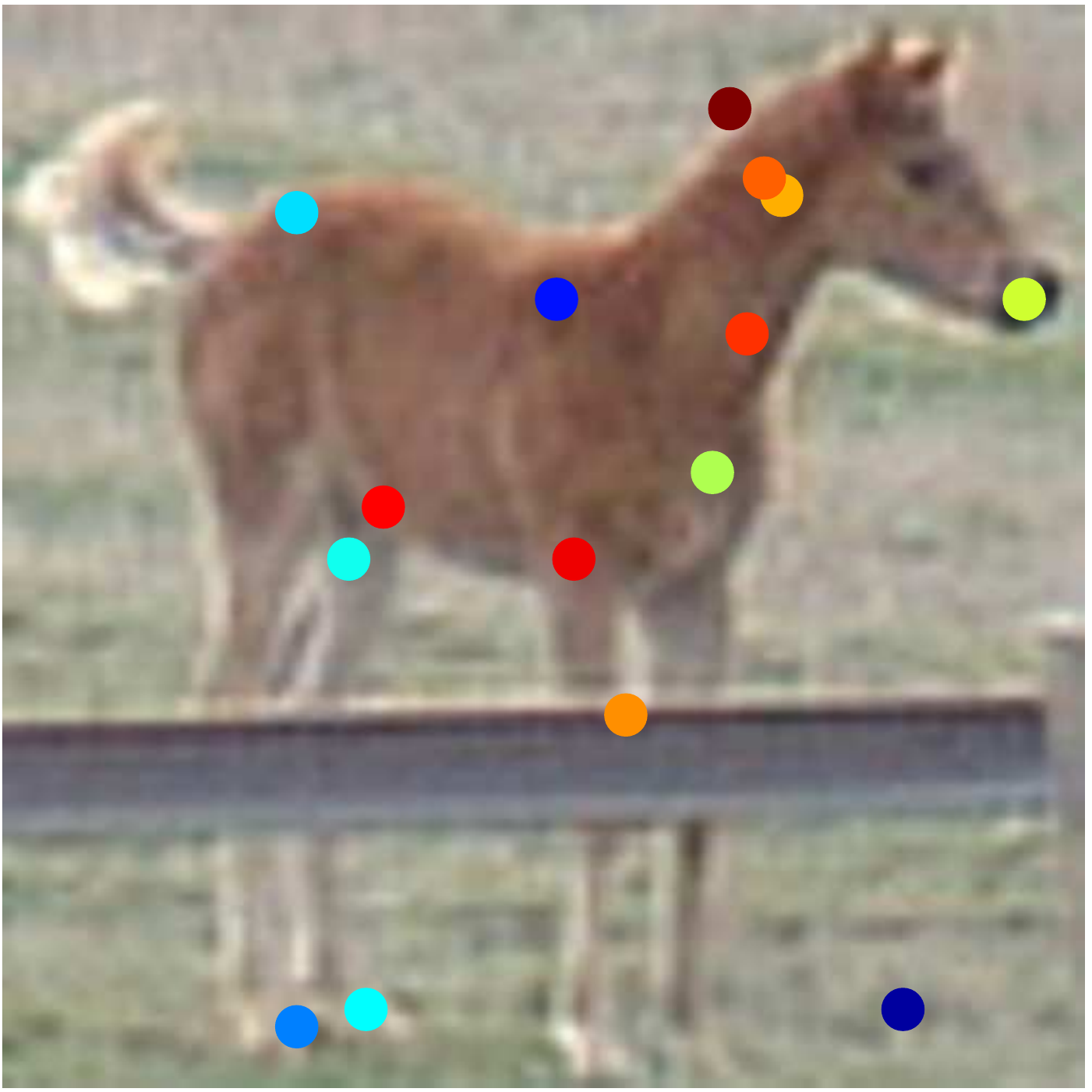} &
\includegraphics[width=0.17\textwidth]{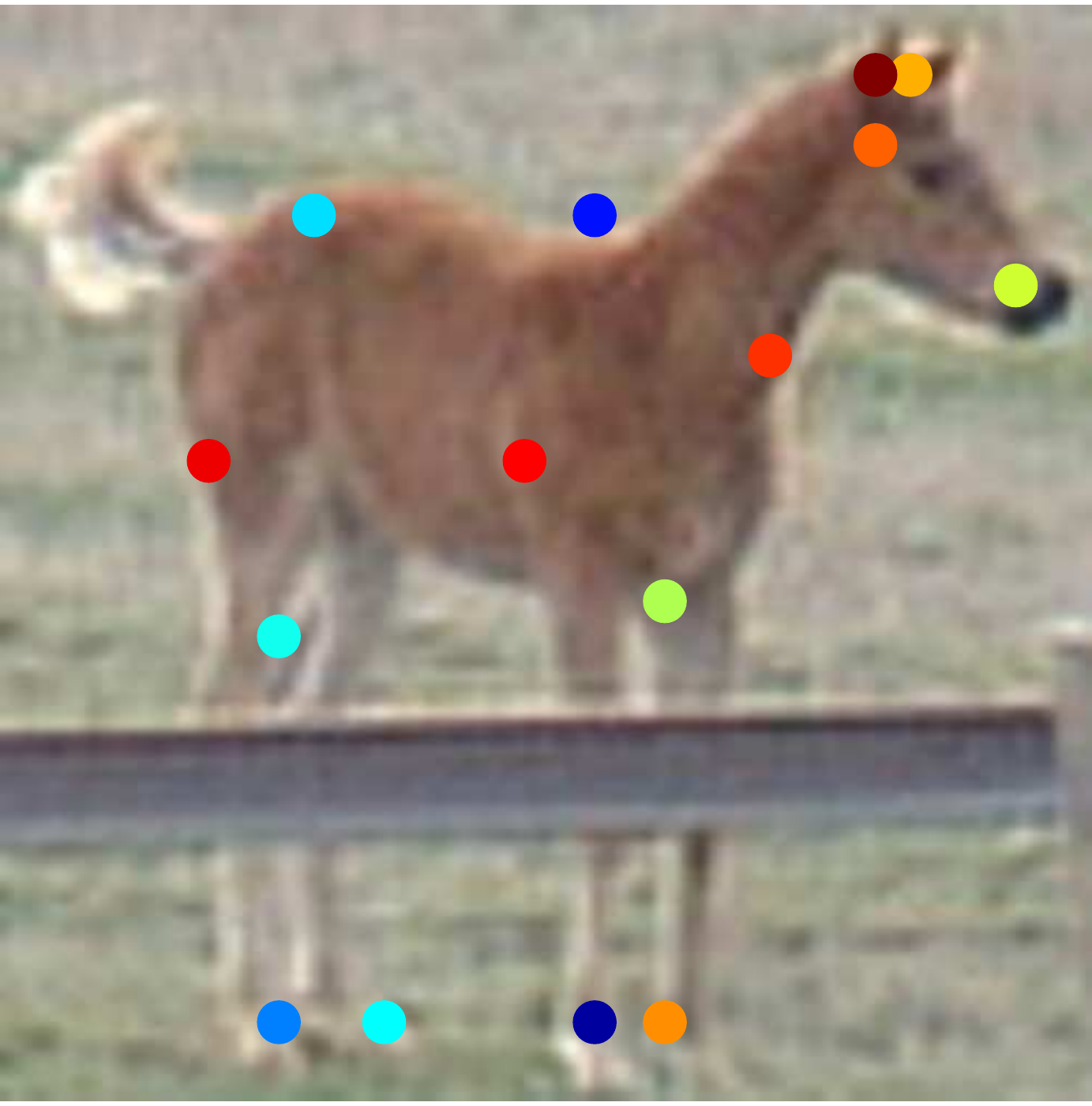} \\
\end{tabular}
}
\captionof{figure}{
Examples of keypoint prediction on five classes of the PASCAL dataset: aeroplane, cat, cow, potted plant, and horse. Each keypoint is associated with one color. The first column is the ground truth annotation, the second column is the prediction result of SIFT+prior and the third column is conv5+prior. (Best viewed in color).
}
\label{fig:keypointprediction}
\end{minipage}
\end{figure}

We evaluate the predicted keypoints using the measure PCK introduced
in Section \ref{sec:flow}, taking $\alpha = 0.1$. A predicted keypoint is defined as correct
if the distance between it and the ground truth keypoint is less than $\alpha \cdot \max(h, w)$ where $h$
 and $w$ are the height and width of the bounding box.
The results using \texttt{conv}5 and SIFT
with and without the prior are shown in Table \ref{table:keypointprediction}. From
the table, we can see that local part detectors trained on the \texttt{conv}5 feature
outperform SIFT by a large margin and that the prior information is helpful
in both cases.
To our knowledge, these are the first keypoint prediction results reported on this
dataset.
We
show example results from five different categories in Figure
\ref{fig:keypointprediction}. Each set consists of rescaled bounding box images
with ground truth keypoint annotations and predicted keypoints using SIFT and
\texttt{conv}5 features,  where each color corresponds to one keypoint. As the
figure
shows, \texttt{conv}5 outperforms SIFT,
often managing satisfactory outputs despite the challenge of this
task. A small offset can be noticed for some keypoints like eyes and noses, likely
due to the limited
stride of our scanning windows. A final regression or finer
stride could mitigate this issue.

\section{Conclusion}

Through visualization, alignment, and keypoint prediction, we have studied the
ability of the intermediate features implicitly learned in a state-of-the-art
convnet classifier to understand specific, local correspondence.
Despite their large receptive fields and weak label training, we have found in
all cases that convnet features are at least as useful (and sometimes
considerably more useful) than conventional ones for extracting local
visual information.
% add a broad statement of grand philosophical implications here?
% Include some future work?

{\small
\textbf{Acknowledgements}\hspace{1ex} This work was supported in part by DARPA's MSEE and SMISC programs, by NSF awards
IIS-1427425, IIS-1212798, and IIS-1116411, and by support from Toyota.
}

\subsubsection*{References}

\begingroup
\renewcommand{\section}[2]{}
\small{
\bibliographystyle{unsrt}
\bibliography{bib}

\begin{thebibliography}{10}

\bibitem{ImageNet}
J.~Deng, W.~Dong, R.~Socher, L.-J. Li, K.~Li, and L.~Fei-Fei.
\newblock {ImageNet: A Large-Scale Hierarchical Image Database}.
\newblock In {\em CVPR}, 2009.

\bibitem{Krizhevsky}
A.~Krizhevsky, I.~Sutskever, and G.~Hinton.
\newblock Imagenet classification with deep convolutional neural networks.
\newblock In {\em NIPS}, 2012.

\bibitem{RossJeff}
R.~Girshick, J.~Donahue, T.~Darrell, and J.~Malik.
\newblock Rich feature hierarchies for accurate object detection and semantic
  segmentation.
\newblock In {\em CVPR}, 2014.

\bibitem{pascal}
M.~Everingham, L.~Van~Gool, C.~K.~I. Williams, J.~Winn, and A.~Zisserman.
\newblock The {PASCAL} {V}isual {O}bject {C}lasses {C}hallenge 2011 {(VOC2011)}
  {R}esults.
\newblock
  http://www.pascal-network.org/challenges/VOC/voc2011/workshop/index.html.

\bibitem{YannLeCunsGooglePlusPage}
{Debate on Yann LeCun's Google+ page}.
\newblock https://plus.google.com/+YannLeCunPhD/posts/JBBFfv2XgWM.
\newblock Accessed: 2014-5-31.

\bibitem{congealing}
G.~B. Huang, V.~Jain, and E.~Learned-Miller.
\newblock Unsupervised joint alignment of complex images.
\newblock In {\em ICCV}, 2007.

\bibitem{deep_congealing}
G.~B. Huang, M.~A. Mattar, H.~Lee, and E.~Learned-Miller.
\newblock Learning to align from scratch.
\newblock In {\em NIPS}, 2012.

\bibitem{sift-flow}
C.~Liu, J.~Yuen, and A.~Torralba.
\newblock Sift flow: Dense correspondence across scenes and its applications.
\newblock {\em PAMI}, 33(5):978--994, 2011.

\bibitem{iccv13_keypoint}
J.~Liu and P.~N. Belhumeur.
\newblock Bird part localization using exemplar-based models with enforced pose
  and subcategory consistenty.
\newblock In {\em ICCV}, 2013.

\bibitem{poof}
T.~Berg and P.~N. Belhumeur.
\newblock {POOF}: Part-based one-vs.-one features for fine-grained
  categorization, face verification, and attribute estimation.
\newblock In {\em CVPR}, 2013.

\bibitem{Belhumeur_Localizing_2011}
P.~N. Belhumeur, D.~W. Jacobs, D.~J. Kriegman, and N.~Kumar.
\newblock Localizing parts of faces using a consensus of exemplars.
\newblock In {\em CVPR}, 2011.

\bibitem{YangRamananCVPR11}
Y.~Yang and D.~Ramanan.
\newblock Articulated pose estimation using flexible mixtures of parts.
\newblock In {\em CVPR}, 2011.

\bibitem{Min_Sun_ICCV11}
M.~Sun and S.~Savarese.
\newblock Articulated part-based model for joint object detection and pose
  estimation.
\newblock In {\em ICCV}, 2011.

\bibitem{Deva_2012}
X.~Zhu and D.~Ramanan.
\newblock Face detection, pose estimation, and landmark localization in the
  wild.
\newblock In {\em CVPR}, 2012.

\bibitem{BourdevMalikICCV09}
L.~Bourdev and J.~Malik.
\newblock Poselets: Body part detectors trained using 3d human pose
  annotations.
\newblock In {\em ICCV}, 2009.

\bibitem{kposelet}
G.~Gkioxari, B.~Hariharan, R.~Girshick, and J.~Malik.
\newblock Using k-poselets for detecting people and localizing their keypoints.
\newblock In {\em CVPR}, 2014.

\bibitem{armlet}
G.~Gkioxari, P.~Arbelaez, L.~Bourdev, and J.~Malik.
\newblock Articulated pose estimation using discriminative armlet classifiers.
\newblock In {\em CVPR}, 2013.

\bibitem{Lecun89}
Y.~LeCun, B.~Boser, J.S. Denker, D.~Henderson, R.~E. Howard, W.~Hubbard, and
  L.~D. Jackel.
\newblock Backpropagation applied to hand-written zip code recognition.
\newblock In {\em Neural Computation}, 1989.

\bibitem{Lecun98OCR}
Y.~Lecun, L.~Bottou, Y.~Bengio, and P.~Haffner.
\newblock Gradient-based learning applied to document recognition.
\newblock In {\em Proceedings of the IEEE}, pages 2278--2324, 1998.

\bibitem{decaf}
J.~Donahue, Y.~Jia, O.~Vinyals, J.~Hoffman, N.~Zhang, E.~Tzeng, and T.~Darrell.
\newblock {DeCAF}: A deep convolutional activation feature for generic visual
  recognition.
\newblock In {\em ICML}, 2014.

\bibitem{sermanet-cvpr13}
P.~Sermanet, K.~Kavukcuoglu, S.~Chintala, and Y.~LeCun.
\newblock Pedestrian detection with unsupervised multi-stage feature learning.
\newblock In {\em CVPR}, 2013.

\bibitem{deep_pose}
A.~Toshev and C.~Szegedy.
\newblock {DeepPose}: Human pose estimation via deep neural networks.
\newblock In {\em CVPR}, 2014.

\bibitem{HumanPoseICLR}
A.~Jain, J.~Tompson, M.~Andriluka, G.~W. Taylor, and C.~Bregler.
\newblock Learning human pose estimation features with convolutional networks.
\newblock In {\em ICLR}, 2014.

\bibitem{ZF}
M.~D Zeiler and R.~Fergus.
\newblock Visualizing and understanding convolutional neural networks.
\newblock In {\em ECCV}, 2014.

\bibitem{Szegedy}
C.~Szegedy, W.~Zaremba, I.~Sutskever, J.~Bruna, D.~Erhan, I.~Goodfellow, and
  R.~Fergus.
\newblock Intriguing properties of neural networks.
\newblock In {\em ICLR}, 2014.

\bibitem{Fischer}
P.~{Fischer}, A.~{Dosovitskiy}, and T.~{Brox}.
\newblock {Descriptor Matching with Convolutional Neural Networks: a Comparison
  to SIFT}.
\newblock {\em ArXiv e-prints}, May 2014.

\bibitem{caffe}
Y.~Jia, E.~Shelhamer, J.~Donahue, S.~Karayev, J.~Long, R.~Girshick,
  S.~Guadarrama, and T.~Darrell.
\newblock Caffe: Convolutional architecture for fast feature embedding.
\newblock {\em arXiv preprint arXiv:1408.5093}, 2014.

\bibitem{hoggles}
C.~Vondrick, A.~Khosla, T.~Malisiewicz, and A.~Torralba.
\newblock {HOGgles: Visualizing Object Detection Features}.
\newblock In {\em ICCV}, 2013.

\bibitem{FHBP}
P.~Felzenszwalb and D.~P. Huttenlocher.
\newblock Efficient belief propagation for early vision.
\newblock {\em International journal of computer vision}, 70(1):41--54, 2006.

\bibitem{FHDT}
P.~Felzenszwalb and D.~Huttenlocher.
\newblock Distance transforms of sampled functions.
\newblock Technical report, Cornell University, 2004.

\bibitem{scipy}
E.~Jones, T.~Oliphant, P.~Peterson, et~al.
\newblock {SciPy}: Open source scientific tools for {Python}, 2001.

\bibitem{PCP}
Y.~Yang and D.~Ramanan.
\newblock Articulated human detection with flexible mixtures of parts.
\newblock In {\em PAMI}, 2013.

\bibitem{SIFT}
D.G. Lowe.
\newblock Object recognition from local scale-invariant features.
\newblock In {\em ICCV}, 1999.

\bibitem{overfeat}
P.~Sermanet, D.~Eigen, X.~Zhang, M.~Mathieu, R.~Fergus, and Y.~LeCun.
\newblock Overfeat: Integrated recognition, localization and detection using
  convolutional networks.
\newblock In {\em ICLR}, 2014.

\bibitem{selsearch}
J.~Uijlings, K.~van~de Sande, T.~Gevers, and A.~Smeulders.
\newblock Selective search for object recognition.
\newblock {\em IJCV}, 2013.

\bibitem{vedaldi08vlfeat}
A.~Vedaldi and B.~Fulkerson.
\newblock {VLFeat}: An open and portable library of computer vision algorithms.
\newblock http://www.vlfeat.org/, 2008.

\end{thebibliography}
}
\endgroup

\end{document}